\documentclass{article}

\PassOptionsToPackage{numbers, compress}{natbib}

\usepackage[final]{neurips_2019}

\usepackage{microtype}
\usepackage{graphicx}
\usepackage{subfigure}
\usepackage{booktabs} 
\usepackage{multirow}
\usepackage[export]{adjustbox}
\usepackage{natbib}
\usepackage[utf8]{inputenc} 
\usepackage[T1]{fontenc}    
\usepackage{url}            
\usepackage{booktabs}       
\usepackage{amsfonts}       
\usepackage{nicefrac}       
\usepackage{xspace}
\usepackage{amsmath}
\usepackage{caption}
\graphicspath{{../diagrams/}{./}}
\usepackage{epstopdf}
\usepackage{pstricks}
\usepackage{smartdiagram}
\usepackage{tikz}
\usepackage{tcolorbox}
\usepackage{subfigure}
\usepackage[textsize=footnotesize,color=green!40]{todonotes}
\usepackage{algorithm}
\usepackage{hyperref}
\usepackage{bm}
\usepackage[color=green!40]{todonotes}
\usepackage{amsmath,amsthm,amssymb, amsfonts}
\usepackage{mathtools}
\usepackage{graphicx} 
\usepackage{algpseudocode}
\usepackage{algorithm}
\usepackage{wrapfig}

\newcommand{\profet}{{\sc{Profet}{}}}

\newcommand{\vx}{{\bm{x}}}
\newcommand{\vy}{{\bm{y}}}

\newcommand{\vh}{\mathbf{h}}

\newcommand{\vtheta}{{\bm{\theta}}}

\newcommand{\ie}{i.\,e.}

\newcommand{\eg}{e.\,g.}

\newcommand{\gauss}{\mbox{${\cal N}$}}

\newcommand{\acr}[1]{\textsc{#1}\xspace}

\newcommand{\coco}{\acr{COCO}}

\newcommand{\hpo}{\acr{HPO}}
\newcommand{\gp}{\acr{GP}}

\newcommand{\svm}{\acr{\svm}}
\newcommand{\I}{\mathbf{I}}

\newcommand{\gplvm}{\acr{GP-LVM}}
\newcommand{\bnn}{\acr{BNN}}

\title{Meta-Surrogate Benchmarking for Hyperparameter Optimization}

\author{
Aaron Klein$^1$~~~~~~Zhenwen Dai$^2$~~~~~~Frank Hutter$^1$~~~~~~Neil Lawrence$^3$~~~~~~Javier Gonz\'{a}lez$^2$ \\
  $^1$University of Freiburg ~~~~~~ $^2$Amazon Cambridge ~~~~~~ $^3$University of Cambridge\\ 
  \texttt{\{kleinaa,fh\}@cs.uni-freiburg.de} \\ \texttt{\{zhenwend, gojav\}@amazon.com} \\ \texttt{ndl21@cam.ac.uk} \\
}

\begin{document}

\maketitle

\begin{abstract}

  Despite the recent progress in hyperparameter optimization (\hpo), available benchmarks that resemble real-world scenarios consist of a few and very large problem instances that are expensive to solve.
This blocks researchers and practitioners not only from systematically running large-scale comparisons that are needed to draw statistically significant results but also from reproducing experiments that were conducted before.
This work proposes a method to alleviate these issues by means of a meta-surrogate model for \hpo tasks trained on off-line generated data.
The model combines a probabilistic encoder with a multi-task model such that it can generate inexpensive and realistic tasks of the class of problems of interest.
We demonstrate that benchmarking \hpo methods on samples of the generative model allows us to draw more coherent and statistically significant conclusions that can be reached orders of magnitude faster than using the original tasks. We provide evidence of our findings for various \hpo methods on a wide class of problems.

\end{abstract}

\section{Introduction}\label{sec:intro}

Automated Machine Learning (AutoML) \citep{automl-book} is an emerging field that studies the progressive automation of machine learning.
A core part of an AutoML system is the hyperparameter optimization (\hpo) of a machine learning algorithm.
It has already shown promising results by outperforming human experts in finding better hyperparameters \citep{snoek-nips12a}, an thereby, for example, substantially improved AlphaGo~\citep{chen-arxiv18a}. 

Despite recent progress (see \eg{} the review by \citet{feurer-automlbook18a}), during the phases of developing and evaluating new \hpo methods one frequently faces the following problems:
\begin{itemize}
  \item  Evaluating the objective function is often expensive in terms of wall-clock time; \emph{e.g.}, the evaluation of a single hyperparameter configuration may take several hours or days. This renders extensive \hpo or repeated runs of \hpo methods computationally infeasible.

  \item  Even though repositories of datasets, such as OpenML~\citep{vanschoren-sigkdd13a} provide thousands of datasets, a large fraction cannot meaningfully be used for \hpo since they are too small or too easy (in the sense that even simple methods achieve top performance). Hence, useful available datasets are scarce, making it hard to produce a comprehensive evaluation of how well a \hpo method will generalize across tasks.
\end{itemize}

\begin{figure}[t]
\begin{center}
 \includegraphics[width=0.32\columnwidth]{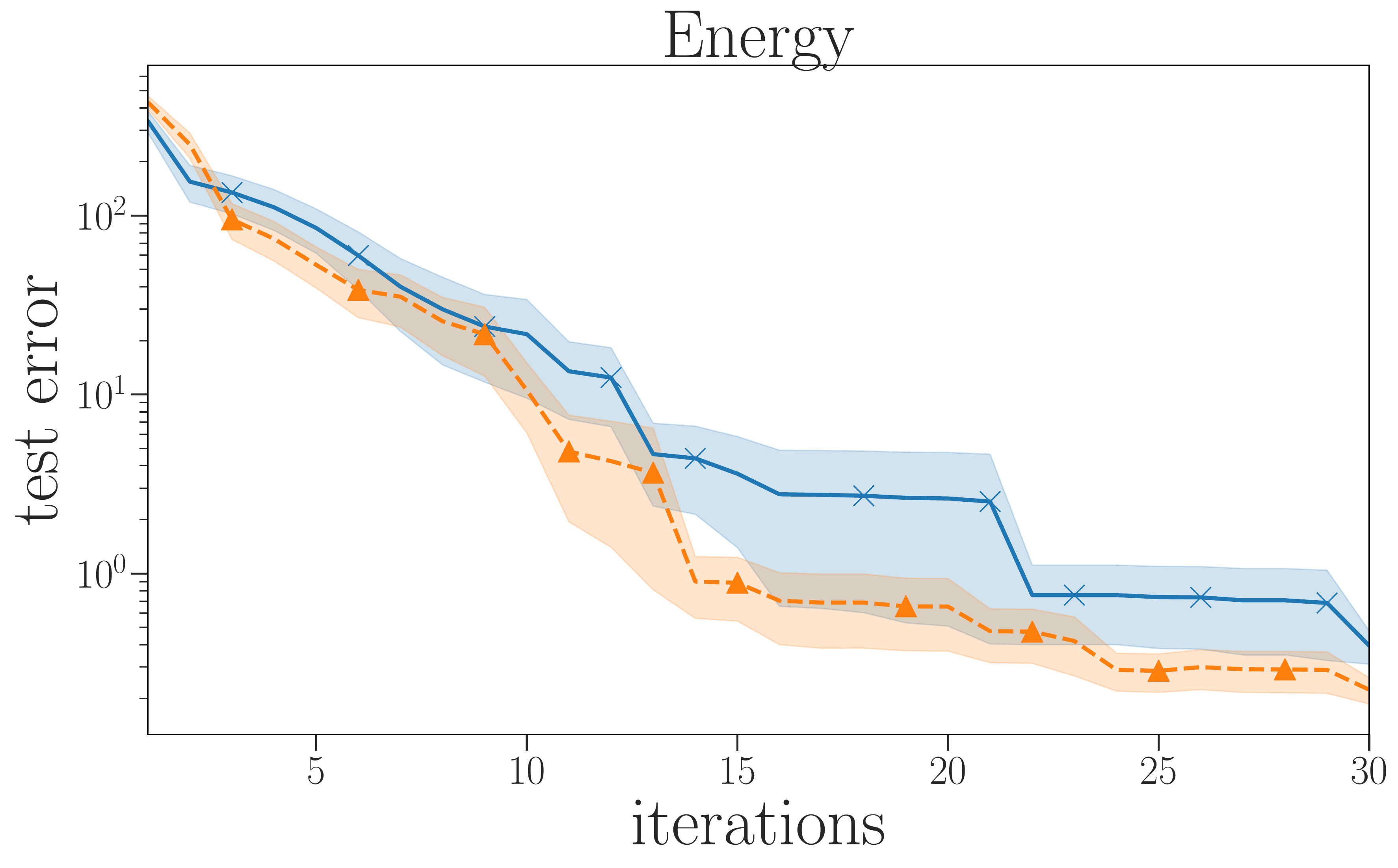}
 \includegraphics[width=0.32\columnwidth]{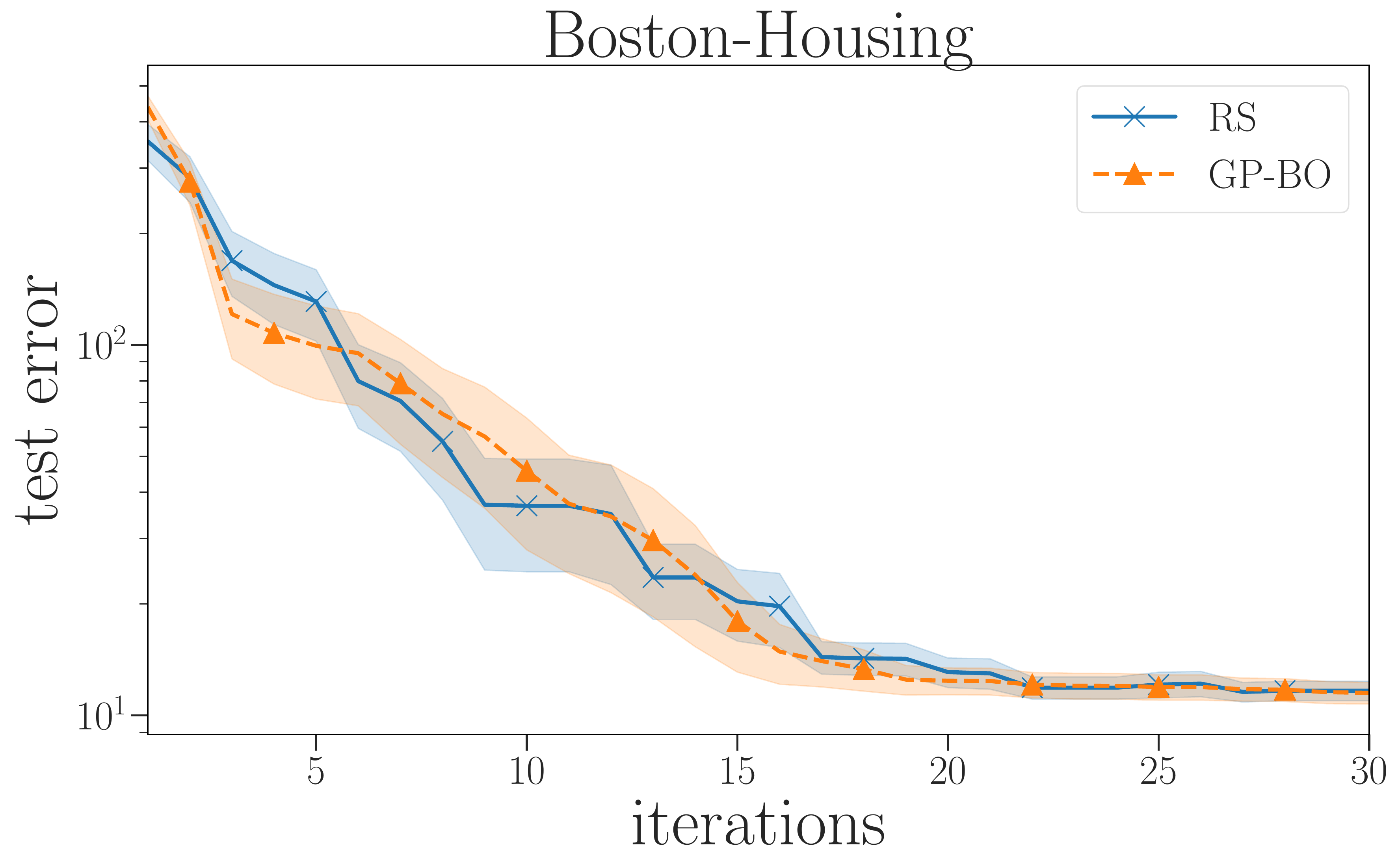}
 \includegraphics[width=0.32\columnwidth]{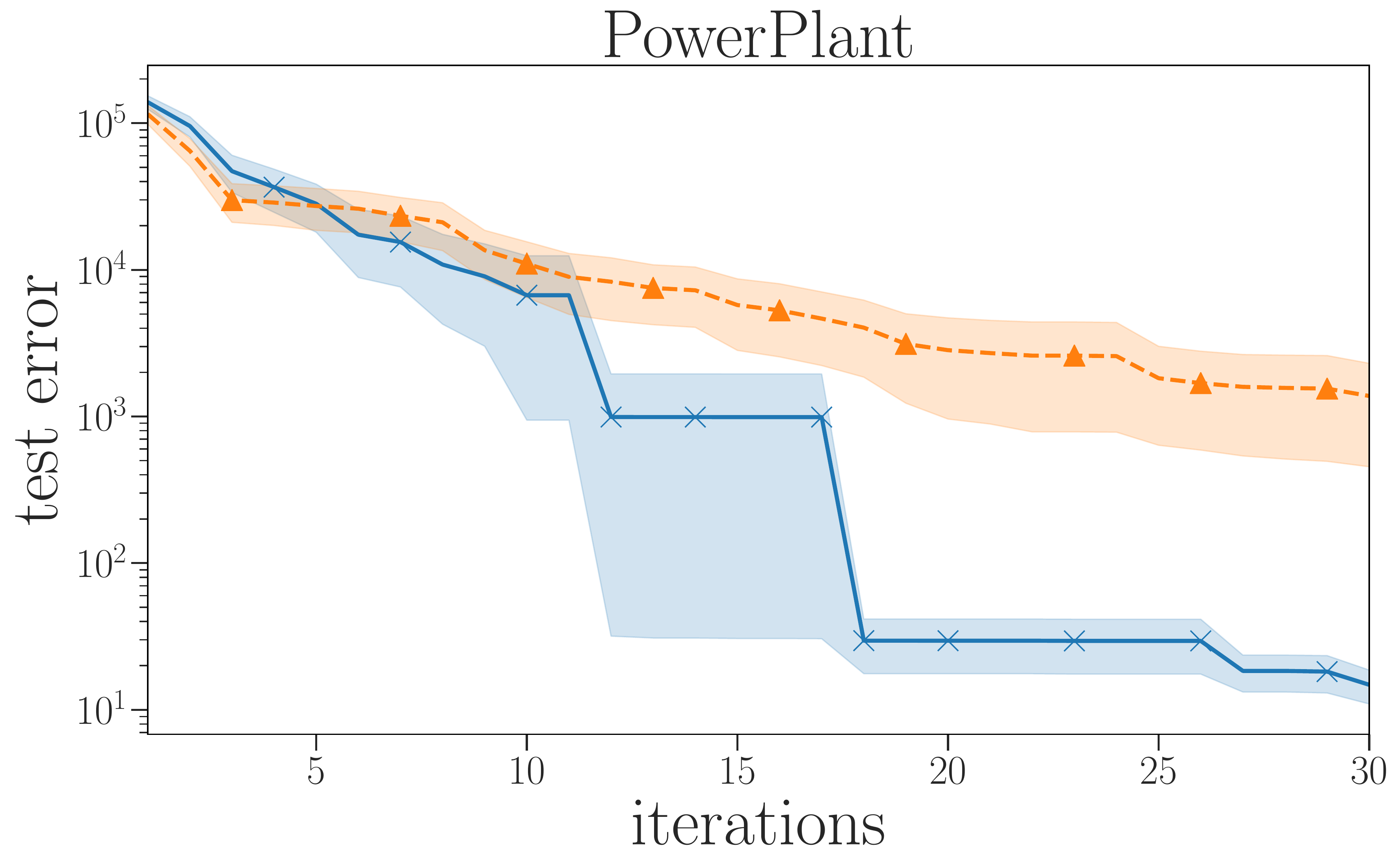}
 \caption{Common pitfalls in the evaluation of \hpo methods: we compare two different \hpo methods for optimizing the hyperparameters of XGBoost on three UCI regression datasets (see Appendix B for more datasets). The small number of tasks makes it hard to draw any conclusions, since the ranking of the methods varies between tasks. Furthermore, a full run might take several hours which makes it prohibitively expensive to average across a large number of runs.}
 \label{fig:teaser1}
 \vspace{-8mm}
\end{center}
\end{figure}

Due to these two problems researchers can only carry out a limit number of comparisons within a reasonable computational budget.
This delays the progress of the field as statistically significant conclusions about the performance of different \hpo methods may not be possible to draw.
See Figure \ref{fig:teaser1} for an illustrative experiment of the \hpo of XGBoost~\citep{chen-kdd16}.
It is well known that Bayesian optimization with Gaussian processes (BO-GP)~\citep{shahriari-ieee16a} outperforms naive random search (RS) in terms of number of function evaluations on most \hpo problems.
While we show clear evidence for this in Appendix B on a larger set of datasets, this conclusion cannot be reached when optimizing on the three "unlucky" picked datasets in Figure~\ref{fig:teaser1}.
Surprisingly, the community has not paid much attention to this issue of proper benchmarking, which is a key step required to generate new scientific knowledge.

\begin{wrapfigure}[23]{r}{0.48\linewidth}
 \vspace{-6mm}
 \includegraphics[width=\linewidth]{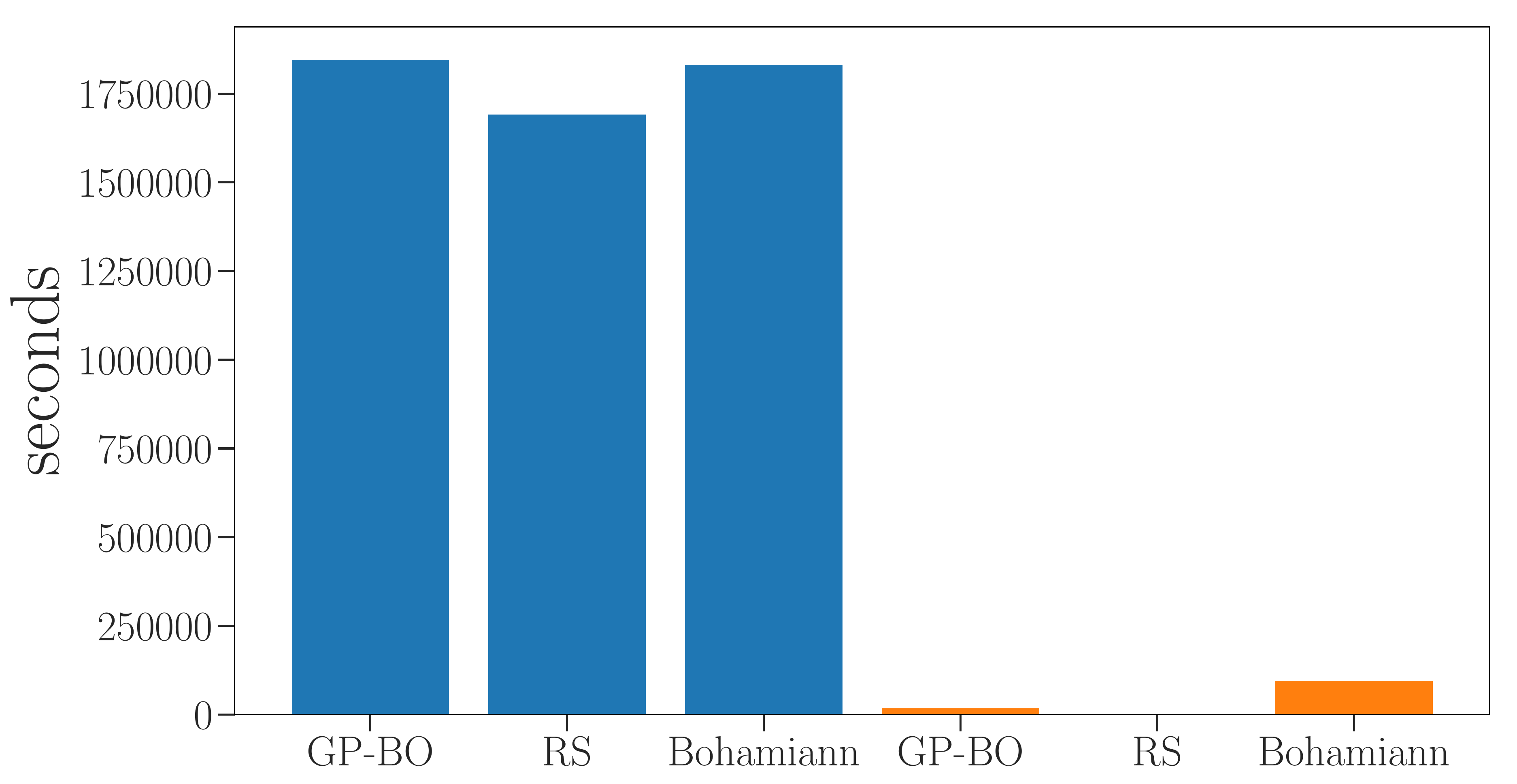}
 \caption{The three blue bars on the left show the total wall-clock time of executing 20 independent runs of GP-BO, RS  and Bohamiann (see Section~\ref{sec:experiments_profet}) with 100 function evaluations for the \hpo of a feed forward neural network on MNIST. The orange bars show the same for optimizing a task sampled from our proposed meta-model, where benchmarking  is orders of magnitude cheaper in terms of wall-clock time than the original benchmarks, thereby the computational time is almost exclusively spend for the optimizer (hence the larger bars for GP-BO and Bohamiann compared to RS).}
 \label{fig:teaser2}
\end{wrapfigure}

In this work we present a generative meta-model that, conditioned on off-line generated data, allows to sample an unlimited number of new tasks that share properties with the original ones.
There are several advantages to this approach. 
First, the new problem instances are inexpensive to evaluate as they are generated with a parameteric form, which drastically reduces the resources needed to compare \hpo methods, bounded only by the optimizer's computational overhead (see Figure \ref{fig:teaser2} for an example).
Second, there is no limit in the number of tasks that can be generated, which helps to draw statistically more reliable conclusions.
Third, the shape and properties of the tasks are not predefined but learned using a few real tasks of an \hpo problem.
While the \emph{global} properties of the initial tasks are preserved in the samples, the generative model allows the exploration of instances with diverse \emph{local} properties making comparisons more robust and reliable (see Appendix D for some example tasks).

In light of the recent call for more reproducibility, we are convinced that our meta-surrogate benchmarks enable more reproducible research in AutoML:
First of all, these cheap-to-evaluate surrogate benchmarks allows researches to reproduce experiments or perform many repeats of their own experiments without relying on tremendous computational resources.
Second, based on our-proposed method, we provide a more thorough benchmarking protocol that reduces the risk of extensively tuning an optimization method on single tasks.
Third, surrogate benchmarks in general are less dependent on hardware and technical details, such as complicated training routines or preprocessing strategies.

\section{Related Work}

The use of meta-models that learn across tasks has been investigated by others before.
To warm-start \hpo on new tasks from previously optimized tasks,~\citet{swersky-nips13a} extended Bayesian optimization to the multi-task setting by using a Gaussian process that also takes the correlation between tasks into account.
Instead of a Gaussian process, \citet{springenberg-nips16} used a Bayesian neural network inside multi-task Bayesian optimization which learns an embedding of tasks during optimization.
Similarly, \citet{perrone-nips18} used Bayesian linear regression, where the basis functions are learned by a neural network, to warm-start the optimization from previous tasks.
\citet{feurer-aaai15} used a set of dataset statistics as meta-features to measure the similarity between tasks, such that hyperparameter configurations that were superior on previously optimized similar tasks can be evaluated during the initial design.
This technique is also applied inside the auto-sklearn framework~\citep{feurer-nips2015}.
In a similar vein, \citet{fusi-nips18} proposed to use a probabilistic matrix factorization approach to exploit knowledge gathered on previously seen tasks.
\citet{vanrijn-kdd18} evaluated random hyperparameter configurations on a large range of tasks to learn priors for the hyperparameters of support vector machines, random forests and Adaboosts.
The idea of using a latent variable to represent correlation among multiple outputs of a Gaussian process has been exploited by~\citet{dai-nips17}.

Besides the aforementioned work on BO, also other methods have been proposed for efficient HPO.
\citet{li-iclr17} proposed Hyperband, which based on the bandit strategy successive halving~\citep{jamieson-aistats16}, dynamically allocates resources across a set of random hyperparameter configurations.
Similarly, \citet{jadeberg-arxiv17} presented an evolutionary algorithm, dubbed PBT, which adapts a population of hyperparameter configurations during training by either random perturbations or exploiting values of well-performing configurations in the population.

In the context of benchmarking \hpo methods, HPOlib~\citep{eggensperger-bayesopt13} is a benchmarking library that provides a fixed and rather small set of common HPO problems. 
In earlier work, \citet{eggensperger-aaai15} proposed surrogates to speed up the empirical benchmarking of \hpo methods.
Similar to our work, these surrogates are trained on data generated in an off-line step.
Afterwards, evaluating the objective function only requires querying the surrogate model instead of actually running the benchmark. However, these surrogates only mimic one particular task and do not allow for generating new tasks as presented in this work.
Recently, tabular benchmarks were introduced for neural architecture search~\citep{ying-arxiv19} and hyperparameter optimization~\citep{klein-arxiv19a}, which first perform an exhaustive search of a discrete benchmark problem to store all results in a database and then replace expensive function evaluations by efficient table lookups. 
While this does not introduce any bias due to a model (see Section~\ref{sec:discussion} for a more detailed discussion), tabular benchmarks are only applicable for problems with few, discrete hyperparameters.
Related to our work, but for benchmarking general blackbox optimization methods, is the \coco platform~\citep{hansen-arxiv16}. However, compared to our approach, it is based on handcrafted synthetic functions that do not resemble real world \hpo problems.

\section{Benchmarking \hpo methods with generative models}\label{sec:method}

We now describe the generative meta-model to create \hpo tasks.
First we give a formal definition of benchmarking \hpo methods across tasks sampled from a unknown distribution and then describe how we can approximate this distribution by our new proposed meta-model.

\subsection{Problem Definition}

We denote $t_1,\dots,t_M$ to be a set of related objectives/tasks with the same input domain $\mathcal{X}$, for example $ \mathcal{X} \subset \mathbb{R}^d $.
We assume that each $t_i$ for $i=1,\dots M$, is an instantiation of an unknown distribution of tasks $t_i \sim p(t)$. 
Every task $t$ has an associated objective function $f_t: \mathcal{X} \rightarrow \mathbb{R}$ where $\vx \in \mathcal{X}$ represents a hyperparameter configuration and we assume that we can observe $f_t$ only through noise: $y_t \sim \gauss(f_t(\vx),\sigma_t^2)$.

Let us denote by $r(\alpha,t)$ the performance of an optimization method $\alpha$ on a task $t$; for instance, a common example for $r$ is the regret of the best observed solution (called incumbent).
To compare two different methods $\alpha_A$ and $\alpha_B$, the standard practice is to compare $r(\alpha_A, t_i)$ with $r(\alpha_B, t_i)$ on a set of hand-picked tasks $t_i \in \{t_1, \dots t_M \}$. 
However, to draw statistically more significant conclusion, we would ideally like to integrate over all tasks: 
\begin{equation}\label{eq:score_metric}
  S_{p(t)}(\alpha) =  \int r(\alpha,t)p(t) dt.
\end{equation}
Unfortunately, the above integral is intractable as $p(t)$ is unknown.
The main contribution of this paper is to approximate $p(t)$ with a generative meta-model $\hat{p}(t \mid\mathcal{D})$ based on some off-line generated data $\mathcal{D}= \big \{ \{(\vx_{tn}, y_{tn})\}_{n=1}^{N} \big\}_{t=1}^{T}$.
This enables us to sample an arbitrary amount of tasks $t_i \sim \hat{p}(t \mid \mathcal{D})$ in order to perform a Monte-Carlo approximation of Equation~\ref{eq:score_metric}.

\subsection{Meta-Model for Task Generation}\label{sec:model}

In order to reason across tasks, we define a probabilistic encoder $p(\vh_t \mid \mathcal{D})$ that learns a latent representation $\vh_t \in \mathbb{R}^Q$ of a task $t$.

More precisely, we use Bayesian \gplvm \citep{titsias-aistats10} which assumes that the target values that belong to the task $t$, stacked into a vector $\vy_t = (y_{t1}, \dots,  y_{tN})$ follow the generative process:
\begin{equation}
\vy_t = g(\vh_t) + \epsilon,\quad g \sim \mathcal{GP}(0, k),\quad \epsilon \sim \mathcal{N}(0, \sigma^2),
\end{equation}
where $k$ is the covariance function of the \gp.
By assuming that the latent variable $\vh_t$ has an uninformative prior $\vh_t \sim \mathcal{N}(0, \I)$, the latent embedding of each task is inferred as the posterior distribution $p(\vh_t \mid \mathcal{D})$.
The exact formulation of the posterior distribution is intractable, but following the variational inference presented in \citet{titsias-aistats10}, we can estimate a variational posterior distribution $q(\vh_t) = \mathcal{N}(m_t, \Sigma_t) \approx p(\vh_t \mid \mathcal{D})$ for each task $t$.

Similar to Multi-Task Bayesian Optimization~\citep{swersky-nips13a,springenberg-nips16}, we define a probabilistic model for the objective function $p(y_{t} \mid \vx, \vh_t)$ across tasks which gets as an additional input a task embedding based on our independently trained probabilistic encoder.
Following \citet{springenberg-nips16}, we use a Bayesian neural network with $M$ weight vectors $\{ \vtheta_1, \dots, \vtheta_M \}$ to model
\begin{equation}\label{eq:model}
  \begin{aligned}
    p(y_{t} \mid \vx, \vh_t, \mathcal{D}) &= \int  p(y_{t} \mid \vx, \vh_t, \vtheta) p(\theta \mid \mathcal{D}) d\vtheta    &\approx \frac{1}{M} \sum^{M}_{i=1} p(y_t \mid \vx, \vh_t, \vtheta_i).
  \end{aligned} 
\end{equation}
where $\vtheta_i \sim p(\vtheta \mid \mathcal{D})$ is sampled from the posterior of the neural network weights. 

By approximating $p(y_{t} \mid \vx, \vh_t) = \gauss \big(\mu(\vx, \vh_t), \sigma^2(\vx, \vh_t) \big) $ to be Gaussian \citep{springenberg-nips16}, we can compute the predictive mean and variance by:
\begin{equation*}
\begin{aligned}
\mu(\vx, \vh_t) &= \frac{1}{M} \sum_{i=1}^M \hat{\mu}(\vx, \vh_t \mid \vtheta_i) && ; && 
\sigma^2(\vx, \vh_t) &= \frac{1}{M} \sum_{i=1}^M \big( \hat{\mu}(\vx, \vh_t \mid \vtheta_i) - \mu(\vx, \vh_t) \big)^2 + \hat{\sigma}^2_{\vtheta_i},
\end{aligned}
\end{equation*}

where $\hat{\mu}(\vx, \vh_t \mid \vtheta_i)$ and $\hat{\sigma}^2_{\vtheta_i}$ are the output of a single neural network with parameters $\vtheta_i$\footnote{Note that we model an homoscedastic noise, because of that, $\hat{\sigma}^2_{\vtheta_i}$ does not depend on the input}.
To get a set of weights $\{ \vtheta_1, \dots, \vtheta_M \}$, we use stochastic gradient Hamiltonian Monte-Carlo \citep{chen-icml14} to sample $\vtheta_i \sim p(\vtheta, \mathcal{D})$ from:
\begin{equation*}
  p(\vtheta, \mathcal{D})  = \frac{1}{N} \sum_{n=1}^{N} \frac{1}{H} \sum_{j=1}^{H} \log p(y_n \mid \vx_n, \vh_{nj})
\end{equation*}
with $N = |\mathcal{D}|$ the number of datapoints in our training set and $H$ the number of samples we draw from the latent space $\vh_{tj} \sim q(\vh_t)$.

\subsection{Sampling New Tasks}

In order to generate a new task $t_{\star} \sim \hat{p}(t \mid\mathcal{D})$, we need the associated objective function $f_{t_{\star}}$ in a parameteric form such that we can evaluate it later on any $\vx \in \mathcal{X}$.

Given the meta-model above, we perform the following steps: (i) we sample a new latent task vector $\vh_{t_{\star}} \sim q(\vh_t)$; (ii) given $\vh_{t_{\star}} $ we pick a random $\vtheta_i$ from the set of weights $\{ \vtheta_1, \dots \vtheta_M \}$ of our Bayesian neural network and set the new task to be $f_{t_{\star}}(\vx) = \hat{\mu}(\vx, \vh_{t_{\star}} \mid \vtheta_i)$.

Note that using $f_{t_{\star}}(\vx)$ makes our new task unrealisticly smooth. Instead, we can emulate the typical noise appearing in \hpo benchmarks by returning $y_{t_{\star}}(\vx) \sim \gauss \big(\hat{\mu}(\vx, \vh_{t_{\star}} \mid \vtheta_i), \hat{\sigma}^2_{\vtheta_i} \big)$, which can be done at an insignificant cost.

%
%
%

\section{Profet}\label{sec:profet_benchmarking}

\begin{figure*}[t!]
 \centering
 \includegraphics[width=.42\textwidth]{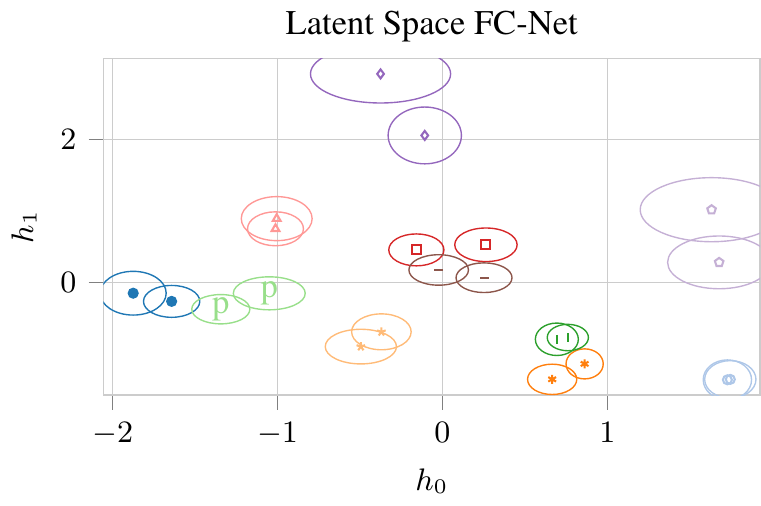}
 \includegraphics[width=.53\textwidth]{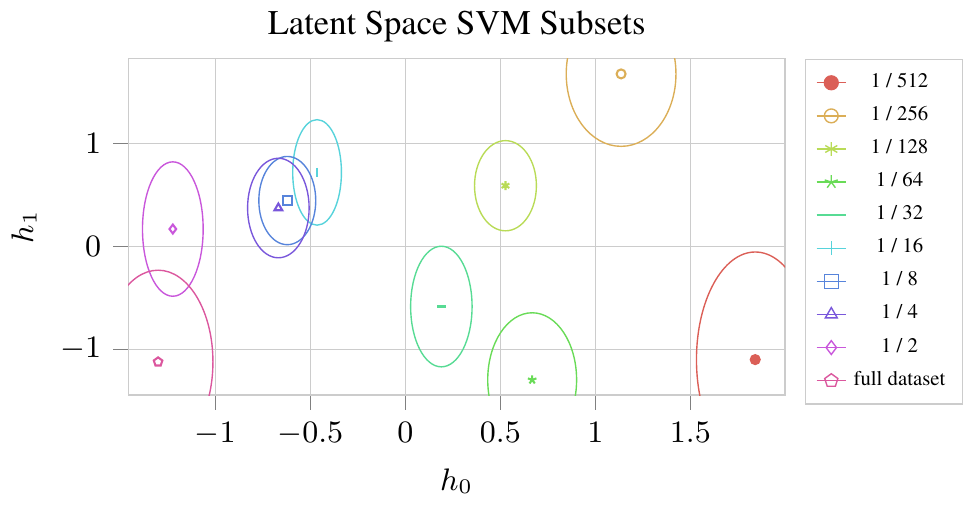}
 \caption[Latent representation]{Latent space representations of our probabilistic encoder. \emph{Left}: Representation of task pairs (same color)  generated by partitioning eleven datasets from the fully connected network benchmark detailed in Section~\ref{sec:data_collection}.  Mean of tasks are visualized with different markers and ellipses represent 4 standard deviations. \emph{Right}: Latent space learned for a model where the input tasks are generated by training a SVM on subsets of MNIST (see \citet{klein-ejs17} for more details).}
\label{fig:representation}
 \vspace{-4mm}
\end{figure*}

We now present our PRObabilistic data-eFficient Experimentation Tool, called \profet, a benchmarking suite for \hpo methods (an open-source implementation is available here:
\href{https://github.com/amzn/emukit}{https://github.com/amzn/emukit}).
We provide pseudo code in Appendix G.
The following section describes first how we collected the data to train our meta-model based on three typical \hpo problems classes.
We then explain how we generated $T=1000$ different tasks for each problem class from our meta-model.
As described above, we provide a noisy and noiseless version of each task.
Last, we discuss two ways that are commonly used in the literature to assess and aggregate the performance of \hpo methods across tasks.

\subsection{Data Collection}\label{sec:data_collection}

We consider three different \hpo problems, two for classification and one for regression, with varying dimensions $D$.
For classification, we considered a support vector machine (SVM) with $D=2$ hyperparameters and a feed forward neural network (FC-Net) with $D=6$ hyperparameters on 16 OpenML~\citep{vanschoren-sigkdd13a} tasks each.
We used gradient boosting (XGBoost)\footnote{We used the implementation from \citet{chen-kdd16}} with $D=8$ hyperparameters for regression on 11 different UCI datasets~\citep{lichman-13}.
For further details about the datasets and the configuration spaces see Appendix A.
To make sure that our meta-model learns a descriptive representation we need a solid coverage over the whole input space.
For that we drew $100 D$ pseudo randomly generated configurations from a Sobol grid~\citep{Sobol1967}.

Details of our meta-model are described in Appendix F.
We show some qualitative examples of our probabilistic encoder in Section~\ref{sec:quality}. 
We can also apply the same machinery to model the cost in terms of computation time for evaluating a hyperparameter configuration to use time rather than function evaluations as budget.
This enables future work to benchmark \hpo methods that explicitly take the cost into account (\eg{} EIperSec \citep{snoek-nips12a}).

\subsection{Performance Assessment}\label{sec:performance}

To assess the performance of a \hpo method aggregated over tasks, we consider two different ways commonly used in the literature.
First, we measure the \textbf{runtime} $r(\alpha, t, y_{target})$ that a \hpo method $\alpha$ needs to find a configuration that achieves a performance that is equal or lower than a certain target value $y_{target}$ on task $t$ \citep{hansen-arxiv16b}.
Here we define runtime either in terms of function evaluations or estimated wall-clock time predicted by our meta-model.
Using a fixed target approach allows us to make quantitative statements, such as: \textit{method A is, on average, twice as fast than method B}. See~\citet{hansen-arxiv16b} for a more detailed discussion.
We average across target values with a different complexity by evaluating the Sobol grid from above on each generated task.
We use the corresponding function values as targets, which, with the same argument as described in Section~\ref{sec:data_collection}, provides a good coverage of the error surface.
To aggregate the runtime we use the empirical cumulative distribution function (ECDF)~\citep{more-siam09}, which, intuitively, shows for each budget on the x-axis the fraction of solved tasks and target pairs on the y-axis (see Figure~\ref{fig:comparison_svm_noiseless} left for an example).

Another common way to compare different \hpo methods is to compute the \textbf{average ranking score} in every iteration and for every task~\citep{bardenet-icml13a}.
We follow the procedure described by~\citet{feurer-aaai15} and compute the average ranking score as follows: assuming we run $K$ different \hpo methods $M$ times for each task, we draw a bootstrap sample of 1000 runs out of the $K^M$ possible combinations.
For each of these samples, we compute the average fractional ranking (ties are broken by the average of the ordinal ranks) after each iteration.
At the end, all the assigned ranks are further averaged over all tasks.
Note that, averaged ranks are a relative performance measurement and can worsen for one method if another method improves (see Figure~\ref{fig:comparison_svm_noiseless} right for an example). 

\section{Experiments}\label{sec:experiments_profet}

In this section we present: (i) some qualitative insights of our meta-model by showing how it is able to coherently represent a sets of tasks in its latent space, (ii) an illustration of why \profet{} helps to obtain statistically meaningful results and (iii) a comparison of various methods from the literature on our new benchmark suite. In particular, we show results for the following state-of-the-art BO methods as well as two popular evolutionary algorithms: 
\begin{itemize}
  \item BO with Gaussian processes (BO-GP)~\citep{jones-jgo98a}. We used expected improvement as acquisition function and marginalize over the Gaussian process' hyperparameters as described by \citet{snoek-nips12a}.
 \item SMAC~\citep{hutter-lion11a}: which is a variant of BO that uses random forests to model the objective function and stochastic local search to optimize expected improvement.\\ We use the implementation from \href{https://github.com/automl/SMAC3}{https://github.com/automl/SMAC3}.
 \item The BO method TPE by \citet{bergstra-nips11a} which models the density of good and bad configurations in the input space with a kernel density estimators. We used the implementation provided from the Hyperopt package~\citep{komer-automl14a}
 \item BO with Bayesian neural networks (BOHAMIANN) as described by~\citet{springenberg-nips16}. To avoid introducing any bias, we used a different architecture with less parameters (3 layers, 50 units in each) than we used for our meta-model (see Section \ref{sec:method}).
 \item Differential Evolution (DE)~\citep{storn-jgo97} (we used our own implementation) with rand1 strategy for the mutation operators and a population size of 10.
 \item Covariance Matrix Adaption Evolution Strategy (CMA-ES) by \citet{hansen-eda06} where we used the implementation from \href{https://github.com/CMA-ES/pycma}{https://github.com/CMA-ES/pycma}
 \item Random Search (RS) \citep{bergstra-jmlr12a} which samples configurations uniformly at random.
\end{itemize}
For BO-GP, BOHAMIANN and RS we used the implementation provided by the RoBO package~\citep{klein-bayesopt17}.
We provide more details for every method in Appendix E.

\subsection{Tasks Representation in the Latent Space}\label{sec:quality}

We demonstrate the interpretability of the learned latent representations of tasks in two examples.
For the first experiment we used the fully connected network benchmark described in Section~\ref{sec:data_collection}. To visualize that our meta-model learns a meaningful latent space, we doubled 11 out of the 18 original tasks to train the model by splitting each one of them randomly in two of the same size.  Thereby, we guarantee that there are pairs of tasks that are similar to each other. In Figure \ref{fig:representation} (left), each color represents the partition of the original task and each ellipse represents the mean and four times the standard deviation of the latent task representations. One can see that the closest neighbour of each task is the other task that belongs to the same original task.

The second experiment targets multi-fidelity problems that arise when training a machine learning model on large datasets and approximate versions of the target objective are generated by considering subsamples of different sizes. We used the SVM surrogate for different dataset subsets from~\citet{klein-ejs17}, which consists of a random forest trained on a grid of hyperparameter configurations of a SVM evaluated on different subsets of the training data. In particular, we defined the following subsets: $\{\nicefrac{1}{512}, \nicefrac{1}{256},\nicefrac{1}{128},\nicefrac{1}{64},\nicefrac{1}{32},\nicefrac{1}{16},\nicefrac{1}{8},\nicefrac{1}{4},\nicefrac{1}{2},1\}$ as tasks and sampled 100 configurations per task to train our meta-model.
Note that we only provide the observed targets and not the subset size to our model.
Figure~\ref{fig:representation} (right) shows the latent space of the trained meta-model: the latent representation of the model captures that similar data subsets are also close in the latent space. In particular, the first latent dimension $h_0$ coherently captures the sample size, which is learned using exclusively the correlation between the datasets and with no further information about their size.

\subsection{Benchmarking with \profet}

\begin{figure}[t!]
\centering
\includegraphics[width=0.32\textwidth]{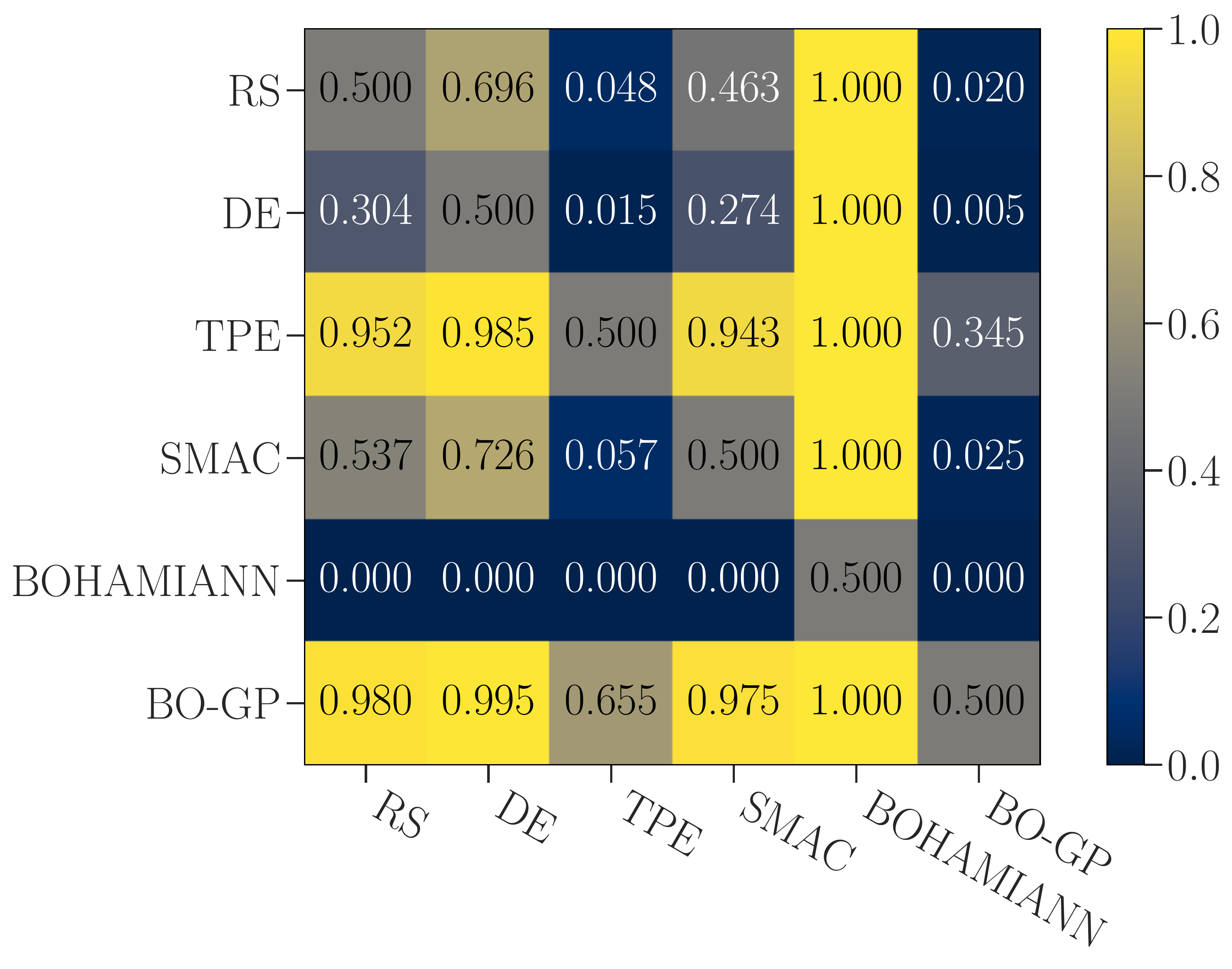}
\includegraphics[width=0.32\textwidth]{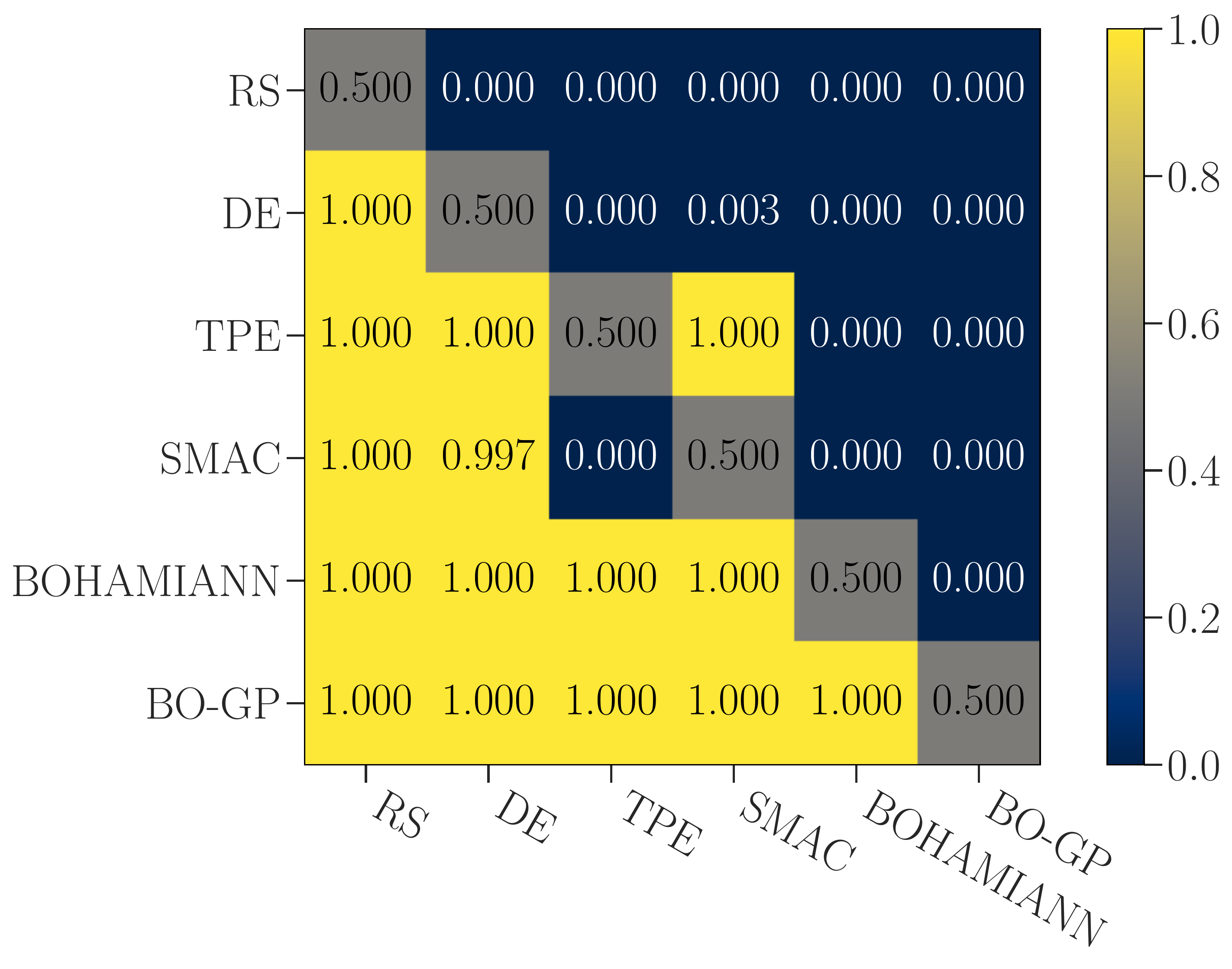}
\includegraphics[width=0.32\textwidth]{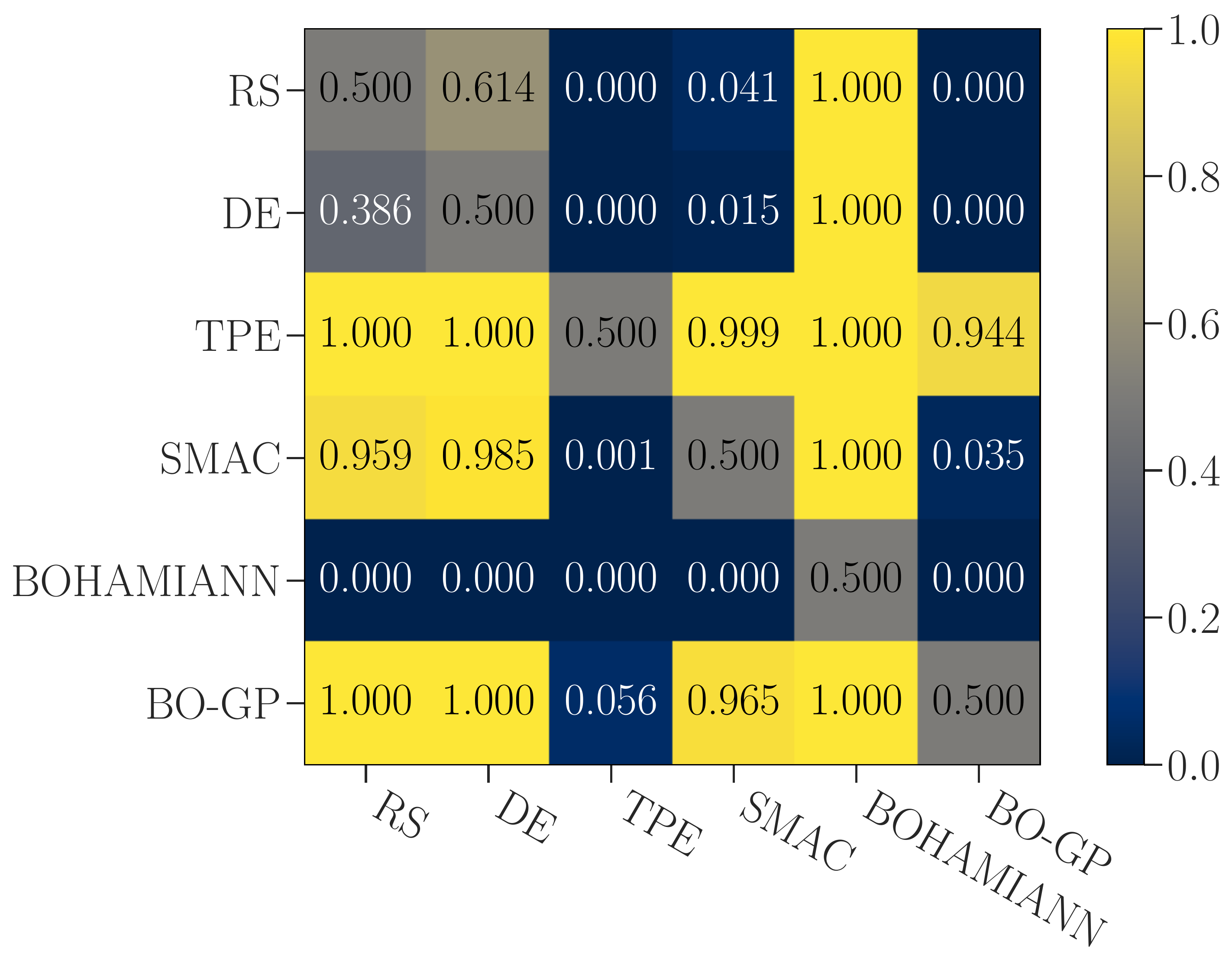}
\caption[Comparison Forrester function]{Heatmaps of the p-values of the pairwise Mann-Whitney U test on three scenarios. Small p-values should be interpreted as finding evidence that the method in the column outperforms the method in the row. Using tasks from our meta-model lead to results that are close to using the large set of original tasks. \emph{Left}: results with 1000 real tasks. \emph{Middle}: subset of only 9 reals tasks. \emph{Right:} results with 1000 tasks generated from our meta-model.}
\label{fig:forrester_1}
\vspace{-4mm}
\end{figure}

Comparing \hpo methods using a small number of instances affects our ability to properly perform statistical tests.
To illustrate this we consider a distribution of tasks that are variations of the Forrester function $ f(x) = ((\alpha x - 2)^2)\sin(\beta x - 4)$.
We generated 1000 tasks by uniformly sampling random $\alpha$ and $\beta$ in $[0,1]$ and compared six \hpo methods: RS, DE, TPE, SMAC, BOHAMIANN and BO-GP (we left CMA-ES out because the python version does not support 1-dimensional optimization problems).

Figure~\ref{fig:forrester_1} (left) shows the p-values of all pairwise comparisons with the null hypothesis `$\text{Method}_{column}$ achieves a higher error after 50 function evaluations averaged over 20 runs than $\text{Method}_{row}$' for the Mann-Whitney U test.
Squares in the figure with a p-value smaller than $0.05$ are comparisons in which with a 95\% confidence we have evidence to show that the method in the column is better that the method in the row (we have evidence to reject the null hypothesis).
To reproduce a realistic setting where one has access to only a small set of tasks, we picked 9 out of the 1000 tasks randomly.
Now, in order to acquire a comparable number of samples to perform a statistical test, we performed 2220 runs of each method on every task, and then computed the average of groups of 20 runs, such that we obtained 999 samples per method to compute the statistical test.
One can see in Figure~\ref{fig:forrester_1} (middle), that although the results are statistically significant, they are misleading: for example, BOHAMIANN is dominating all other methods (except BO-GP), whereas it is significantly worse than all other methods if we consider all 1000 tasks.

To solve this issue and obtain more information from the same limited number of a subset of 9 tasks, we use \profet.
We first train the meta-model on the same 9 selected tasks and then use it to generate 1000 new surrogate tasks (see Appendix C for a visualization).
Next, we use these tasks to run the comparison of the  \hpo methods.
Results are shown in Figure~\ref{fig:forrester_1} (right). The heatmap of statistical comparisons reaches very similar conclusions to those obtained with the original 1000 tasks, contrary to what happened when we did the comparisons with 9 tasks only (\ie{} p-values are closer to the original ones).
We conclude that our meta-model captures the variability across tasks such that using samples from it (generated based on a subset of tasks) allows us to draw conclusion that are more in line with experiments on the full dataset of tasks than running directly on the subset of tasks.

\subsection{Comparing State-of-the-art \hpo Methods}\label{sec:comparison_profet}

\begin{figure*}[t]
\centering
 \includegraphics[width=0.49\textwidth]{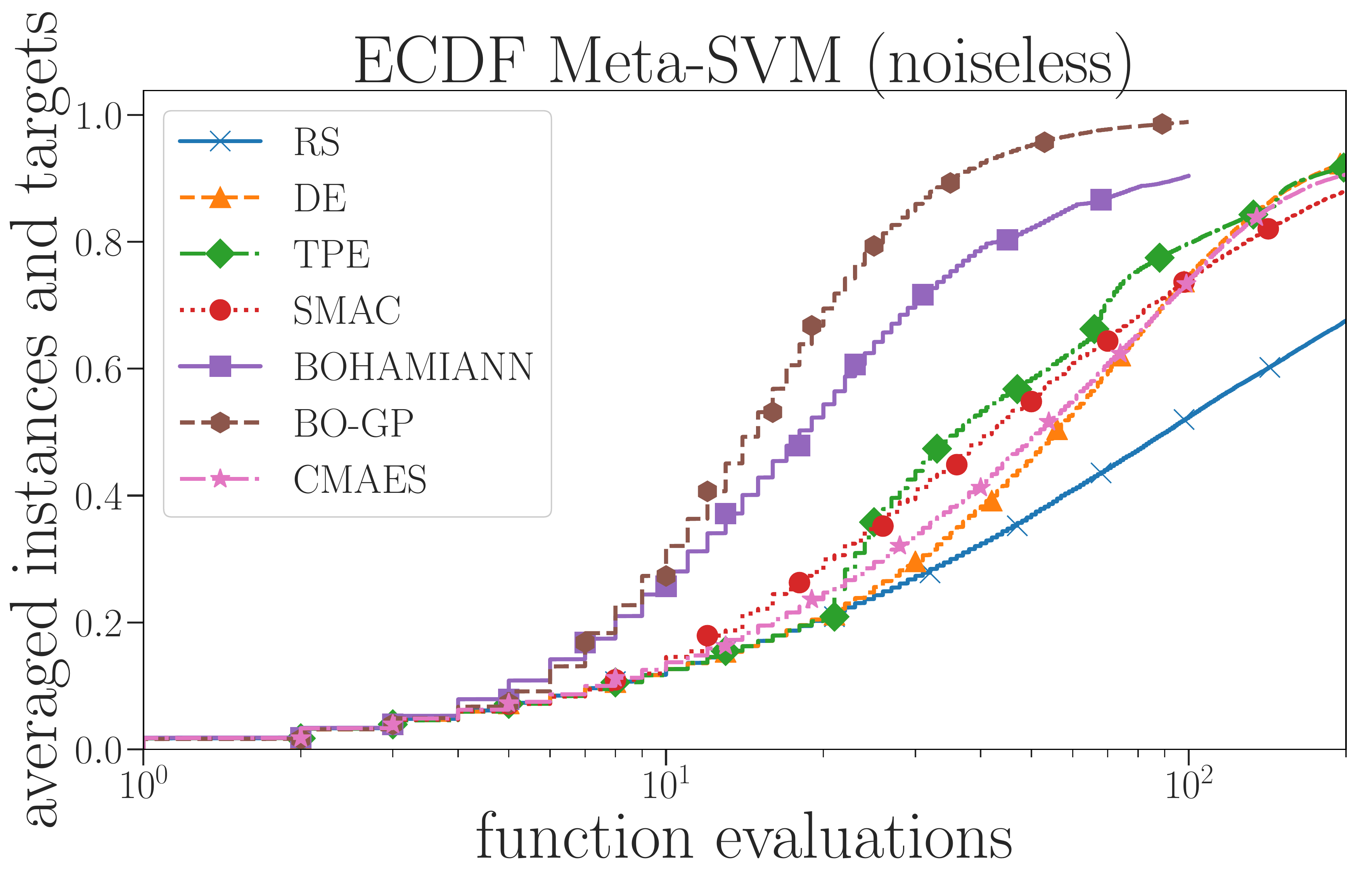}
 \includegraphics[width=0.49\textwidth]{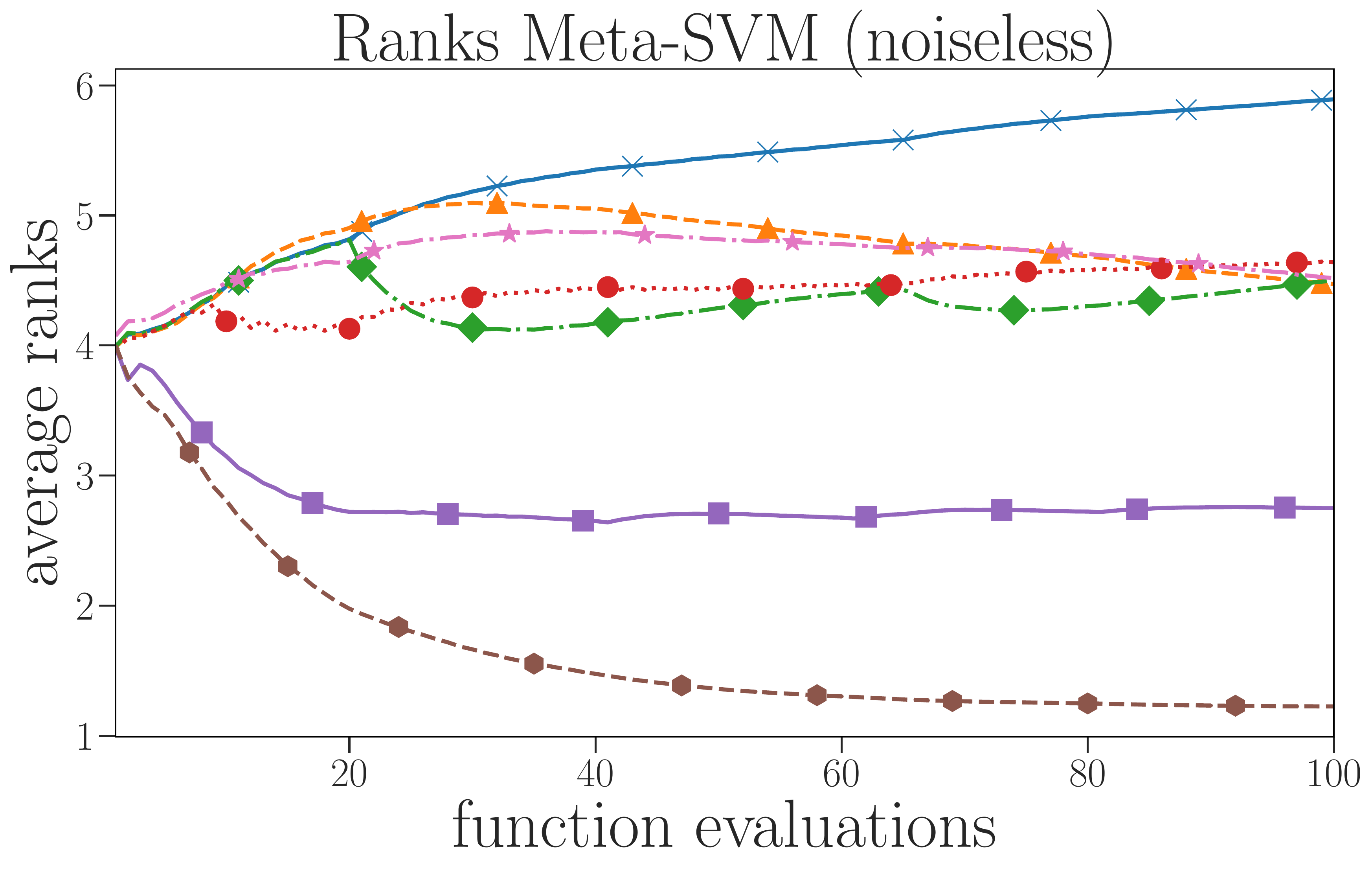}
 \caption[Comparison SVM noiseless]{Comparison of various \hpo methods on 1000 tasks of the noiseless SVM benchmark:  \emph{left:} ECDF for the runtime \emph{right}: average ranks. See Appendix D for the results of all benchmarks.}
 \vspace{-6mm}
 \label{fig:comparison_svm_noiseless}
\end{figure*}

We conducted 20 independent runs for each method on every task of all three problem classes described in Section~\ref{sec:data_collection} with different random seeds.
Each method had a budget of 200 function evaluations per task, except for BO-GP and BOHAMIANN, where, due to their computational overhead, we were only able to perform 100 function evaluations.
Note that conducting this kind of comparison on the original benchmarks would have been prohibitively expensive.
In Figure~\ref{fig:comparison_svm_noiseless} we show the ECDF curves and the average ranking for the noiseless version of the SVM benchmark.
The results for all other benchmarks are shown in Appendix E.
We can make the following observations:

\begin{itemize}
  \item Given enough budget, all methods are able to outperform RS. BO approaches can exploit their internal model such that they start to outperform RS earlier than evolutionary algorithms (DE, CMA-ES). Thereby, more sophisticated models, such as Gaussian processes or Bayesian neural networks are more \emph{sample efficient} than somewhat simpler methods, \eg{} random forests or kernel density estimators.
  \item The performance of BO methods that model the objective function (BO-GP, BOHAMIANN, SMAC) instead of just the distribution of the input space (TPE) decays if we evaluate the function through noise. Also evolutionary algorithms seem to struggle with noise.
  \item Standard BO (BO-GP) works superior on these benchmarks, but its performance decays rapidly with the number of dimensions.
  \item Runner-up is BOHAMIANN which works slightly worse than BO-GP but seems to suffer less under noisy function values.  Note that this result can only be achieved by using \profet{} as we could not have evaluated with and without noise on the original datasets.
  \item Given a sufficient budget, DE starts to outperform CMA-ES as well as BO with simpler (and cheaper) models of the objective function (SMAC, TPE), making it a competitive baseline particularly for higher dimensional benchmarks. 
 \end{itemize}

\section{Discussion and future work}\label{sec:discussion}

We presented \profet, a new tool for benchmarking \hpo algorithms.
The key idea is to use a generative meta-model, trained on offline generated data, to produce new tasks, possibly perturbed by noise.
The new tasks retain the properties of the original ones but can be evaluated inexpensively, which represents a major advance to speed up comparisons of \hpo methods.
In a battery of experiments we have illustrated the representation power of \profet{} and its utility when comparing \hpo methods in families of problems where only a few tasks are available.

Besides these strong benefits, there are certain drawbacks of our proposed method:
First, since we encode new tasks based on a machine learning model, our approach is based on the assumptions that come with this model. 
Second, while we show in Section~\ref{sec:experiments_profet} empirical evidence that conclusions based on \profet{} are virtually identical to the ones based on the original tasks, there are no theoretical guarantees that results translate one-to-one to the original benchmarks.
Nevertheless, we believe that \profet{} sets the ground for further research in this direction to provide more realistic use-cases than commonly used synthetic functions, \eg{} Branin, such that future work on \hpo can rapidly perform reliable experiments during development and only execute the final evaluation on expensive real benchmarks.
Ultimately, we think this is an important step towards more reproducibility, which is paramount in such a empirical-driven field as AutoML.

A possible extension of \profet{} would be to consider multi-fidelity benchmarks~\citep{klein-ejs17,kandasamy-icml17,klein-iclr17,li-iclr17} where cheap, but approximate fidelities of the objective function are available, \eg{} learning curves or dataset subsets.
Also, different types of observation noise, e.g  non-stationary or heavy-tailed distributions as well as higher dimensional input spaces with discrete and continuous hyperparameters could be investigated.
Furthermore, since \profet{} also provides gradient information, it could serve as a training distribution for learning-to-learn approaches~\citep{chen-icml17,volpp-arxiv19}. 

\section*{Acknowledgement}

We want to thank Noor Awad for providing an implementation of differential evolution.
\bibliographystyle{plainnat}
\bibliography{lib}

\section{Hyperparameter Optimization Benchmarks} \label{appendix:classification_profet_supp}

\begin{figure}[h]
\centering
 \includegraphics[width=0.32\textwidth]{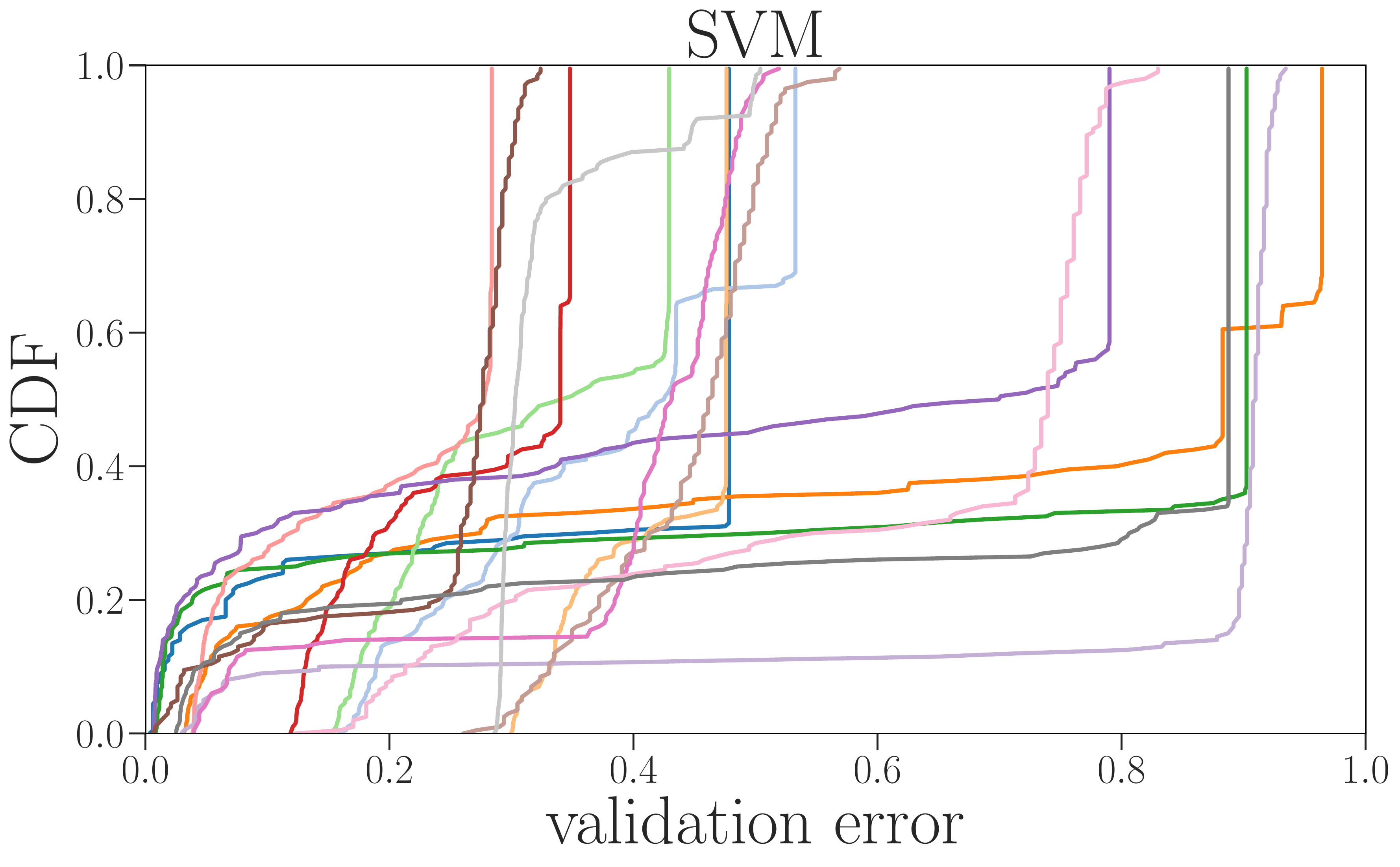}
 \includegraphics[width=0.32\textwidth]{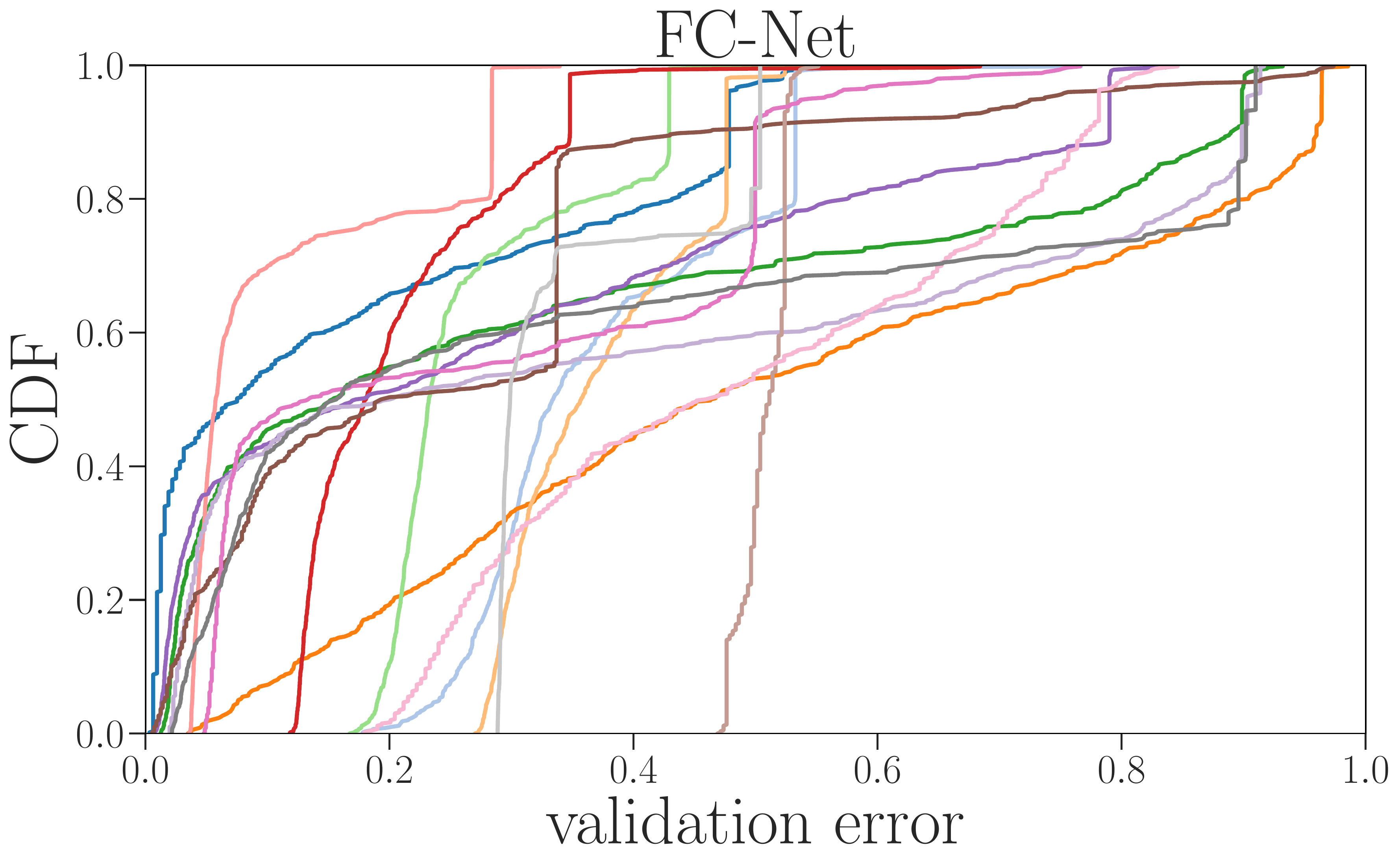}
 \includegraphics[width=0.32\textwidth]{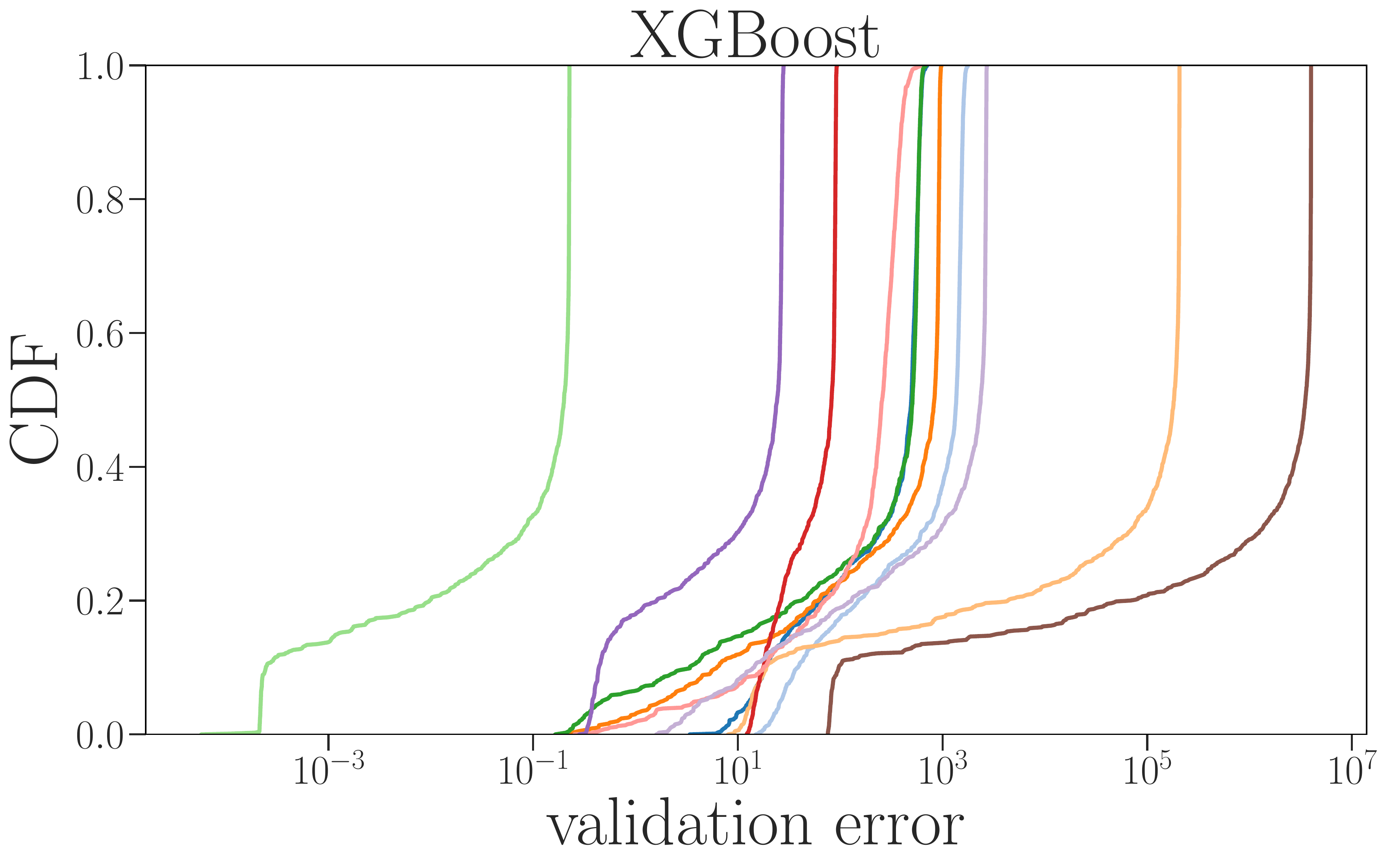}
 \caption[CDF plots meta benchmarks]{The empirical cumulative distribution plots of all observed target values for all tasks.}
 \label{fig:cdf_meta_benchmarks}
\end{figure}

In Table~\ref{tab:openml_profet} we list all OpenML datasets that we used to generate the Meta-SVM and Meta-FCNet benchmarks and in Table~\ref{tab:uci_profet} the UCI datasets that we used for the Meta-XGBoost benchmark.
The ranges of the hyperparameters for all benchmarks are given in Table~\ref{tab:hypers_profet}.
Figure~\ref{fig:cdf_meta_benchmarks} shows the empirical cumulative distribution over the observed target values based on the Sobol grid for all tasks.

\begin{table}[h!]
\centering
\begin{tabular}{|c|c|c|c|}
\toprule
Name & OpenML Task ID & number of features & number of datapoints\\
\midrule
\href{https://www.openml.org/t/3}{kr-vs-kp} & 3 & 37 & 3196\\ 
\href{https://www.openml.org/t/2118}{covertype} & 2118 & 55 & 110393\\ 
\href{https://www.openml.org/t/236}{letter} & 236 & 17 & 20000\\ 
\href{https://www.openml.org/t/75101}{higgs} & 75101 & 29 & 98050\\ 
\href{https://www.openml.org/t/258}{optdigits} & 258 & 65 & 5620\\ 
\href{https://www.openml.org/t/336}{electricity} & 336 & 9 & 45312\\ 
\href{https://www.openml.org/t/75112}{magic telescope} & 75112 & 12 & 19020\\ 
\href{https://www.openml.org/t/146595}{nomao} & 146595 & 119 & 34465\\ 
\href{https://www.openml.org/t/146590}{gas-drift} & 146590 & 129 & 13910\\ 
\href{https://www.openml.org/t/250}{mfeat-pixel} & 250 & 241 & 2000\\ 
\href{https://www.openml.org/t/251}{car} & 251 & 7 & 1728\\ 
\href{https://www.openml.org/t/167079}{churn} & 167079 & 101 & 1212\\ 
\href{https://www.openml.org/t/167202}{dna} & 167202 & 181 & 3186\\ 
\href{https://www.openml.org/t/283}{vehicle small} & 283 & 19 & 846\\ 
\href{https://www.openml.org/t/75191}{vehicle} & 75191 & 101 & 98528\\ 
\href{https://www.openml.org/t/3573}{MNIST} & 3573 & 785 & 50000\\ 
\bottomrule
\end{tabular}
\caption[OpenML datasets for Profet]{OpenML dataset we used for the FC-Net and SVM classification benchmarks}
\label{tab:openml_profet}
\end{table}

\begin{table}[h!]
\centering
\begin{tabular}{|c|c|c|}
\toprule
Name  & number of features & number of datapoints\\
\midrule
boston housing & 13 & 506\\
\href{https://archive.ics.uci.edu/ml/datasets/Concrete+Compressive+Strength}{concrete} & 9 & 1030\\
\href{https://archive.ics.uci.edu/ml/datasets/Parkinsons+Telemonitoring}{parkinsons telemonitoring} & 26 & 5875\\
\href{https://archive.ics.uci.edu/ml/datasets/Combined+Cycle+Power+Plant}{combined cycle power plant} & 4 & 9568\\
\href{https://archive.ics.uci.edu/ml/datasets/Energy+efficiency}{energy} & 8 & 768\\
\href{https://archive.ics.uci.edu/ml/datasets/Condition+Based+Maintenance+of+Naval+Propulsion+Plants#}{naval propulsion} & 16 & 11934 \\
\href{https://archive.ics.uci.edu/ml/datasets/Physicochemical+Properties+of+Protein+Tertiary+Structure}{protein structure} & 9 & 45730\\
\href{https://archive.ics.uci.edu/ml/datasets/Yacht+Hydrodynamics}{yacht-hydrodynamics} & 7 & 308\\
\href{https://archive.ics.uci.edu/ml/datasets/Wine+Quality}{winequality-red} & 12 & 4898\\
\href{https://archive.ics.uci.edu/ml/datasets/Relative+location+of+CT+slices+on+axial+axis}{slice localization} & 386 & 53500\\
\bottomrule
\end{tabular}
\caption[UCI datasets for Profet]{UCI regression dataset we used for the XGBoost benchmark. All dataset can be found at \href{here}{https://archive.ics.uci.edu/ml/datasets.html}}
\label{tab:uci_profet}
\end{table}

\begin{table}[!]
\centering
\begin{tabular}{c|lccccc}
\toprule
& Name & Range & log scale & \\
\midrule
SVM & $C$ & $[e^{-10}, e^{10} ]$ & \checkmark \\
 &  $\gamma$ & $[e^{-10}, e^{10}]$ & \checkmark \\
\midrule
FC-Net & learning rate & $[10^{-6}, 10^{-1}]$ & \checkmark\\
 & batch size & $[8, 128]$ & \checkmark \\
& units layer 1 & $[16, 512]$  & \checkmark \\
& units layer 2 & $[16, 512]$  & \checkmark \\
& drop. rate l1 & $[0.0, 0.99]$  & - \\
& drop. rate l2 & $[0.0, 0.99]$  & - \\

\midrule
XGBoost & learning rate & $[10^{-6}, 10^{-1}]$ & \checkmark\\
 & gamma & $[0, 2]$ & - \\
 & L1 regularization & $[10^{-5}, 10^{3}]$  & \checkmark \\
 & L2 regularization & $[10^{-5}, 10^{3}]$  & \checkmark \\
 & number of estimators & $[10, 500]$  & - \\
 & subsampling & $[0.1, 1]$  & - \\
 & max. depth & $[1, 15]$  & - \\
 & min. child weight & $[0, 20]$ & - \\ 
\bottomrule
\end{tabular}
\caption[Configuration space for Profet]{Hyper-parameter configuration space of the support vector machine (SVM), fully connected neural network (FC-Net) and the gradient tree boosting (XGBoost) benchmark.}
\label{tab:hypers_profet}
\end{table}

\section{Comparison Random Search vs. Bayesian Optimization on XGBoost}\label{sec:xgboost_profet_supp}

\begin{figure}[t!]
\centering
 \includegraphics[width=0.49\textwidth]{plots/error_xgboost_boston_housing.pdf}
 \includegraphics[width=0.49\textwidth]{plots/error_xgboost_combined_cycle_power_plant.pdf}\\
 \includegraphics[width=0.49\textwidth]{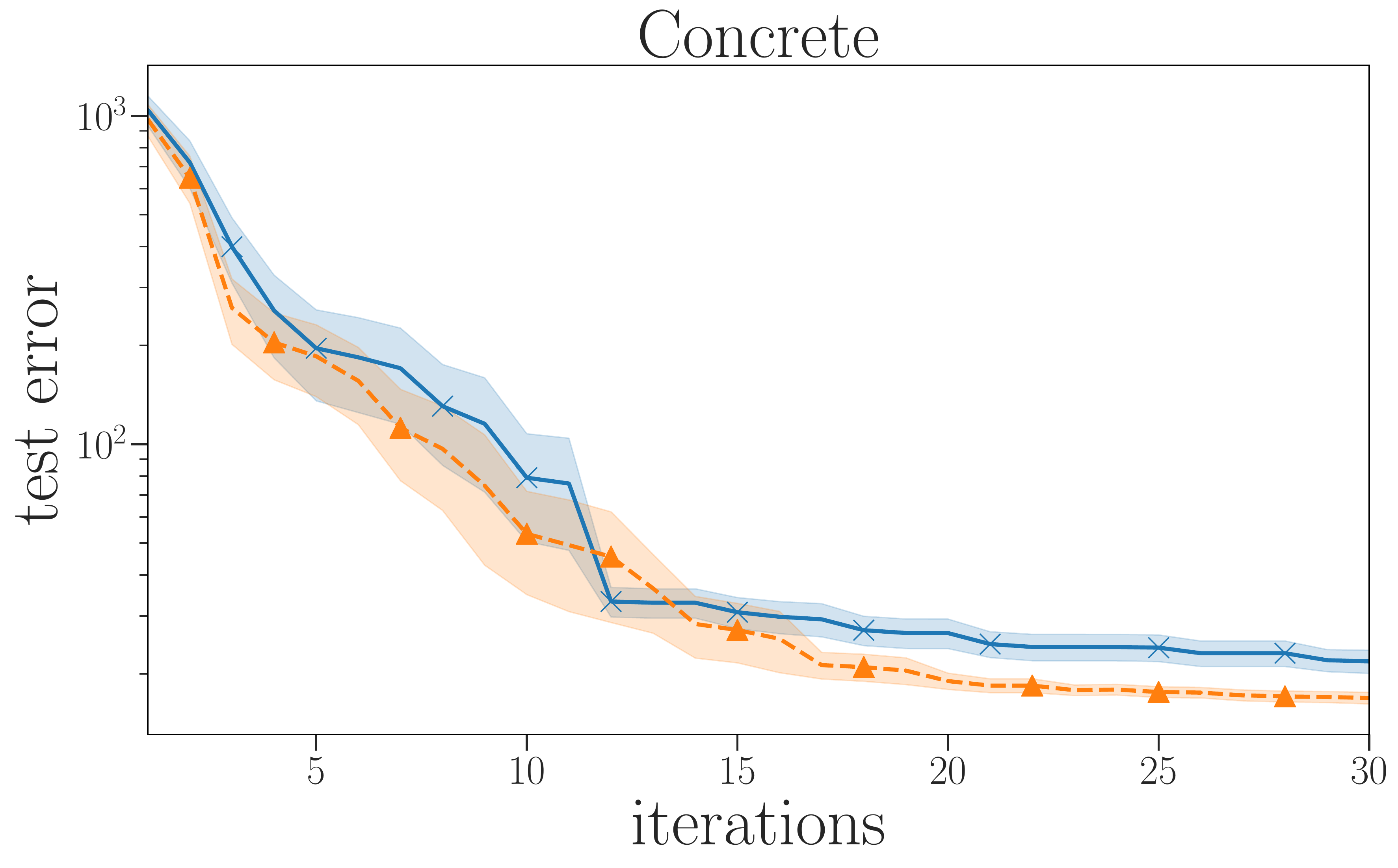}
 \includegraphics[width=0.49\textwidth]{plots/error_xgboost_energy_efficiency.pdf}\\
 \includegraphics[width=0.49\textwidth]{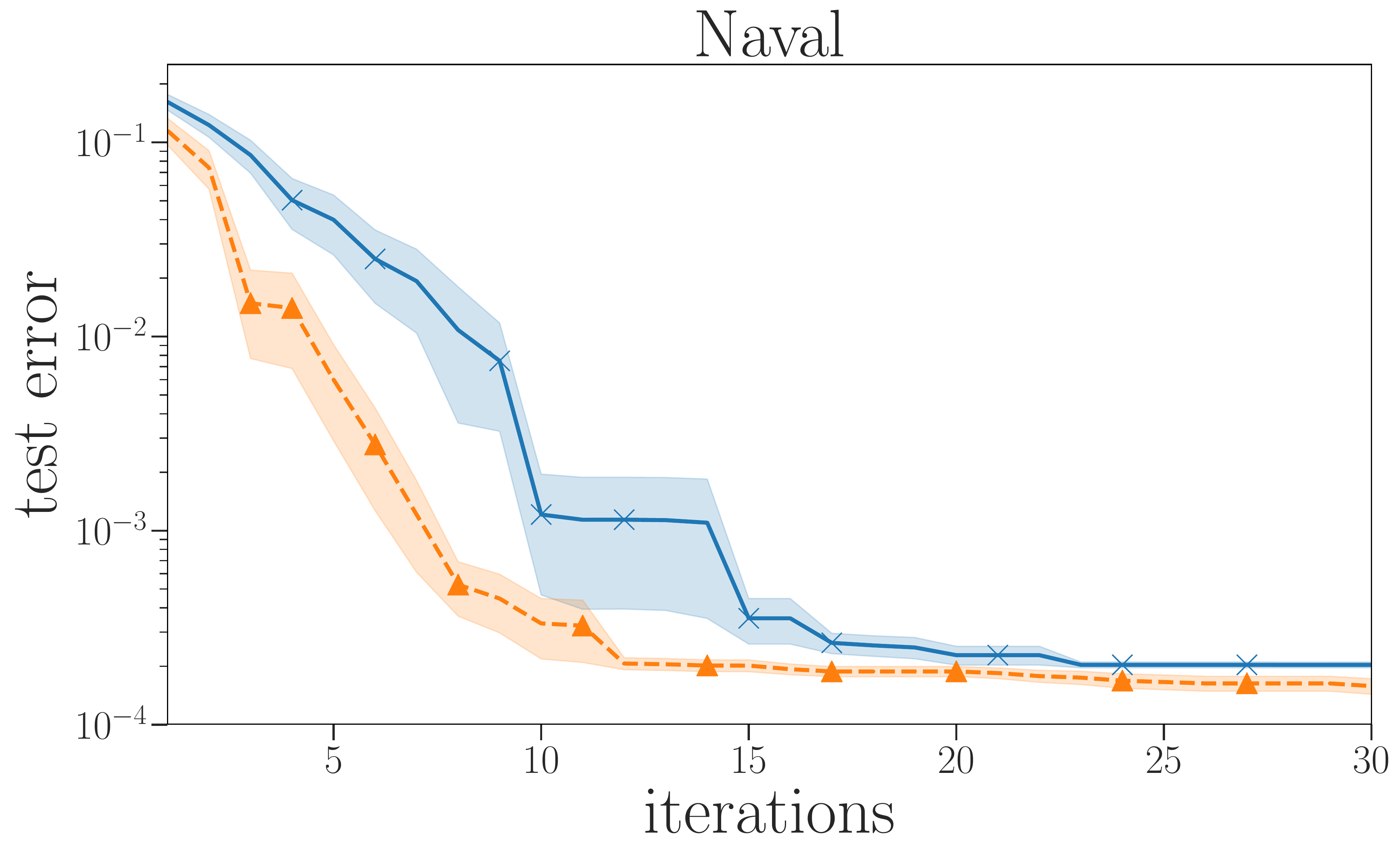}
 \includegraphics[width=0.49\textwidth]{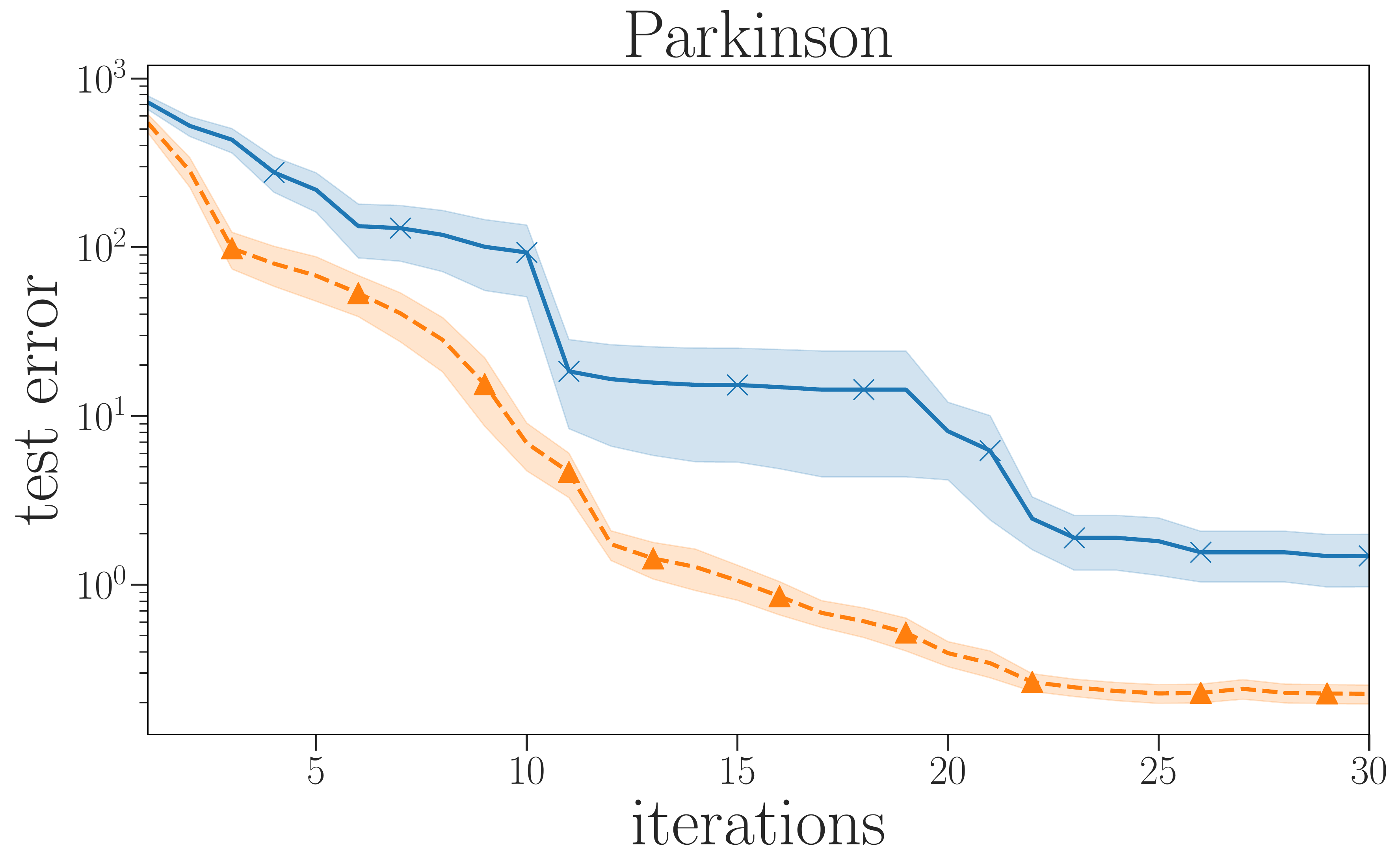}\\
 \includegraphics[width=0.49\textwidth]{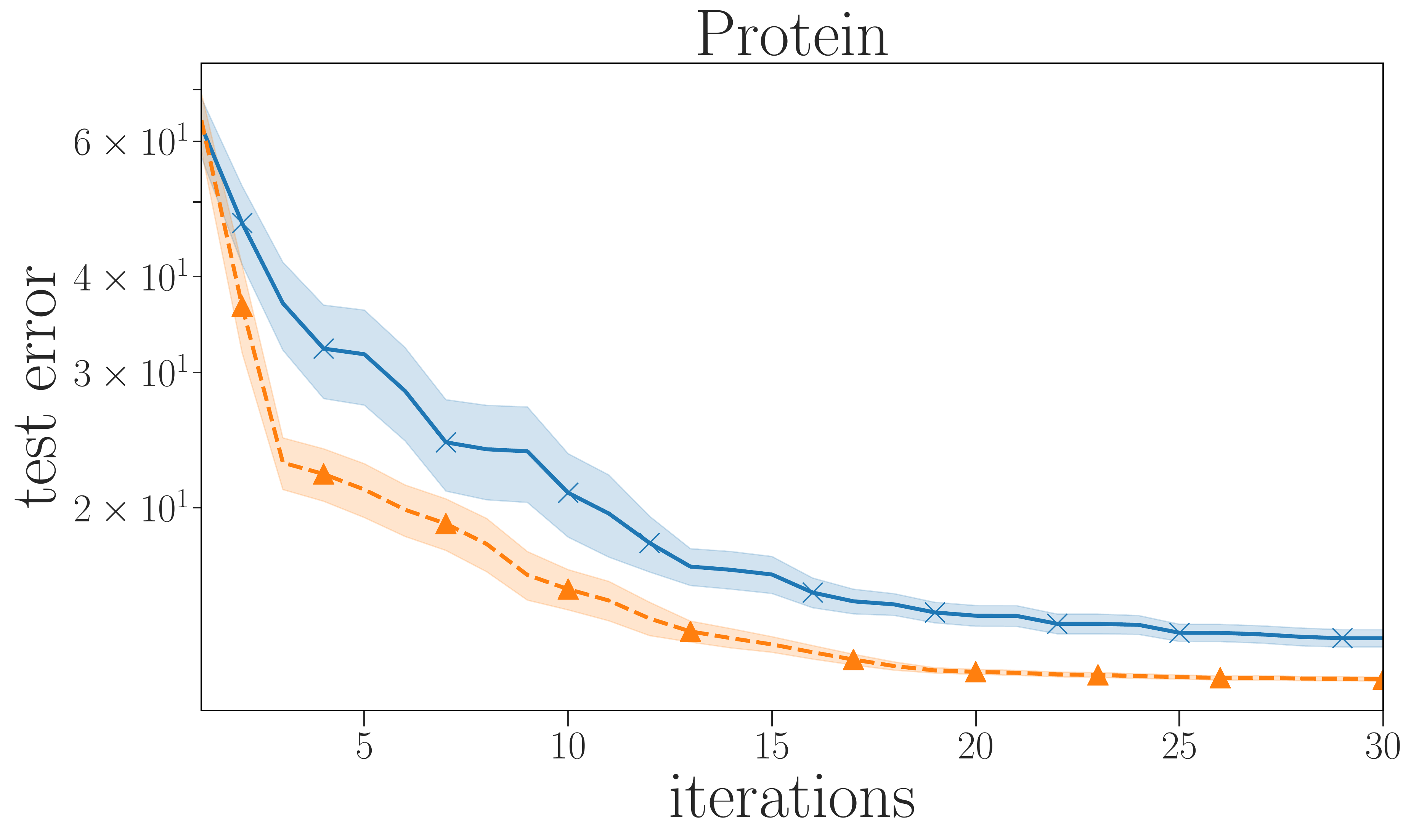}
 \includegraphics[width=0.49\textwidth]{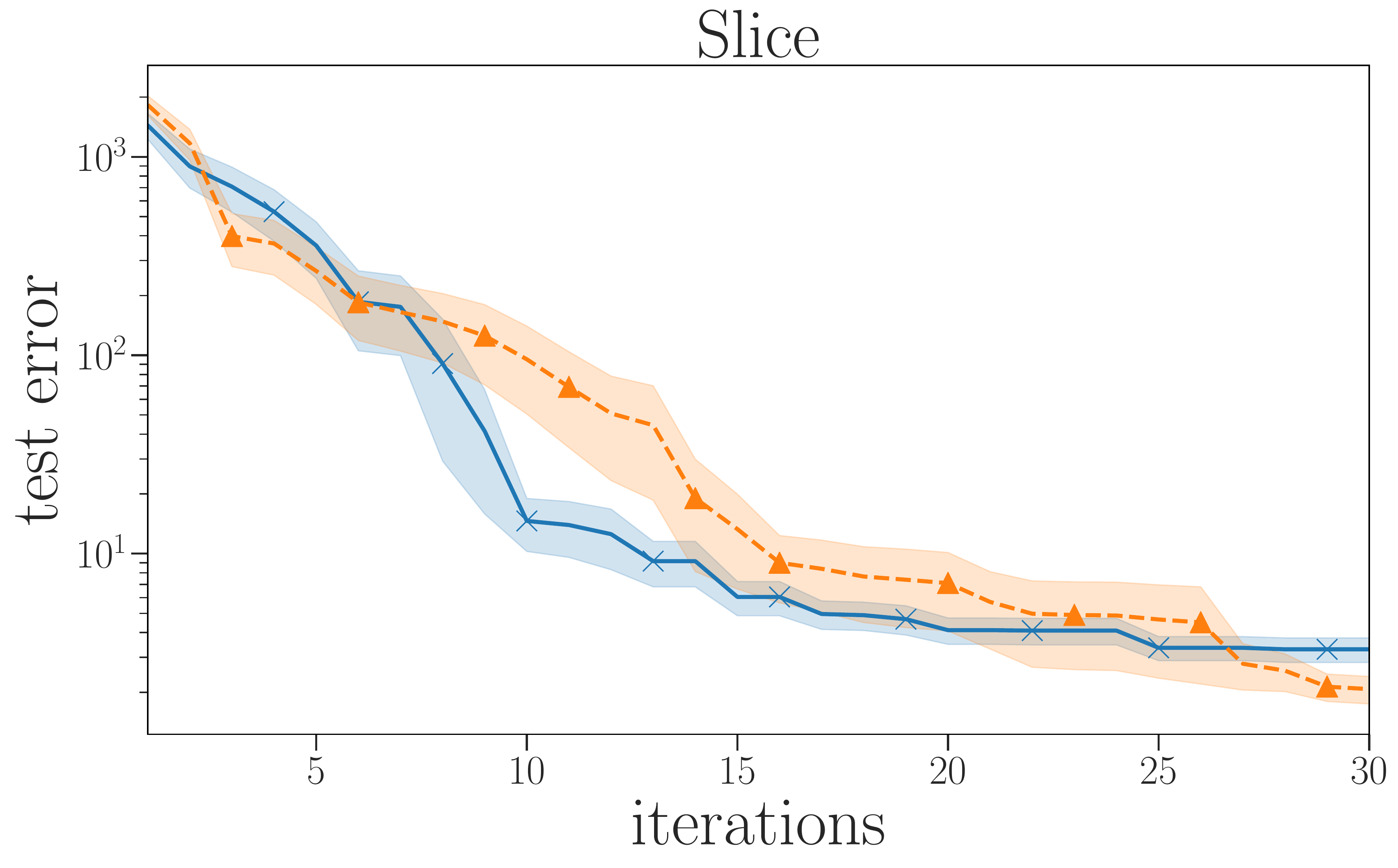}\\
 \includegraphics[width=0.49\textwidth]{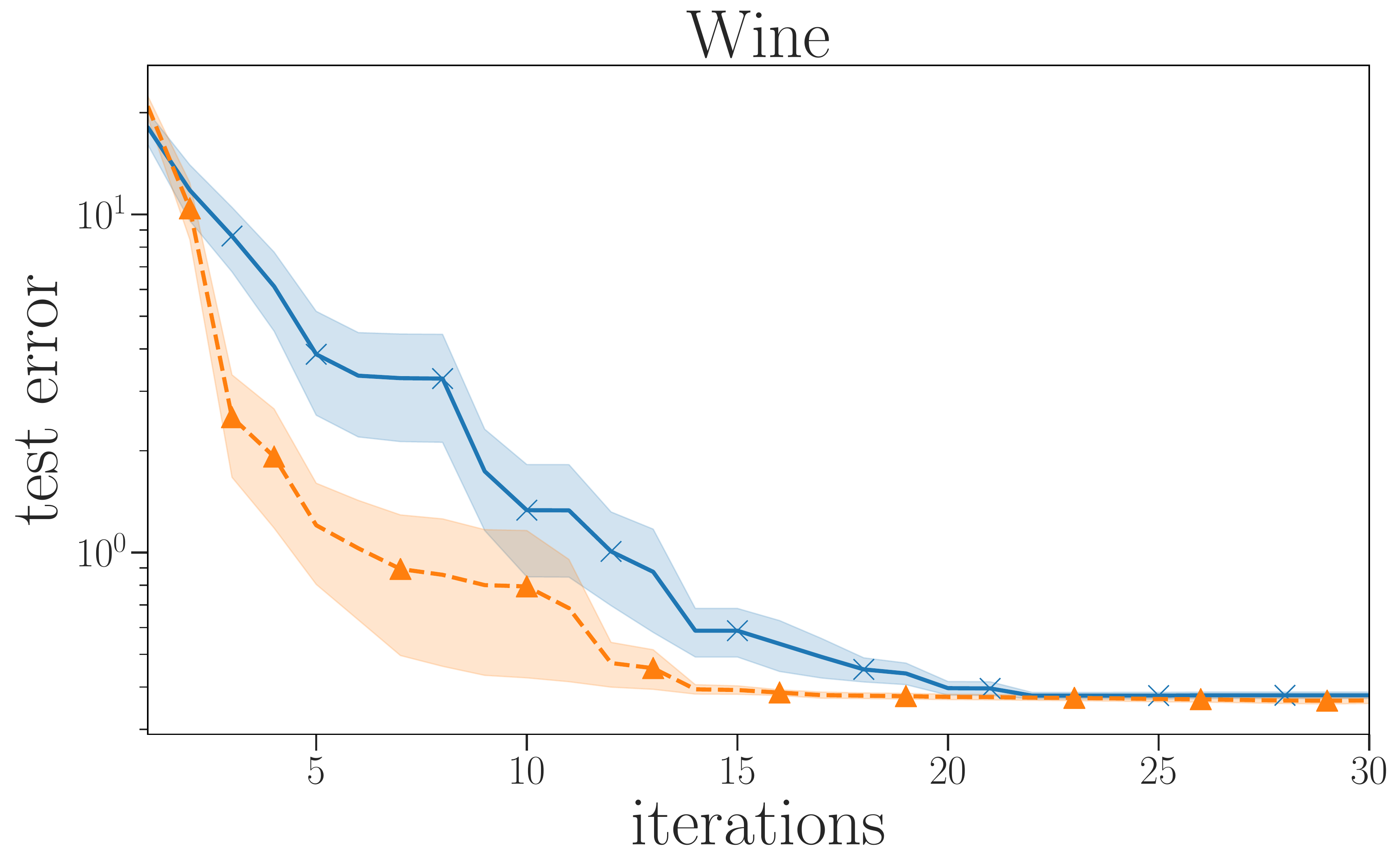}
 \includegraphics[width=0.49\textwidth]{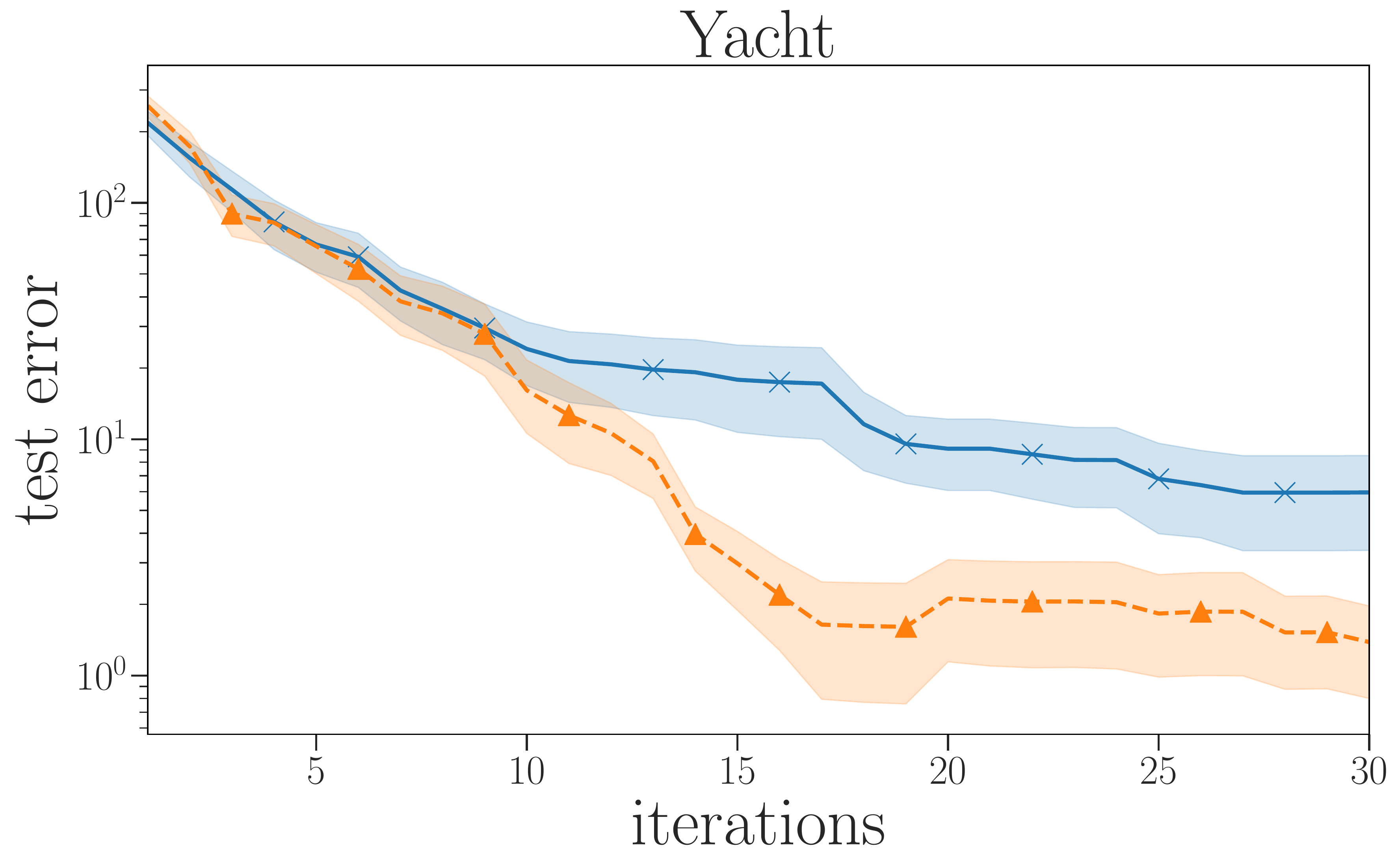}
 \caption[Comparison of all UCI examples]{Comparison of Bayesian optimization with Gaussian processes (GP-BO) and random search (RS) for optimizing the hyperparameters of XGBoost on 10 UCI regression datasets.}
 \label{fig:comparison_uci_all_profet}
\end{figure}

For completeness we show in Figure~\ref{fig:comparison_uci_all_profet} the comparison of random search (RS) and Bayesian optimization with Gaussian processes (BO-GP) on several UCI regression datasets.
Out of the 10 datasets, GP-BO perform better than RS on 7, worse on one, and ties on 2 and hence performs overall better than RS which is inline with the results obtained from out meta-model.
However, if we would look only on the first three datasets: Boston-Housing, PowerPlant and Concrete it would be much harder to draw strong conclusions.

\section{Details about the Forrester benchmark}

\begin{figure*}[t!]
\centering
 \includegraphics[width=0.32\textwidth]{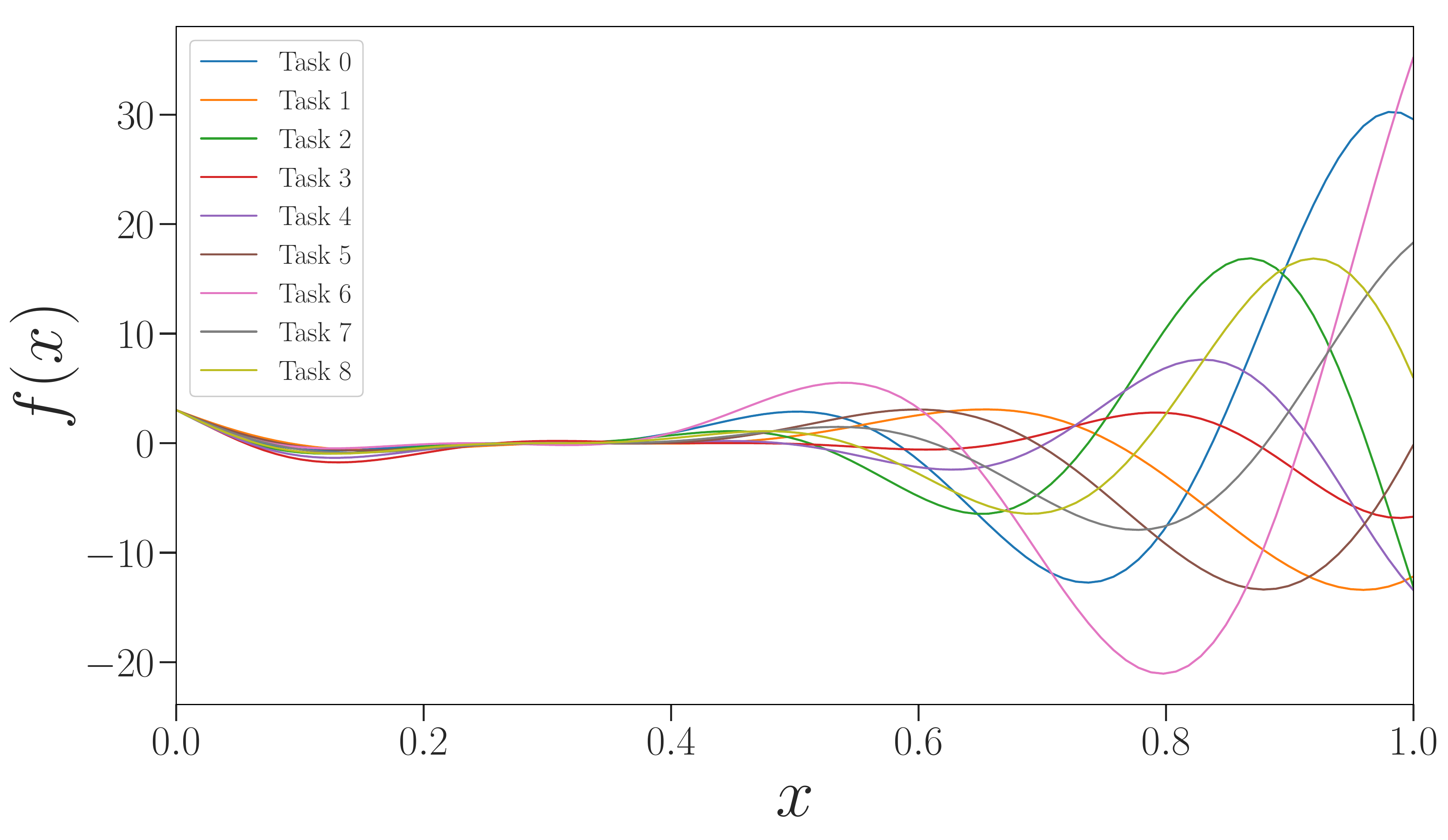}
 \includegraphics[width=0.32\textwidth]{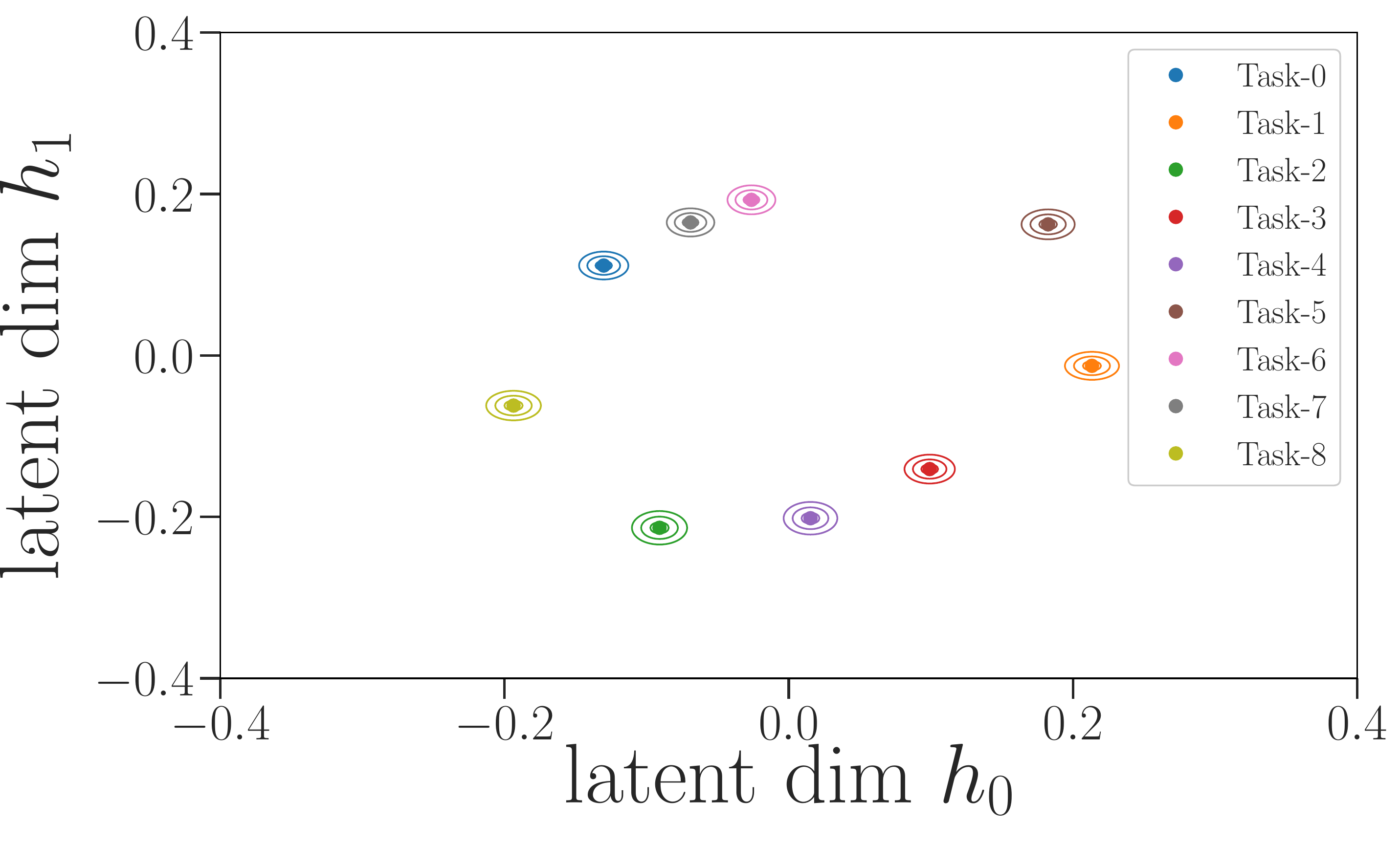}
 \includegraphics[width=0.32\textwidth]{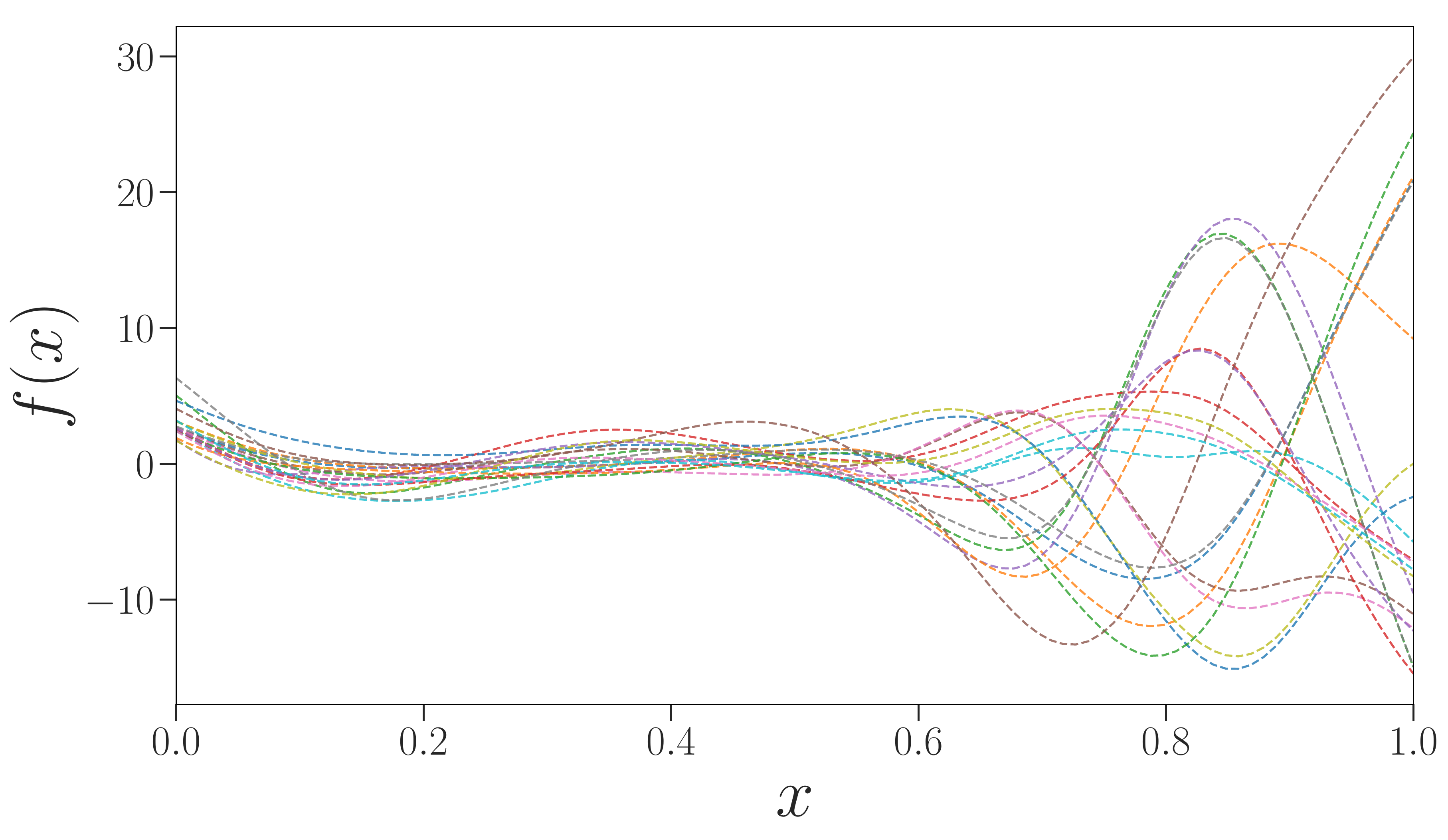}
\caption[Forrest example]{Visualizing the concept of our meta-model on the one-dimensional Forrester function. \emph{Left:} 9 different tasks (solid lines) coming from the same distribution.  \emph{Middle:} We use a probabilistic encoder to learn a two-dimensional latent space for the task embedding. \emph{Right:} Given our encoder and the multi-task model we can generate new tasks (dashed lines) that, based on the collected data, resemble the original tasks.}
 \label{fig:forrester_2}
\end{figure*}

Figure \ref{fig:forrester_2} shows the original 9 tasks (left), their representation on the latent space of the model (middle) and an example of 10 new generated task (right), that resemble the original ones.

\section{Samples for the Meta-SVM benchmark}\label{sec:samples_profet_supp}

\begin{figure*}[h]
\centering
\includegraphics[width=0.19\columnwidth]{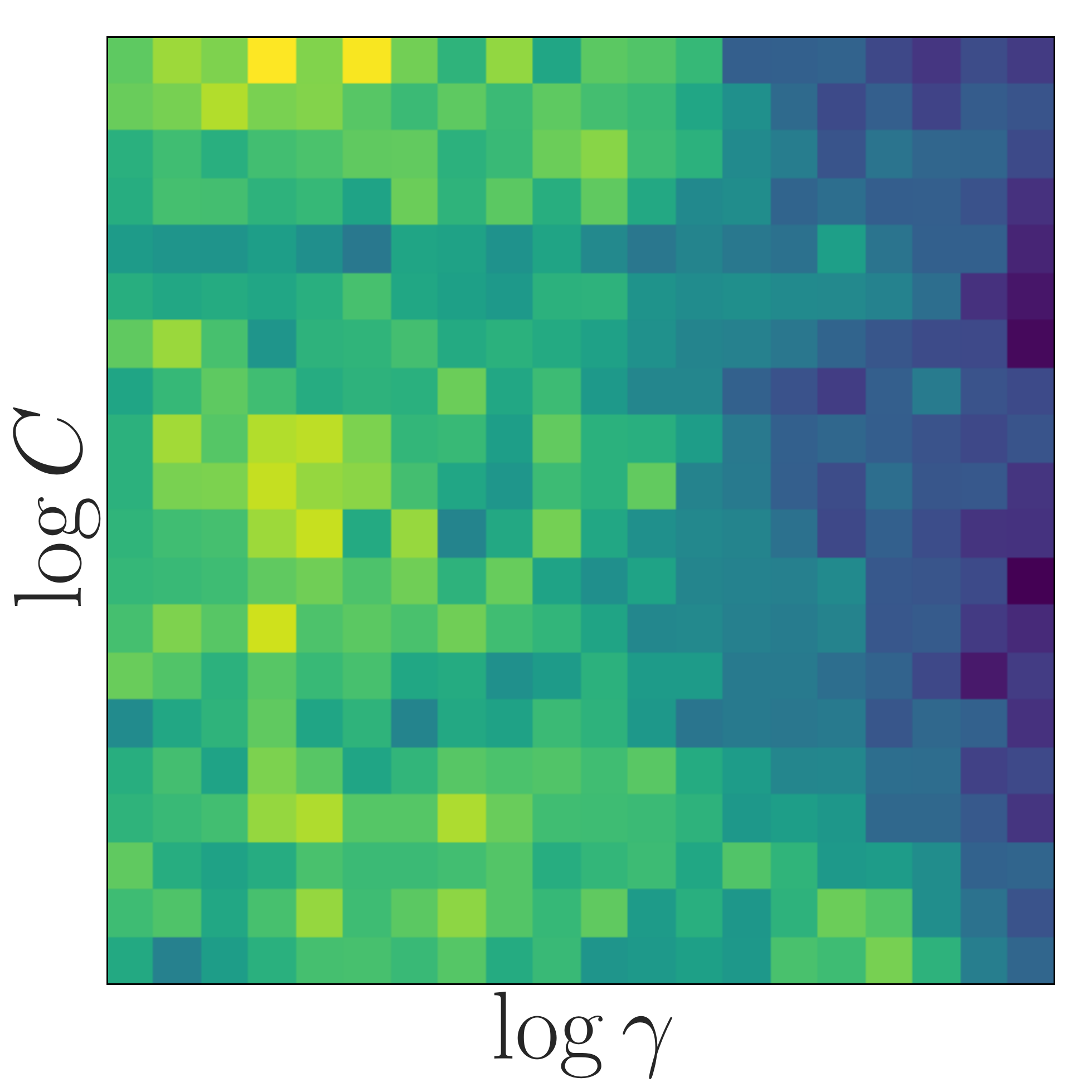}
\includegraphics[width=0.19\columnwidth]{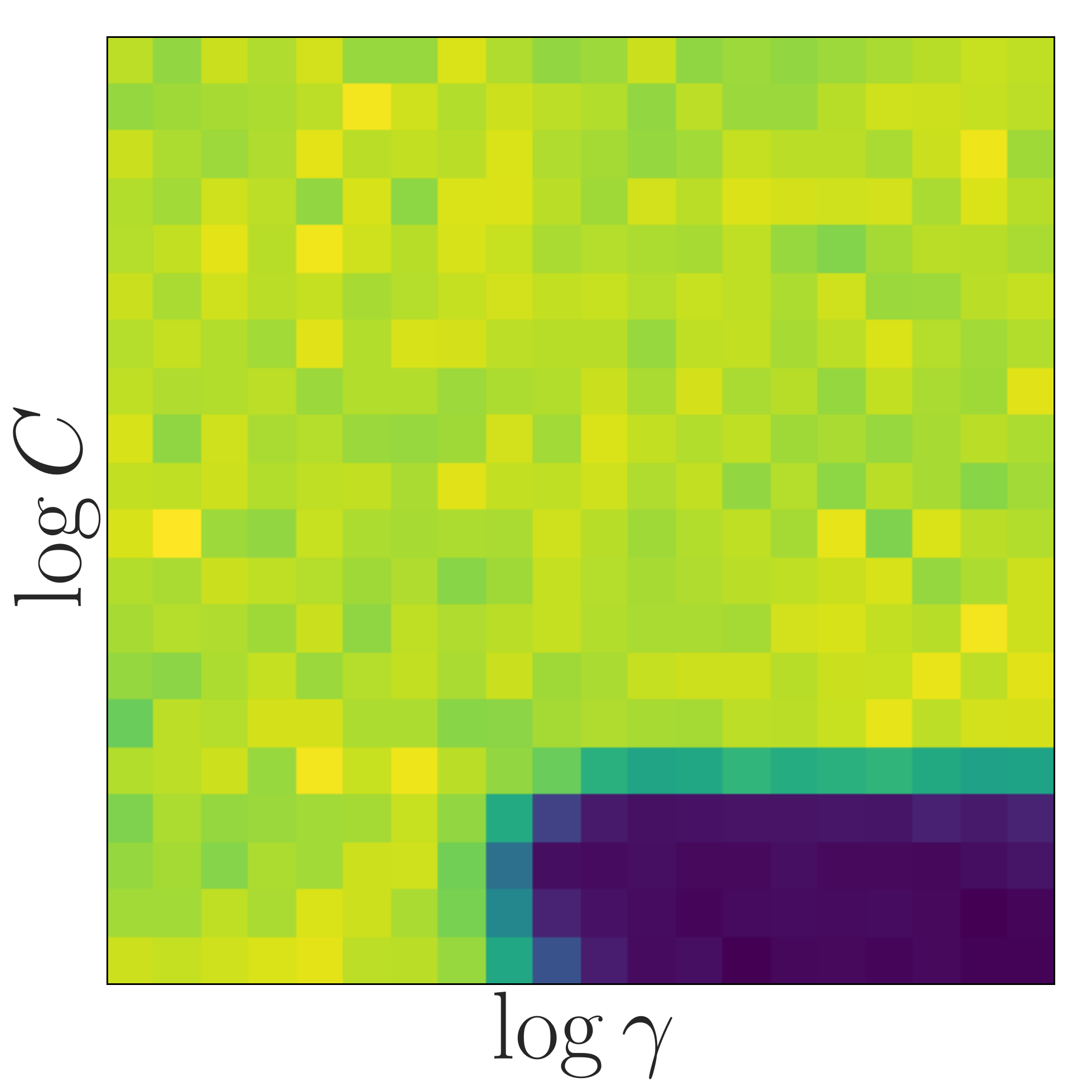}
\includegraphics[width=0.19\columnwidth]{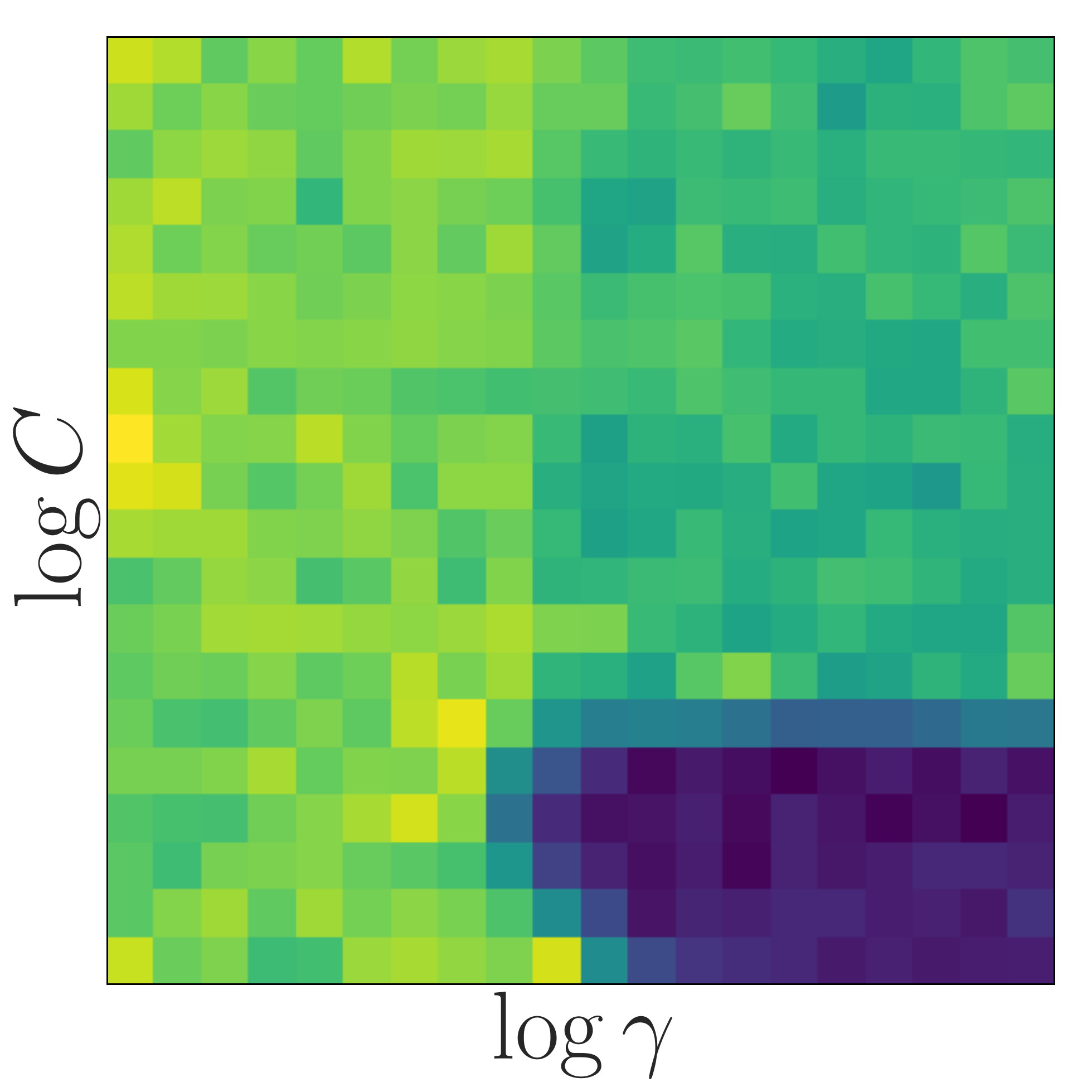}
\includegraphics[width=0.19\columnwidth]{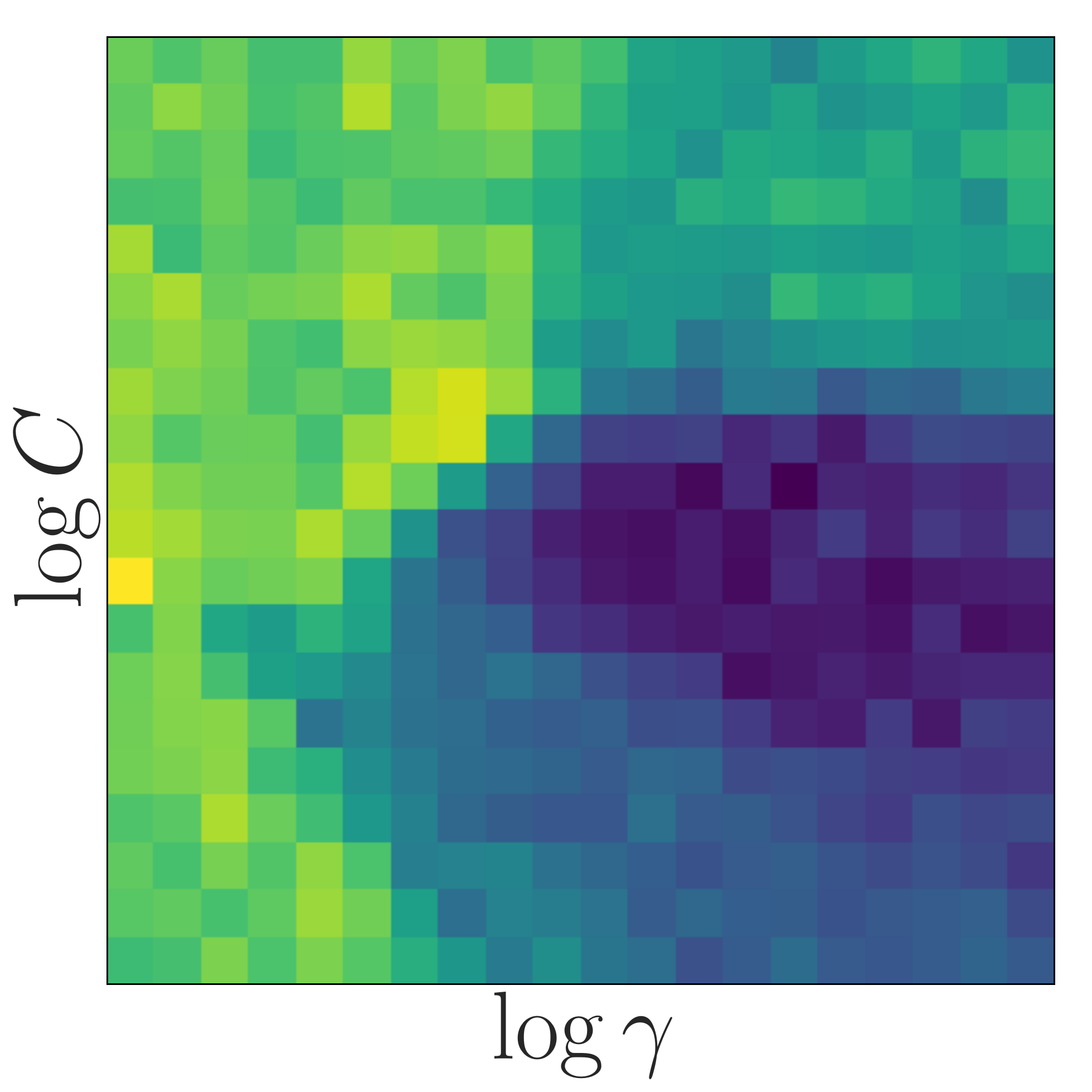}
\includegraphics[width=0.19\columnwidth]{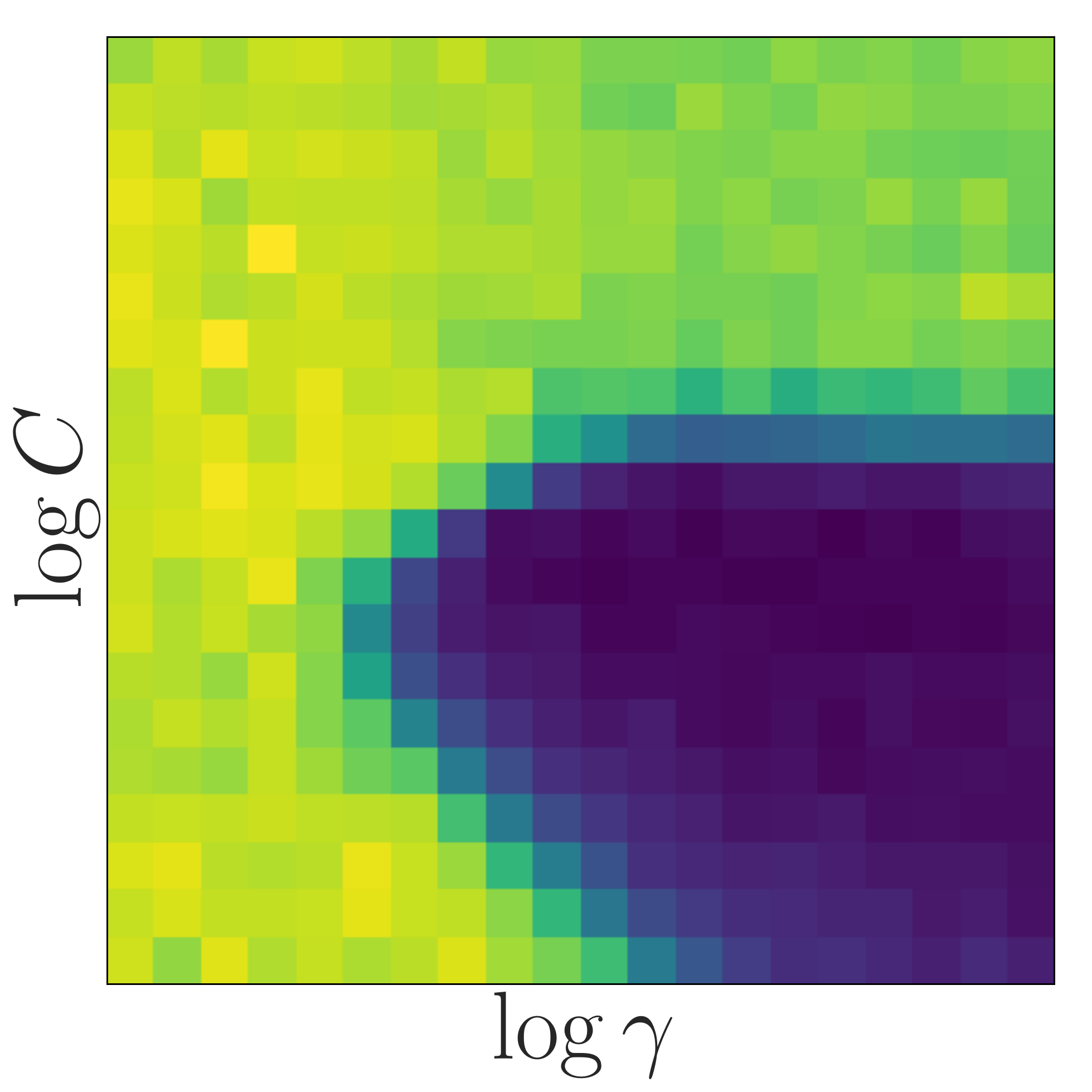}\\
\includegraphics[width=0.19\columnwidth]{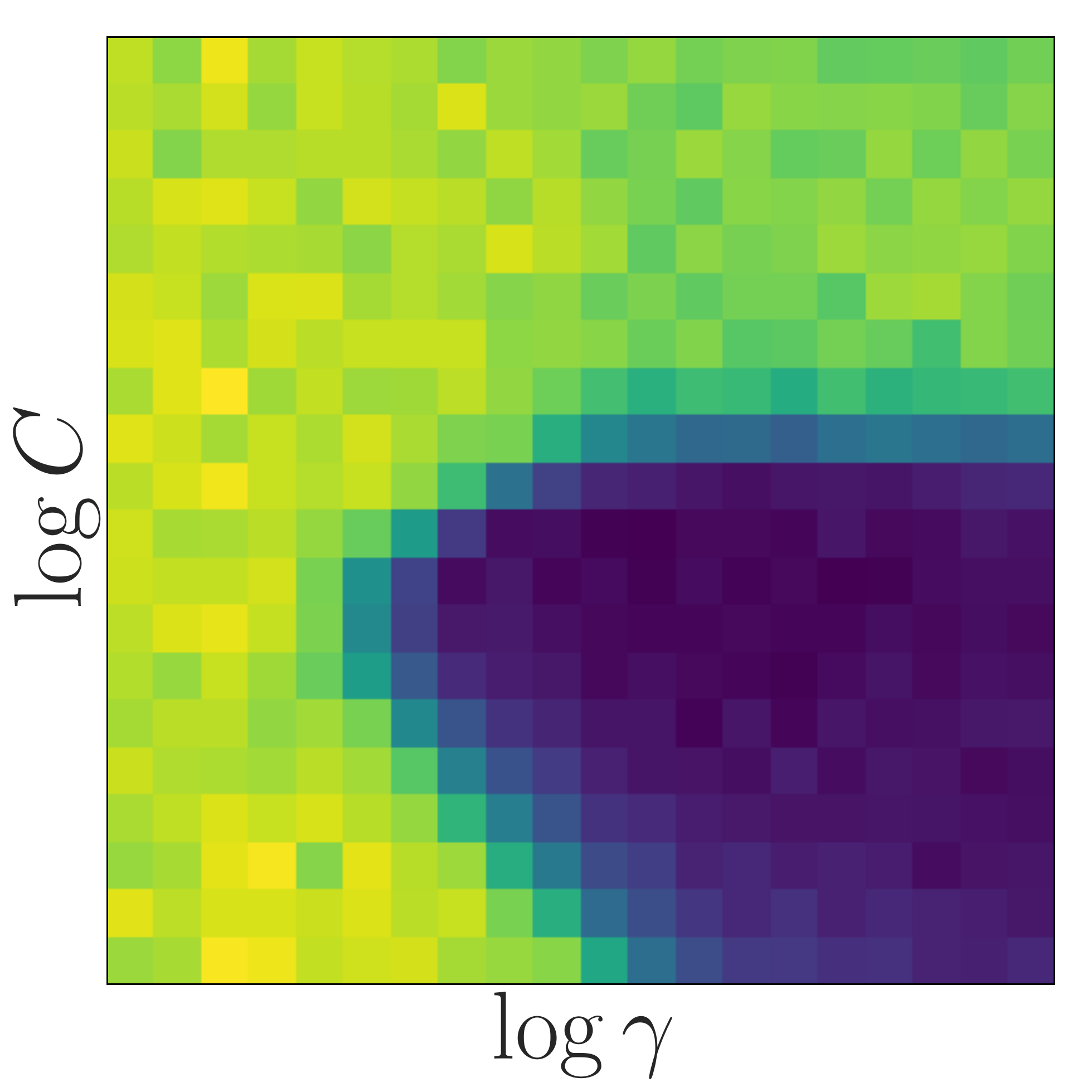}
\includegraphics[width=0.19\columnwidth]{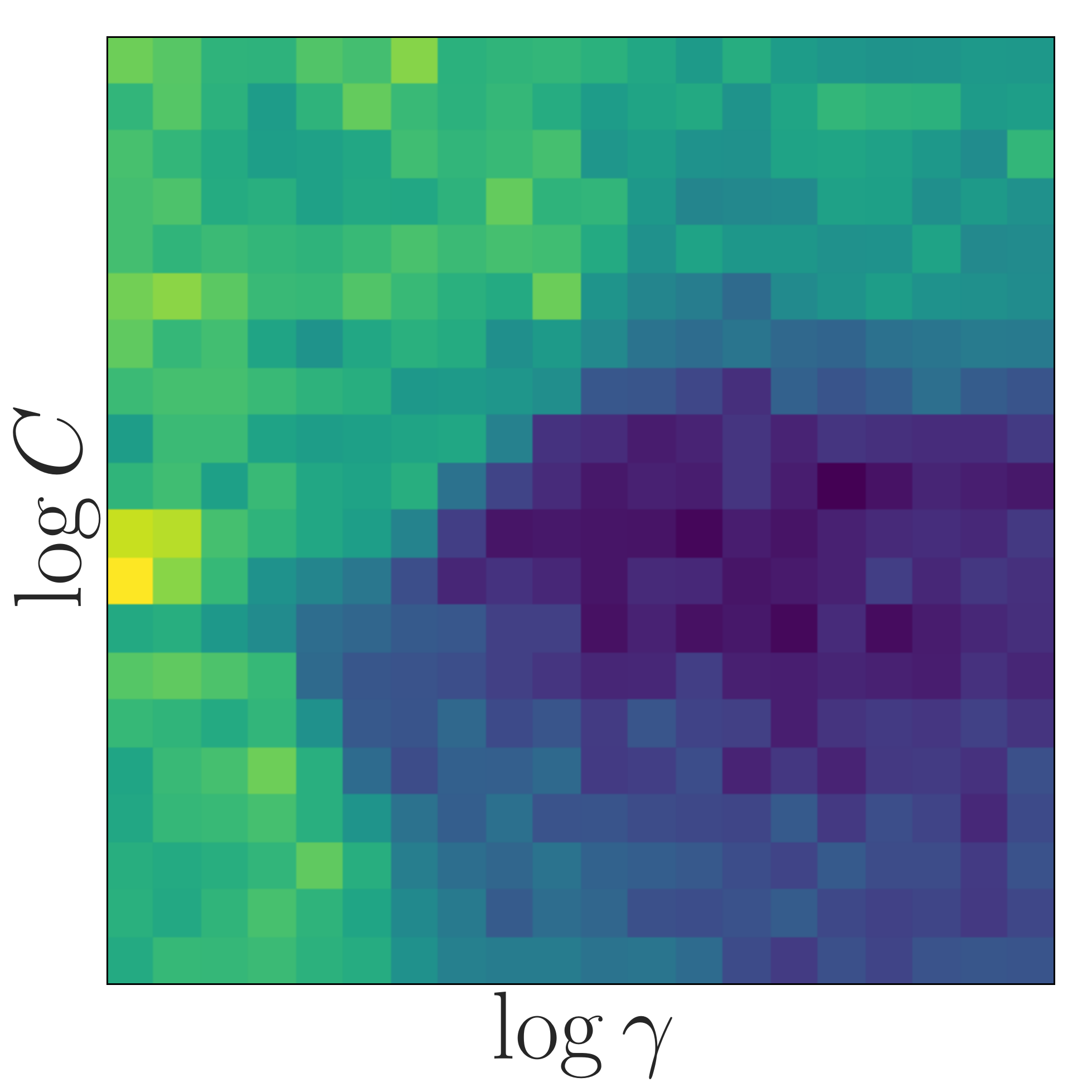}
\includegraphics[width=0.19\columnwidth]{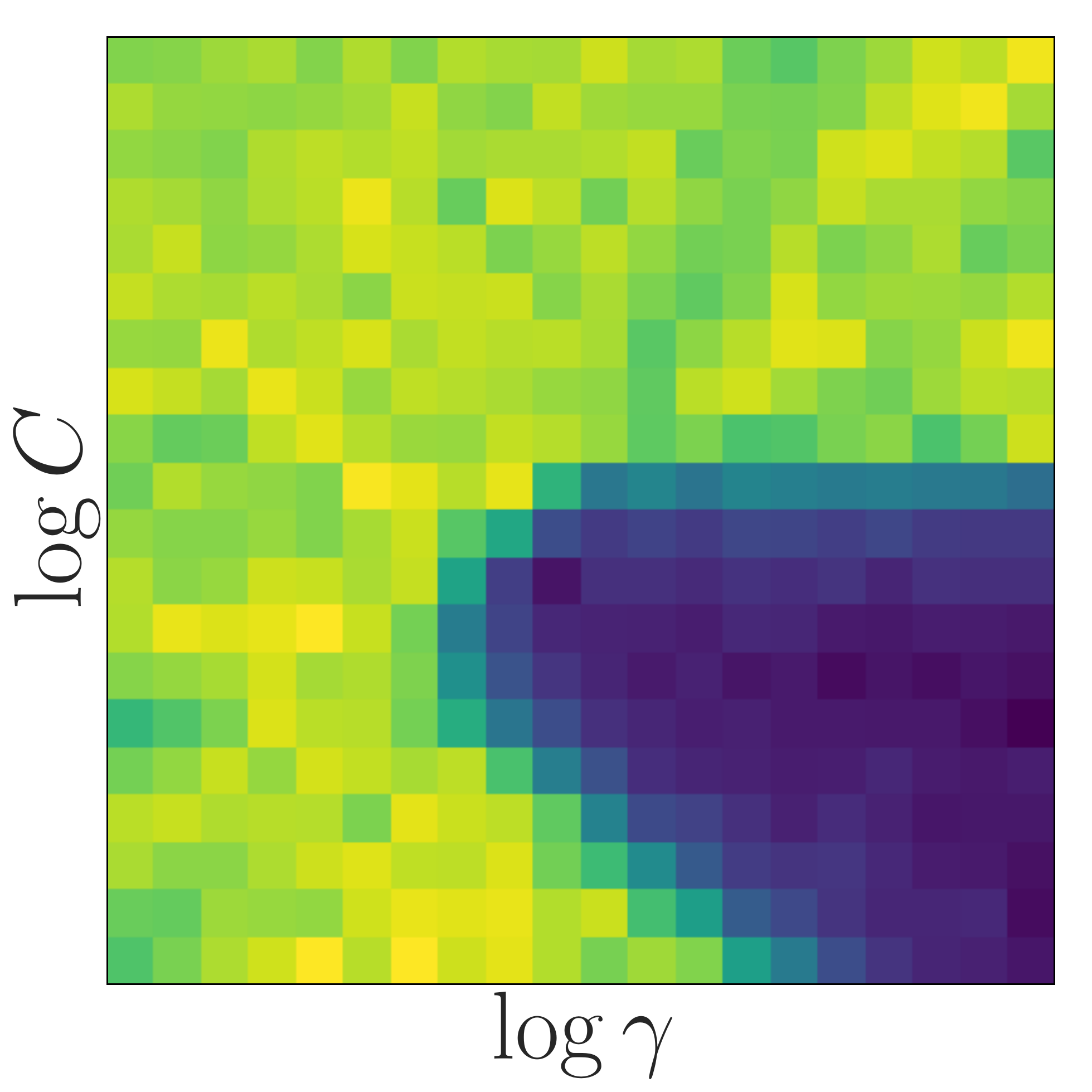}
\includegraphics[width=0.19\columnwidth]{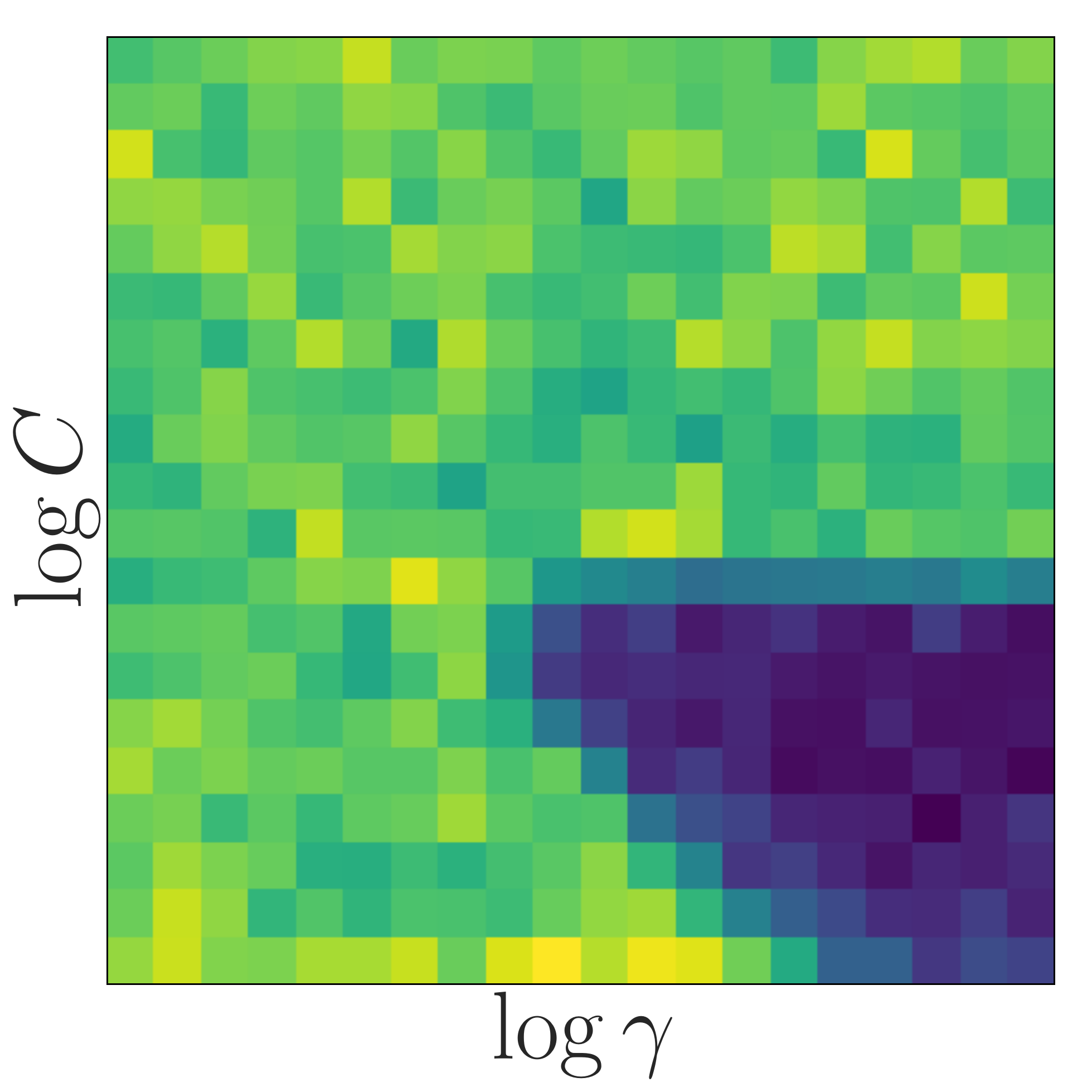}
\includegraphics[width=0.19\columnwidth]{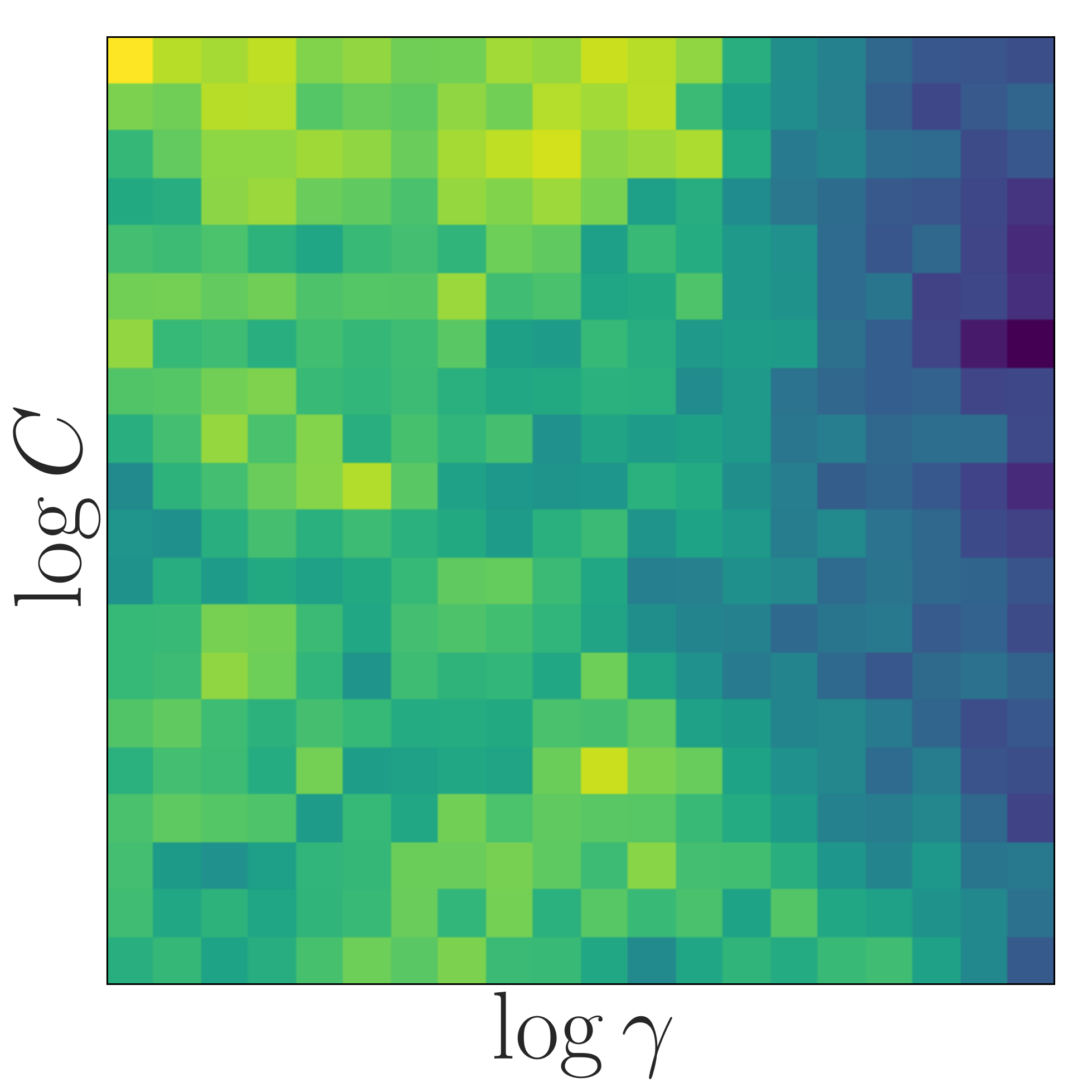}
\caption[Noisy samples for SVM benchmark]{Noisy samples from our meta-model for the SVM benchmark}
\label{fig:all_noisy_samples}
\end{figure*}

In Figure~\ref{fig:all_noisy_samples} and Figure~\ref{fig:all_noisy_samples} we show additional randomly sampled tasks with and without noise.
One can see that, while the general characteristics of the original objective function, \ie{} bowl shaped around the lower right corner, remains, the local structure changes across samples.

\begin{figure*}[h]
\centering
\includegraphics[width=0.19\columnwidth]{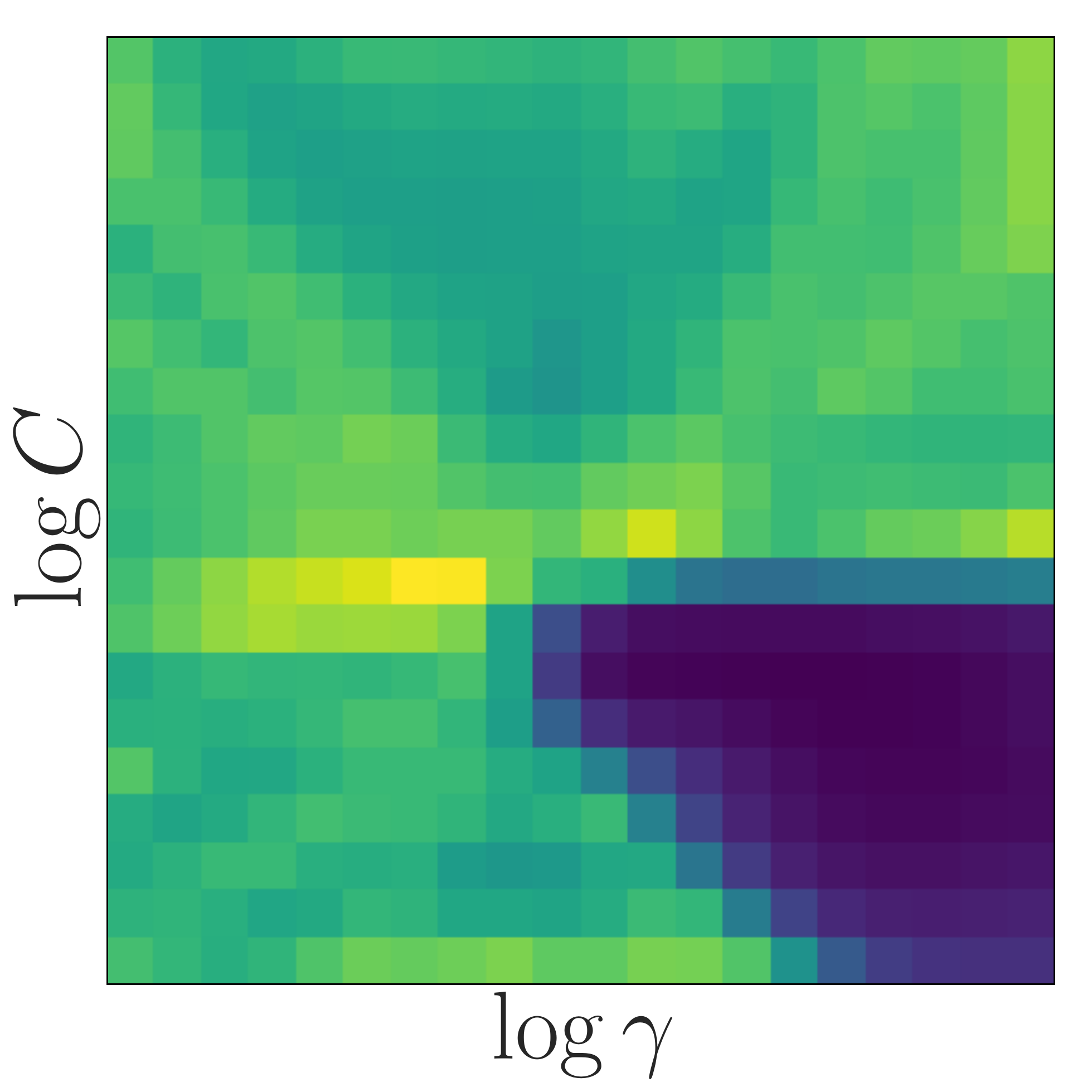}
\includegraphics[width=0.19\columnwidth]{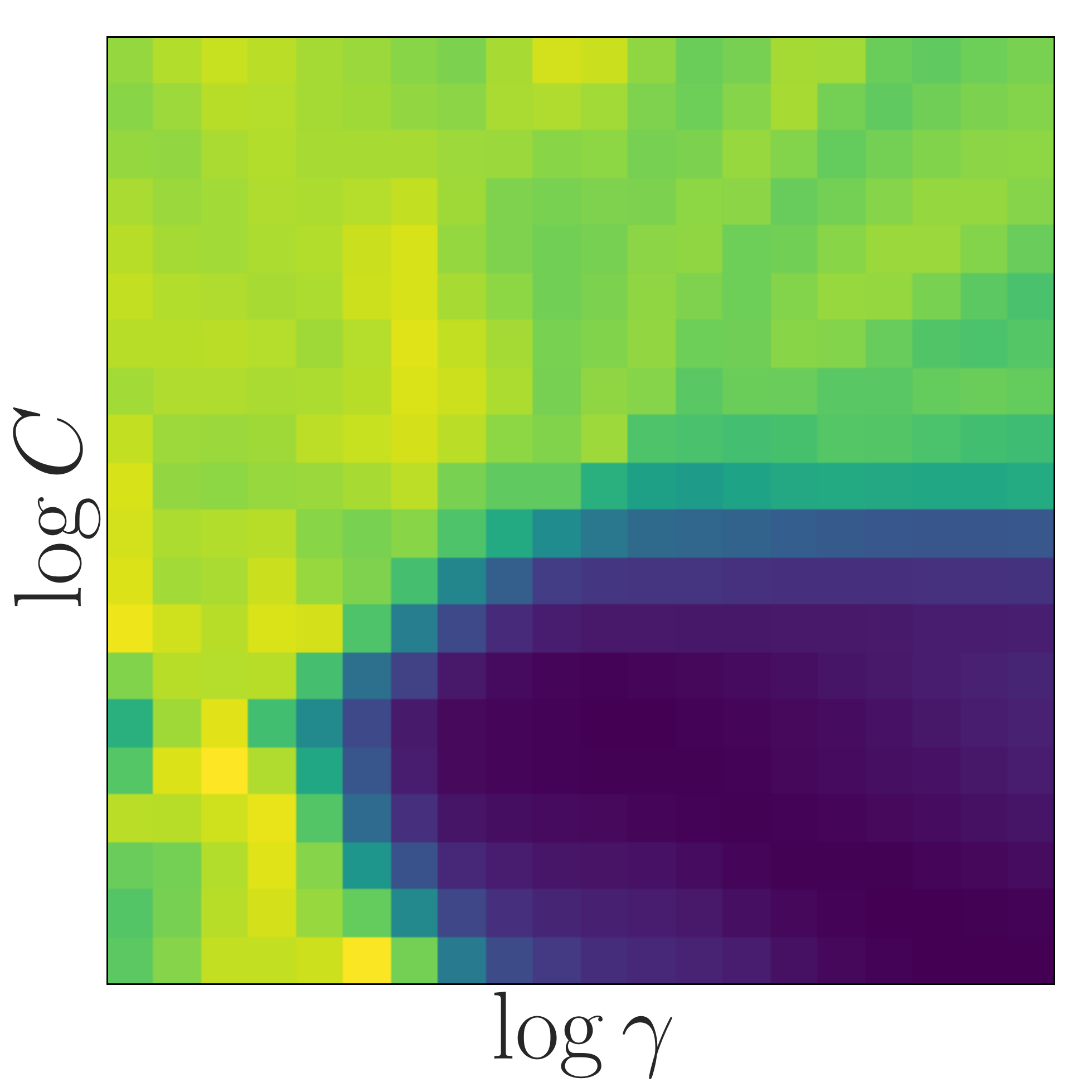}
\includegraphics[width=0.19\columnwidth]{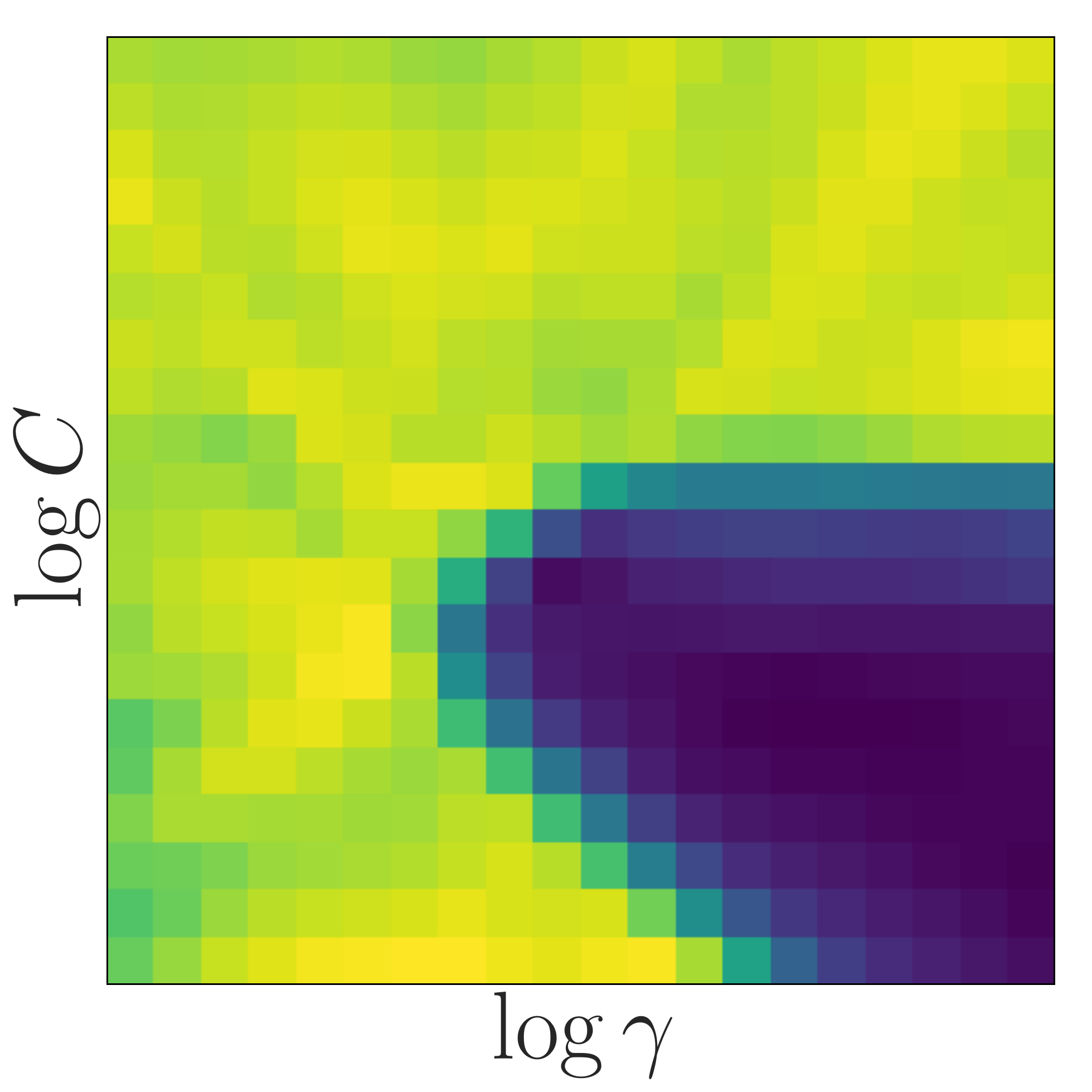}
\includegraphics[width=0.19\columnwidth]{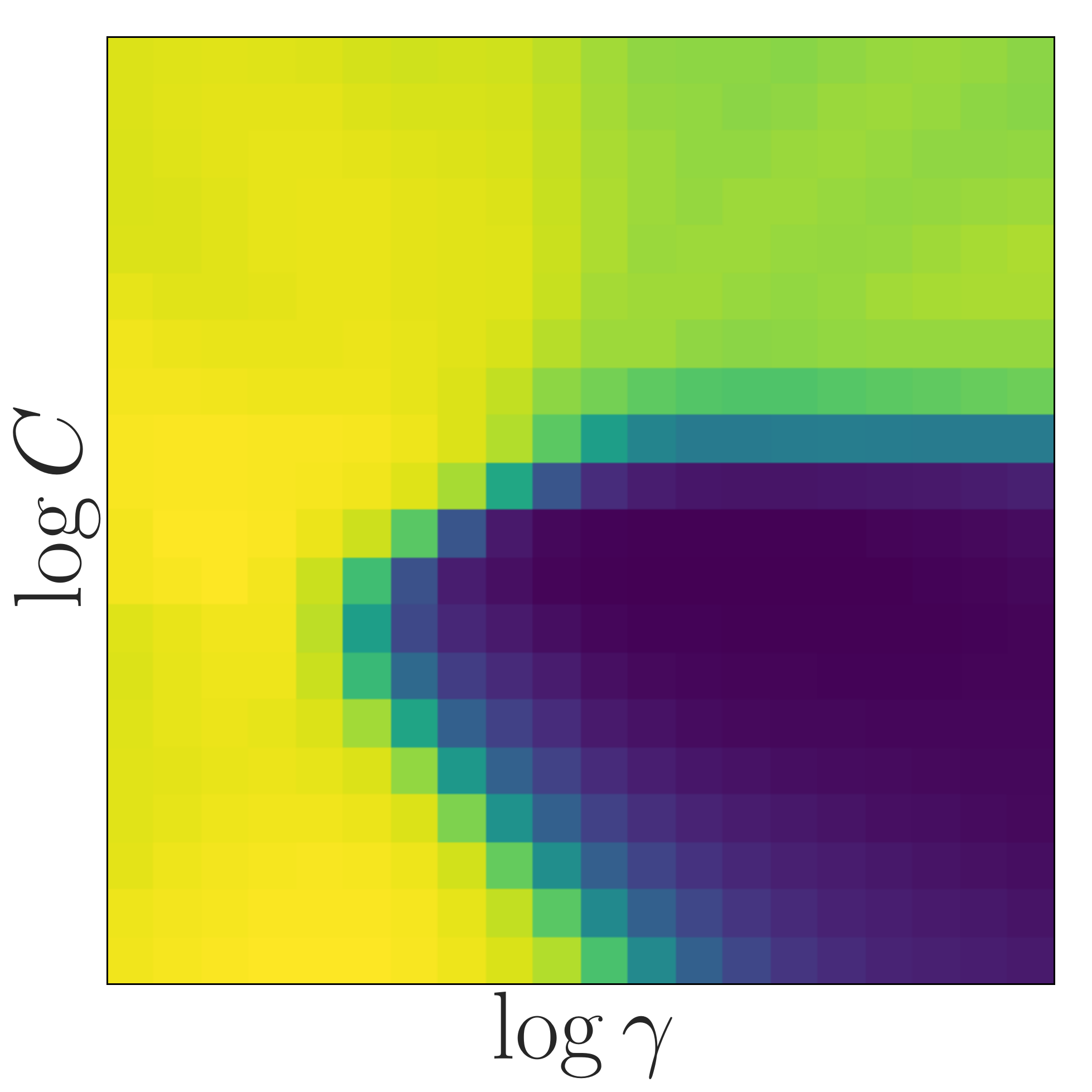}
\includegraphics[width=0.19\columnwidth]{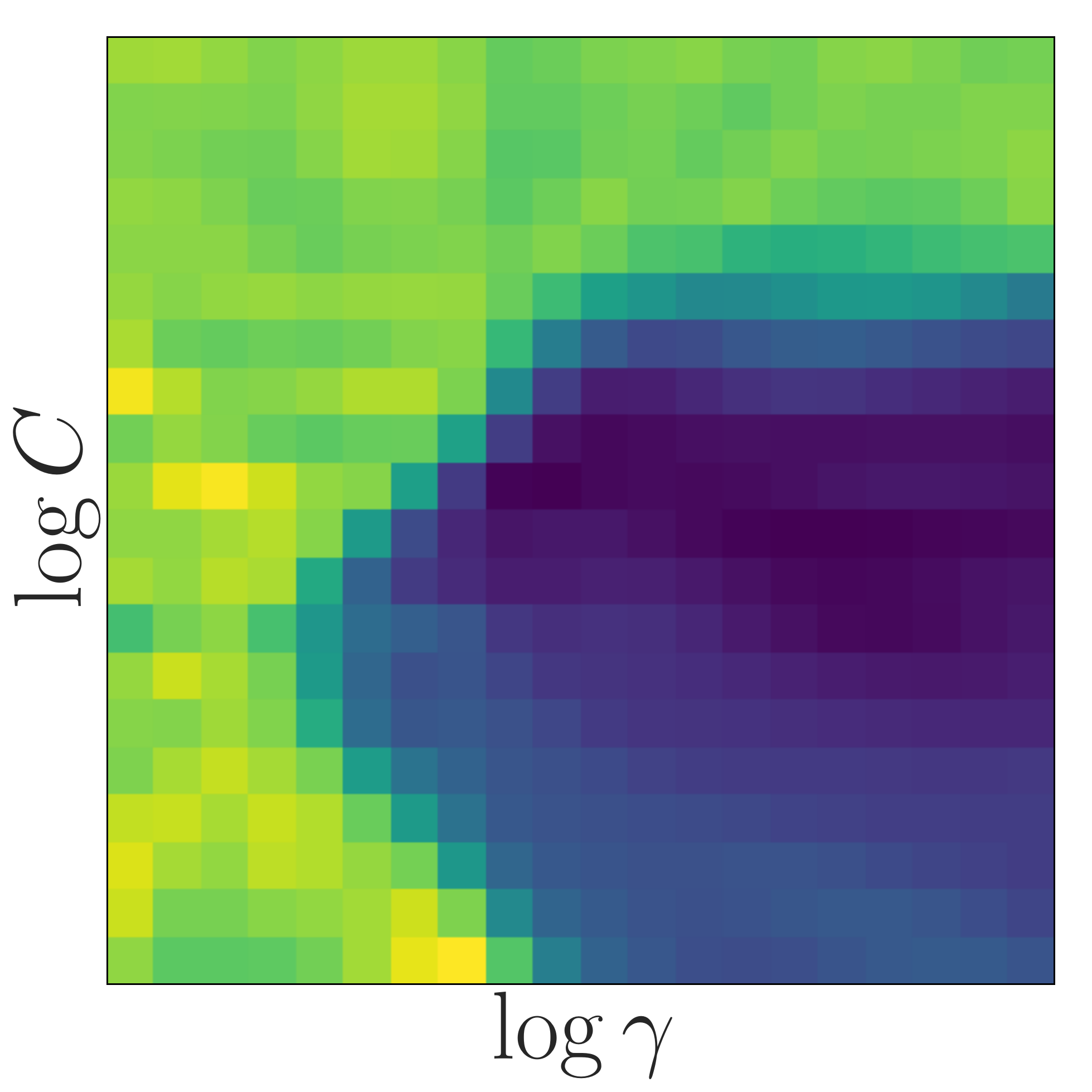}\\
\includegraphics[width=0.19\columnwidth]{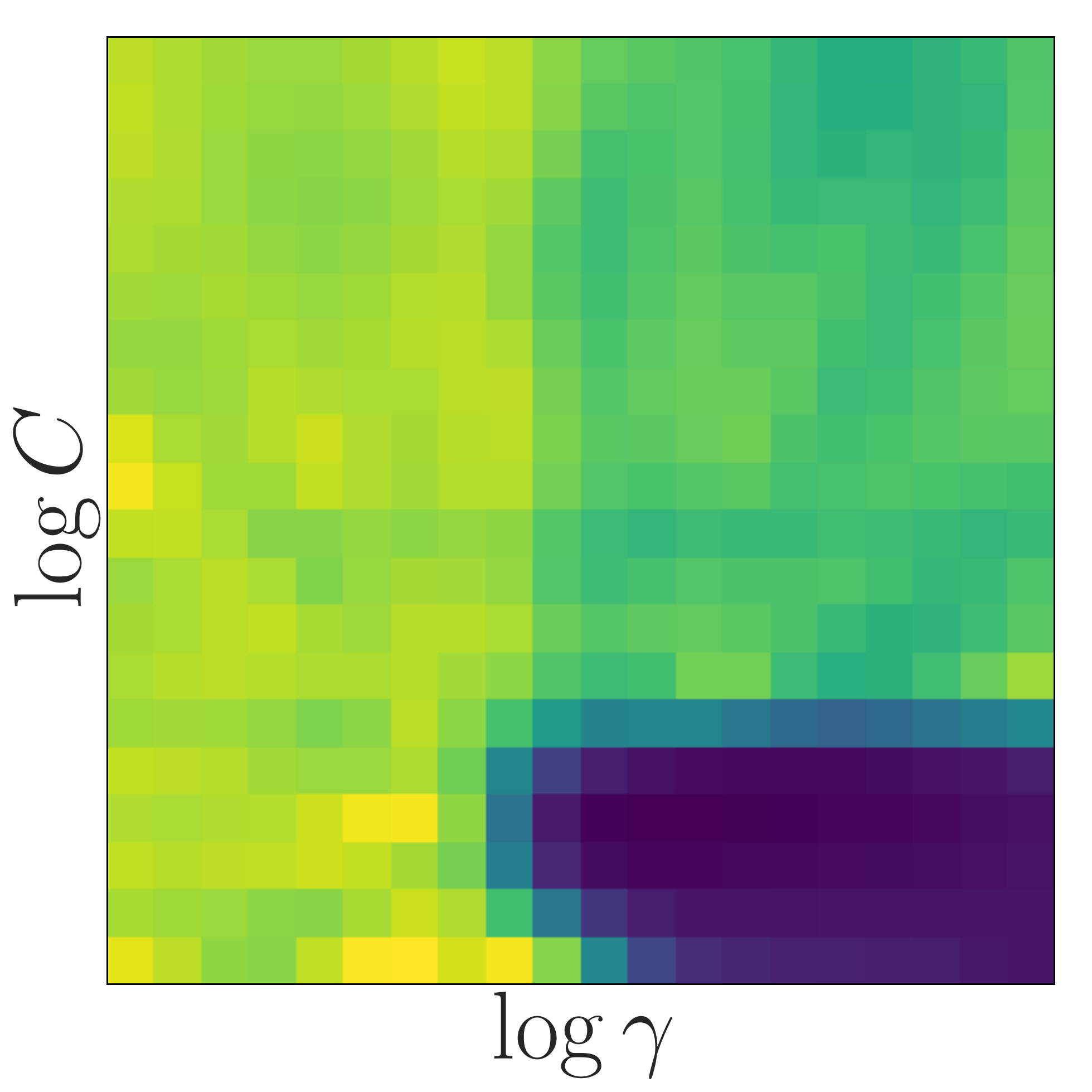}
\includegraphics[width=0.19\columnwidth]{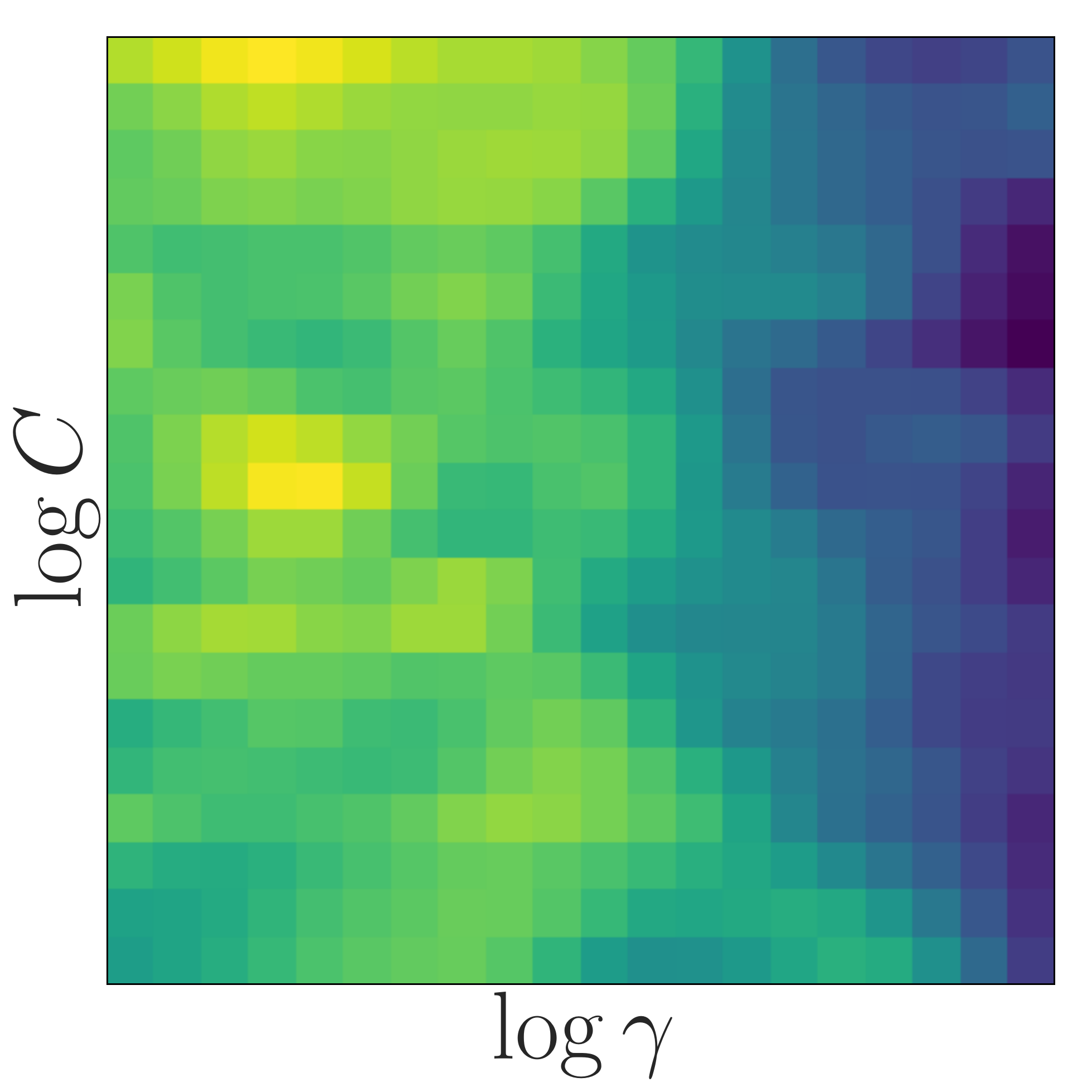}
\includegraphics[width=0.19\columnwidth]{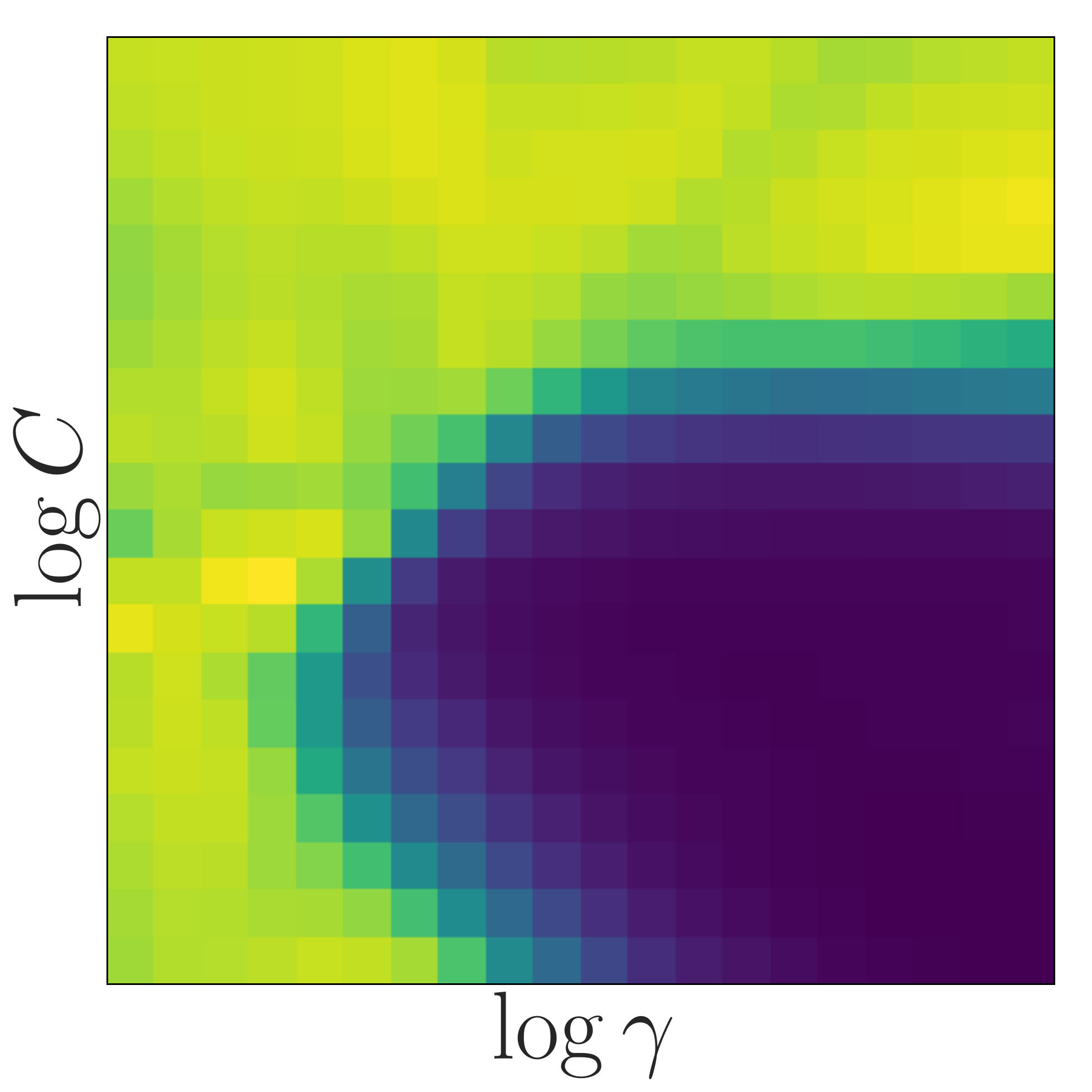}
\includegraphics[width=0.19\columnwidth]{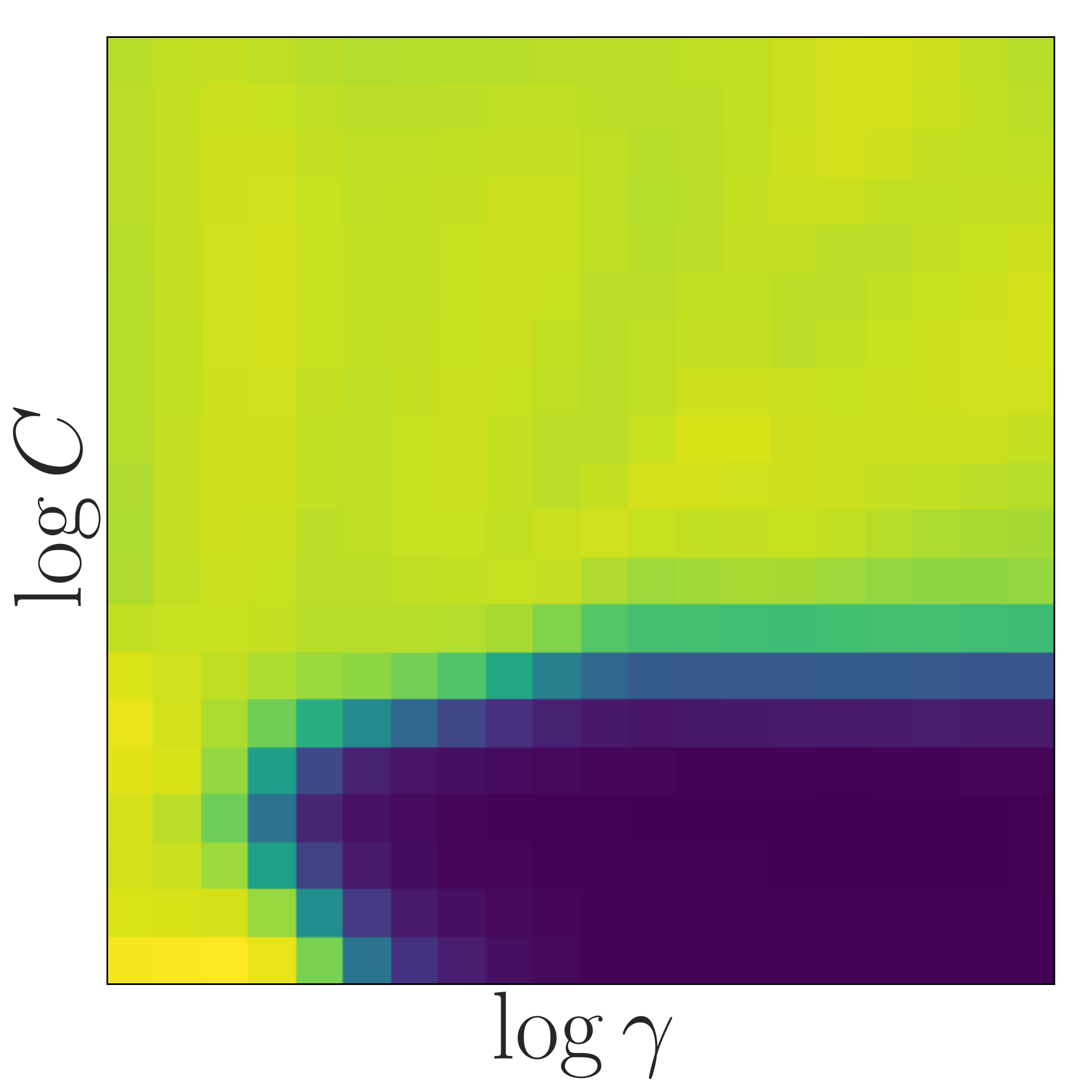}
\includegraphics[width=0.19\columnwidth]{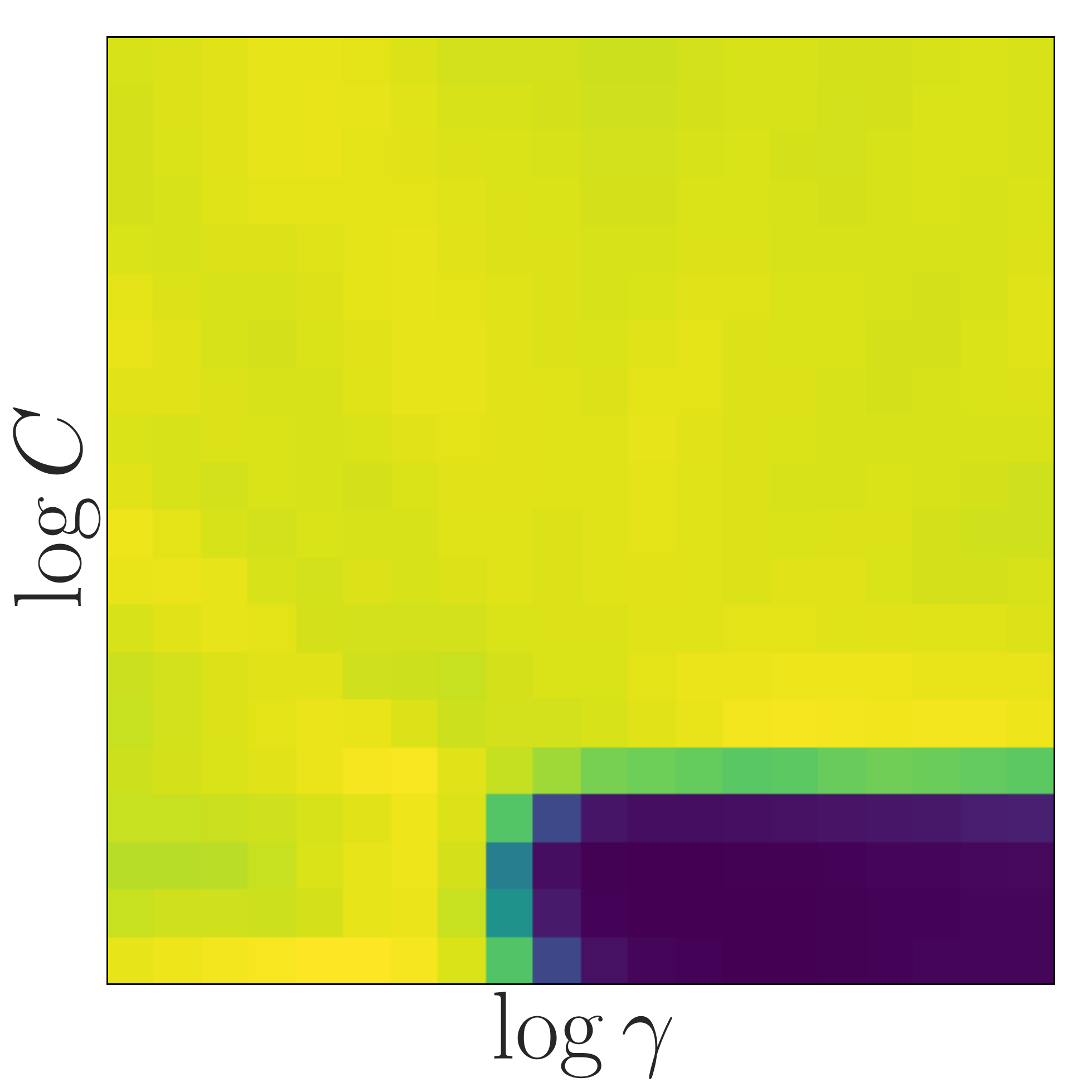}
\caption[Noisless samples for SVM benchmark]{Noiseless samples from our meta-model for the SVM benchmark}
\label{fig:all_noisless_samples}
\end{figure*}

\section{Comparison of \hpo Methods}\label{sec:results_profet_supp}

We now described the specific details of each optimizer in turn. 

\textbf{Random search} (RS)~\citep{bergstra-jmlr12a} We defined a uniform distribution over the input space and, in each iteration, randomly sampled a datapoint from this distribution.

\textbf{Differential Evolution} (DE)~\citep{storn-jgo97} maintains a population of data points and generates new candidate points by mutation random points from this population.
We defined the probability for mutation and crossover to be 0.5. The population size was 10 and we sampled new candidate points based on the 'rand/1/bin' strategy.

\textbf{Tree Parzen Estimator} (TPE)~\citep{bergstra-nips11a} is a Bayesian optimization method that uses kernel density estimators (KDE) to model the probability of 'good' points in the input space, i e. that achieve a function value that is lower than a certain value and 'bad' points that achieve a function value smaller than a certain value. TPE selects candidates by maximizing the ration between the likelihood of the two KDEs which is equivalent to optimizing expected improvement. We used the default hyperparameters provided by the hyperopt (\url{https://github.com/hyperopt/hyperopt}) package.

\textbf{SMAC}~\citep{hutter-lion11a} is a Bayesian optimization method that uses random forests to model the objective function and stochastic local search to optimize the acquisition function.
We used the default hyperparameters of SMAC, and set the number of trees for the random forest to 10.

\textbf{CMA-ES}~\citep{hansen-eda06} is an evolutionary strategy that models a population of points as a multivariate normal distribution. We used the open source pycma package (\url{https://github.com/CMA-ES/pycma}).
We set the initial standard deviation of the normal distribution to 0.6.

\textbf{Gaussian Process based Bayesian optimization} (BO-GP) as described by~\citet{snoek-nips12a}. We used expected improvement as acquisition function and an adapted random search strategy to optimize the acquisition function, which given a maximum number of allowed points $N=500$ samples first 70\% uniformly at random and the rest from a Gaussian with a fixed variance around the best observed point.
While other methods, such as gradient ascent techniques or continuous global optimization methods could also be used, we found this to work faster and more robustly.
We marginalized the acquisition function over the Gaussian process hyperparameters~\citep{snoek-nips12a} and used the emcee package (\url{http://dfm.io/emcee/current/}) to sample hyperparameter configurations from the marginal log-likelihood. We used a Matern 52 kernel for the Gaussian process.

\textbf{BOHAMIANN}~\citep{springenberg-nips16} uses a Bayesian neural network inside Bayesian optimization where the weights are sampled based on stochastic gradient Hamiltonian Monte-Carlo~\citep{chen-icml14}. We use a step length of $10^{-2}$ for the MCMC sampler and increased the number of burnin step by a factor of 100 times the number of observed data points. In each iteration, we sampled 100 weight vectors over 10000 MCMC steps. We used the same random search method to optimize the acquisition function as for BO-GP.

All methods started from a uniformly sampled point and we estimated the incumbent after each function evaluation as the point with the lowest observed function value.

\begin{figure*}[h]
\centering
 \includegraphics[width=0.32\textwidth,valign=t]{plots/cdf_meta_svm_noiseless.pdf}
 \includegraphics[width=0.32\textwidth,valign=t]{plots/ranks_meta_svm_noiseless.pdf}
 \includegraphics[width=0.32\textwidth,valign=t]{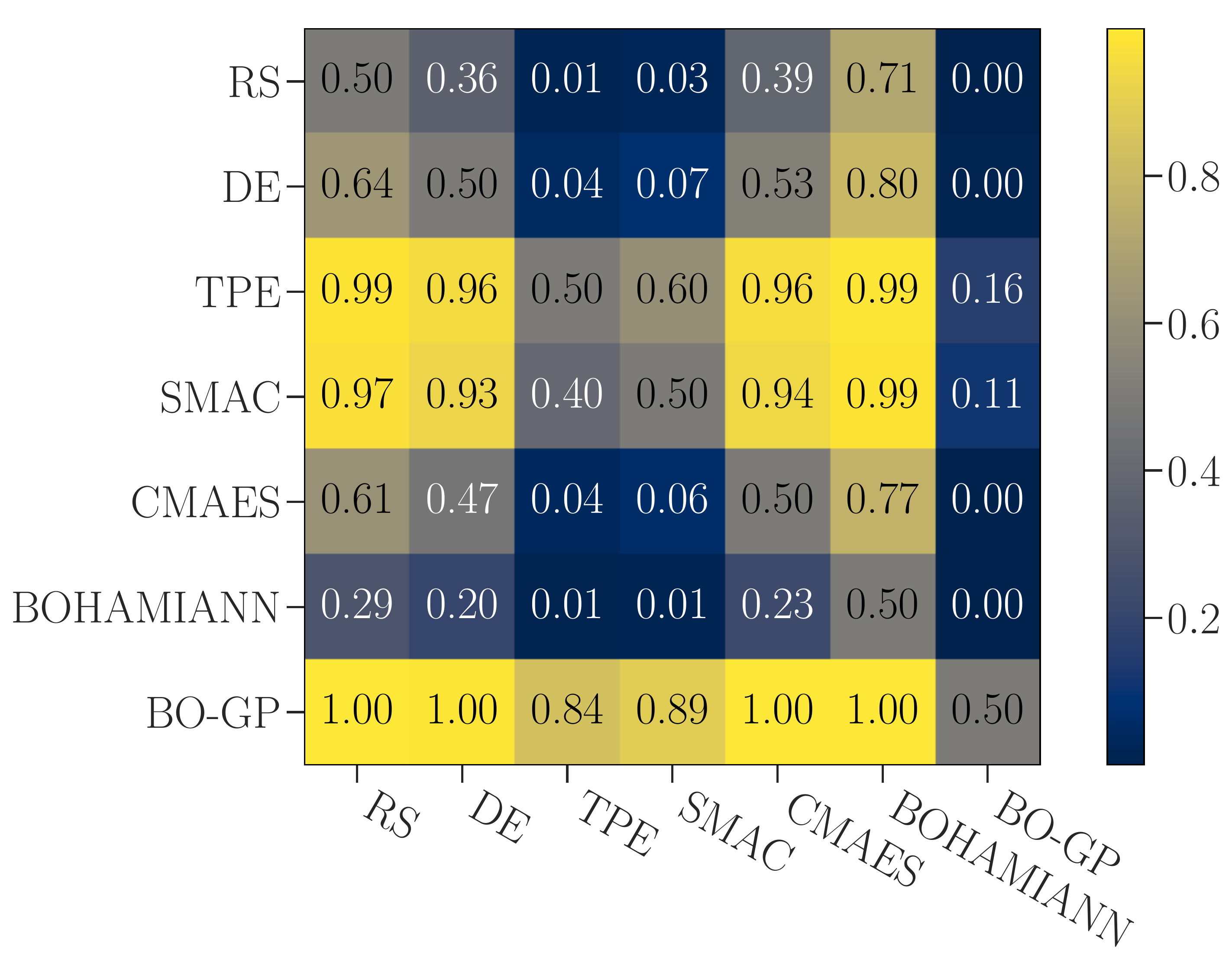}\\
 \includegraphics[width=0.32\textwidth,valign=t]{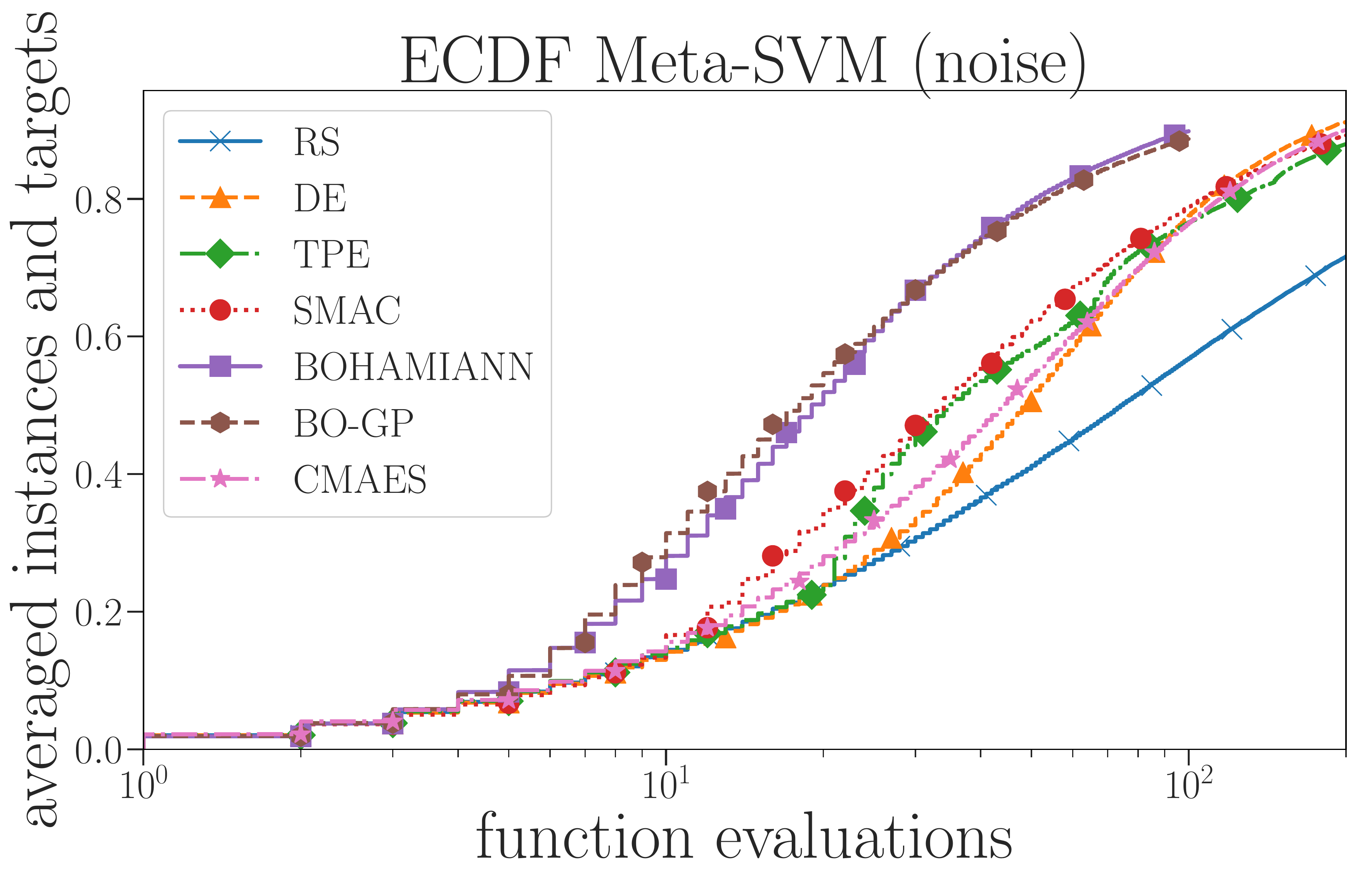}
 \includegraphics[width=0.32\textwidth,valign=t]{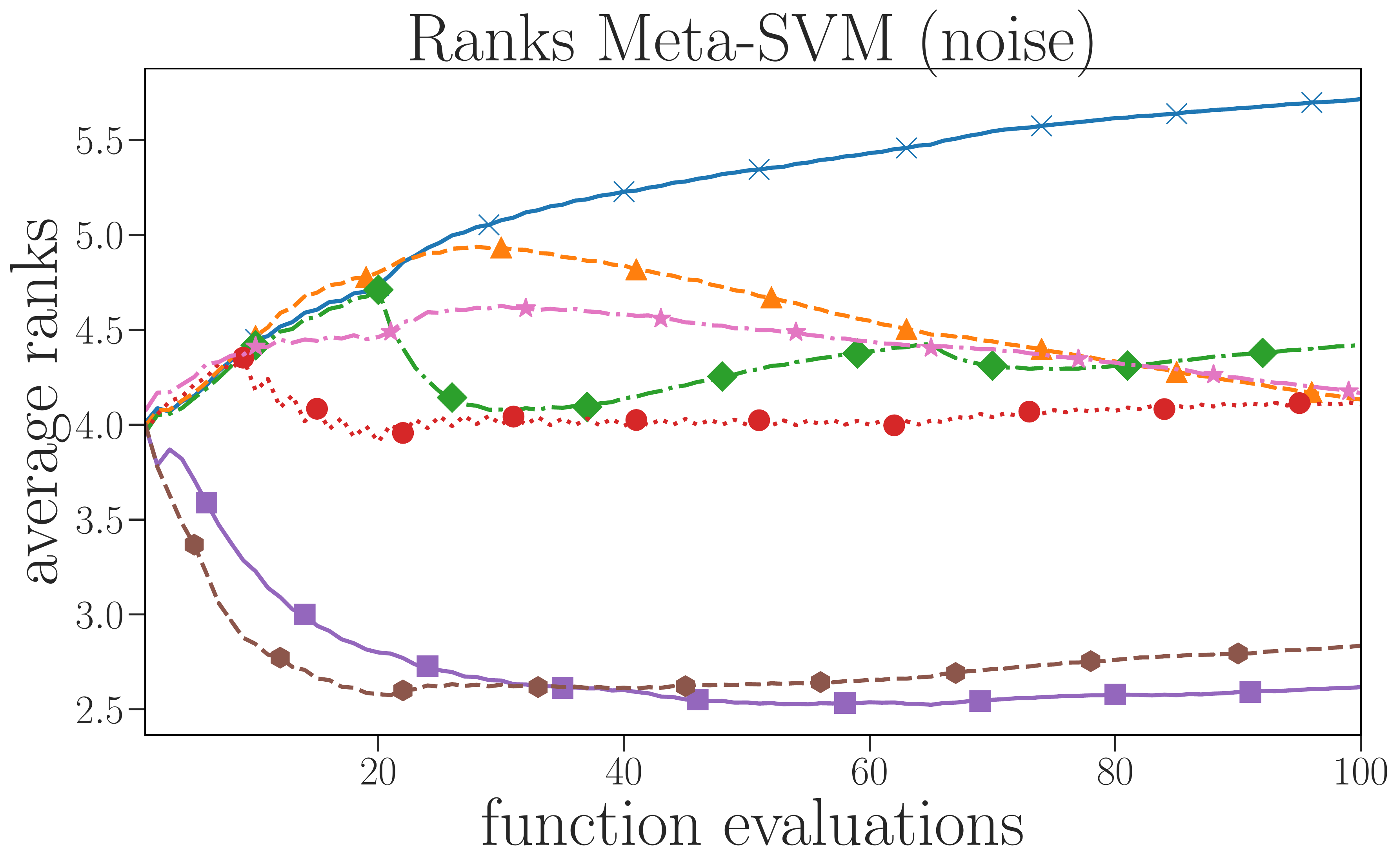}
 \includegraphics[width=0.32\textwidth,valign=t]{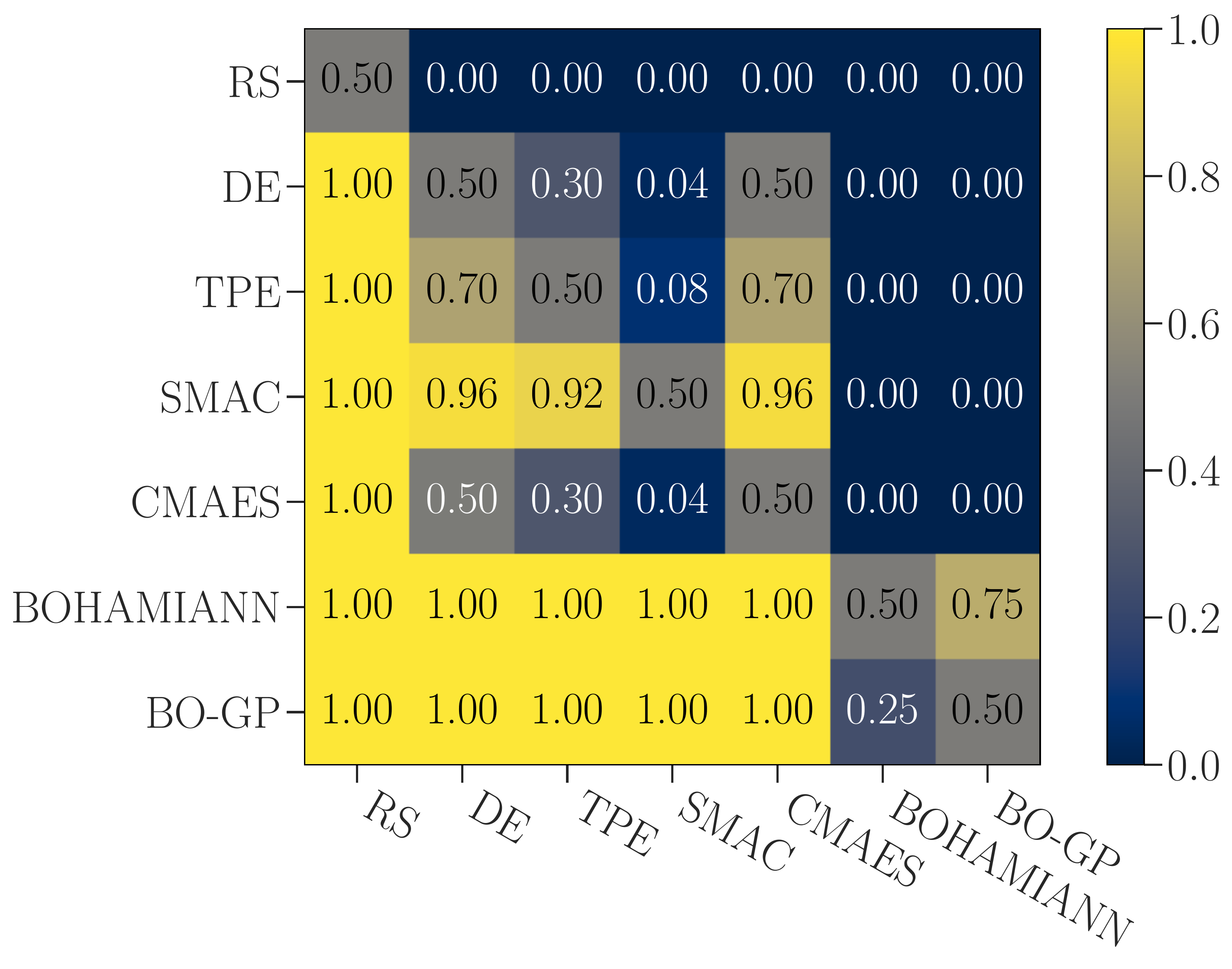}\\
 \includegraphics[width=0.32\textwidth,valign=t]{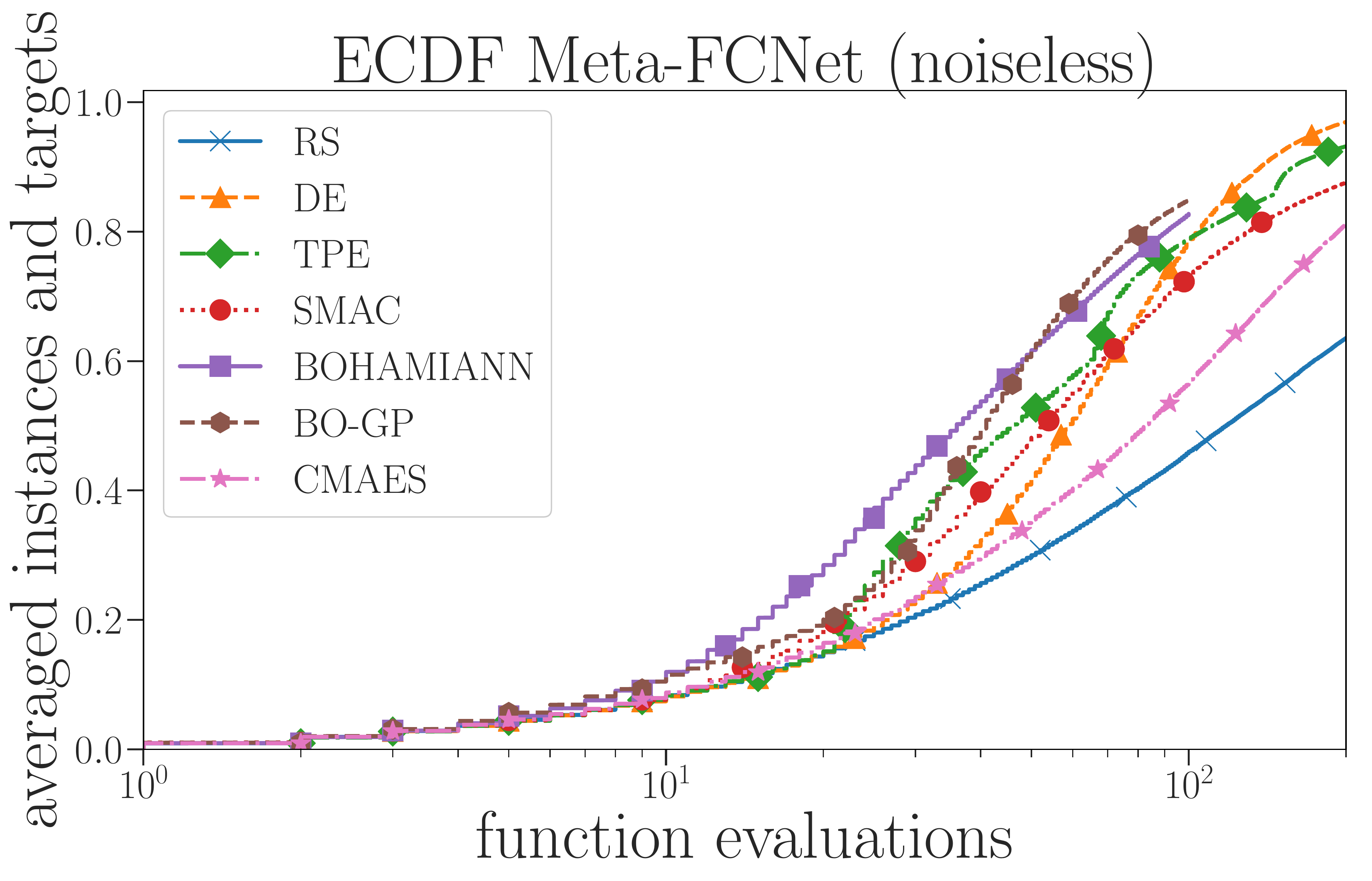}
 \includegraphics[width=0.32\textwidth,valign=t]{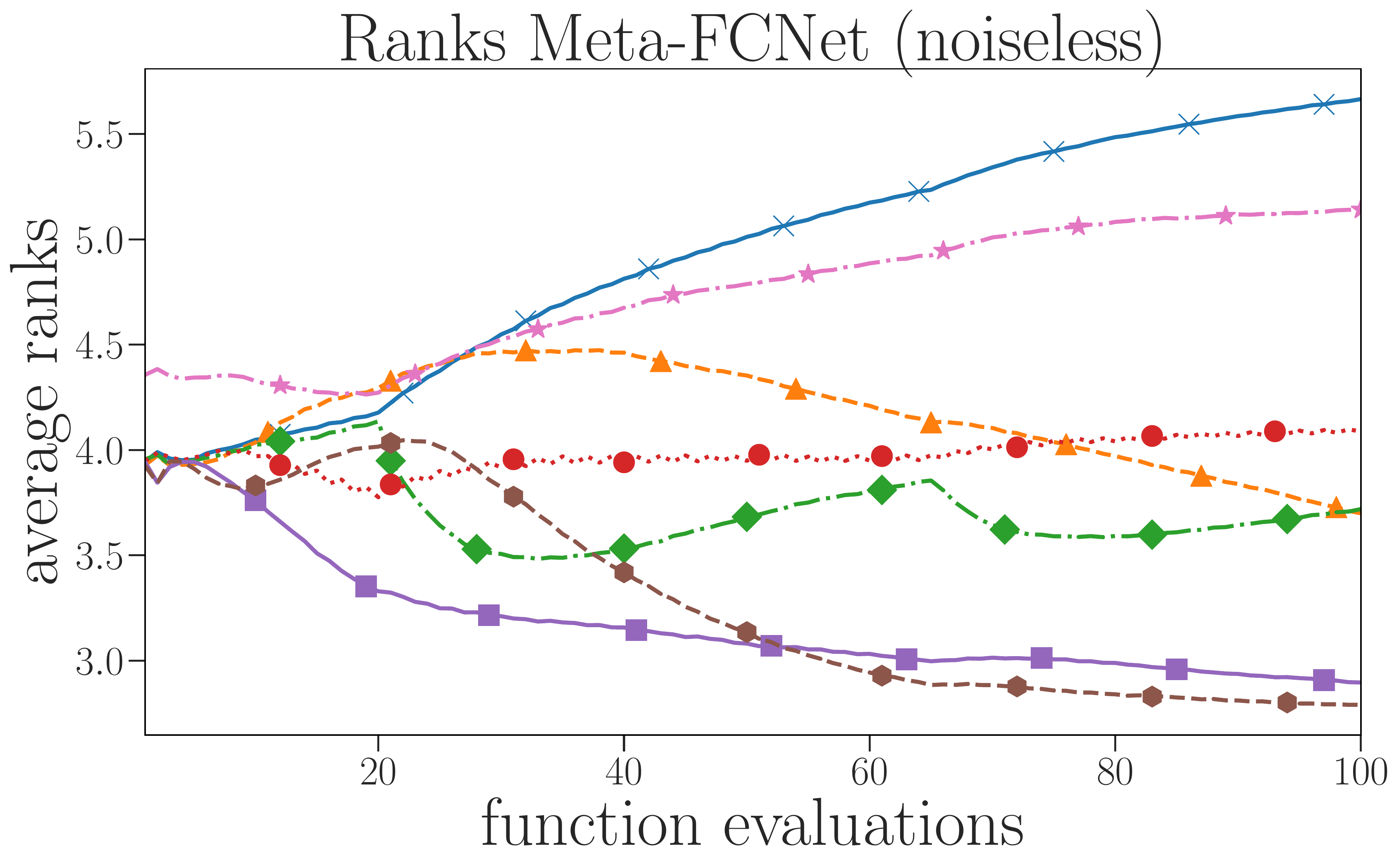}
 \includegraphics[width=0.32\textwidth,valign=t]{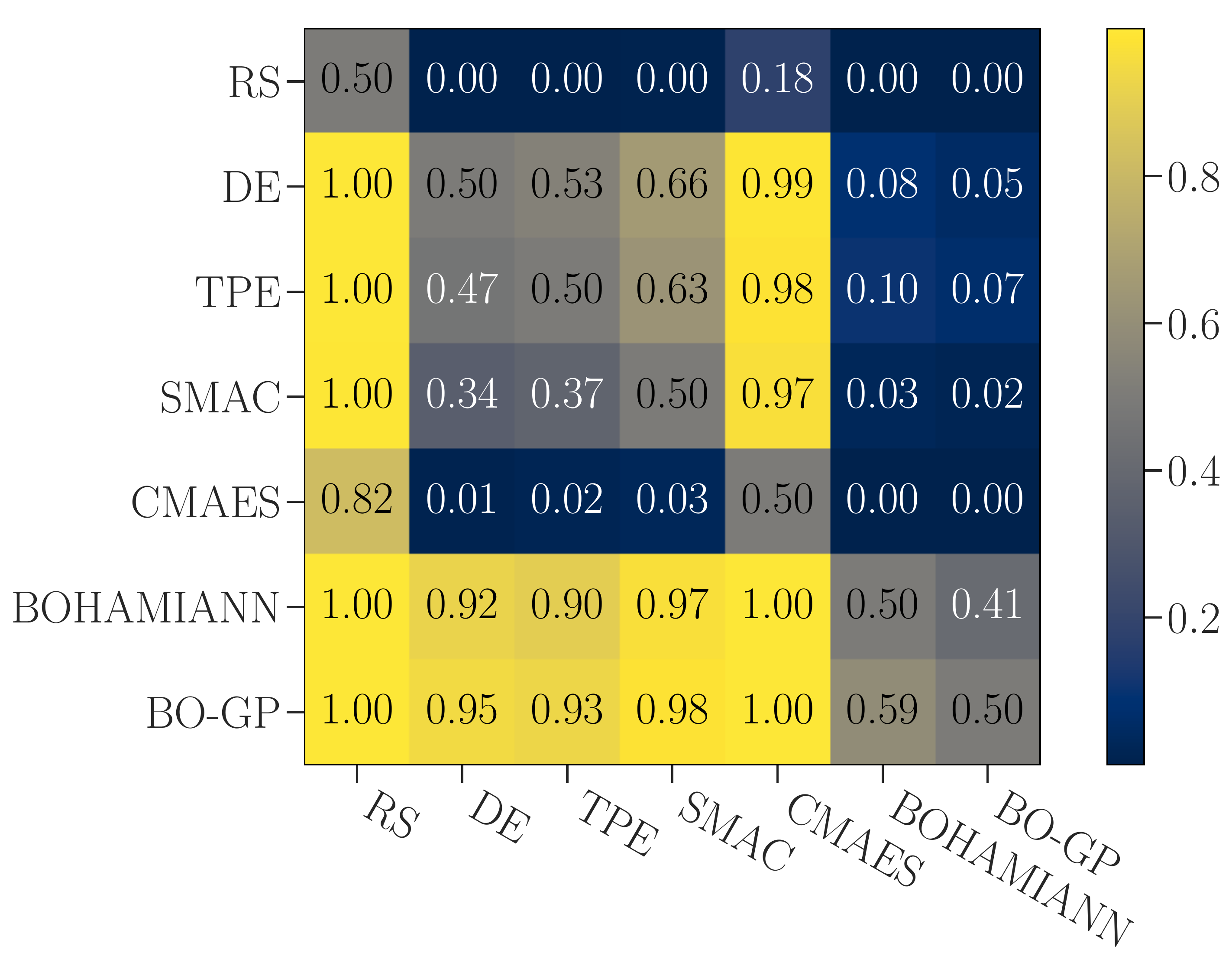}\\
 \includegraphics[width=0.32\textwidth,valign=t]{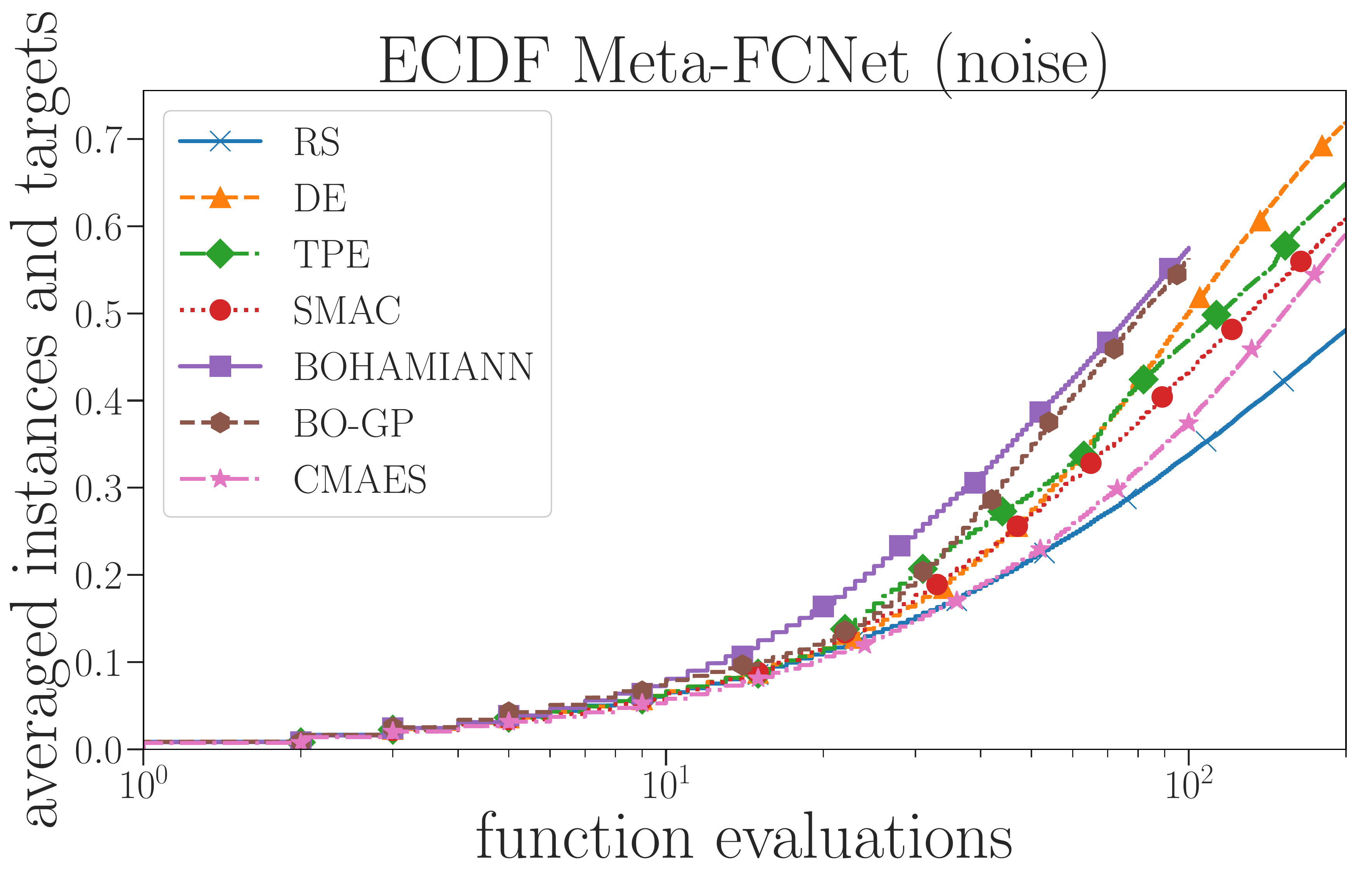}
 \includegraphics[width=0.32\textwidth,valign=t]{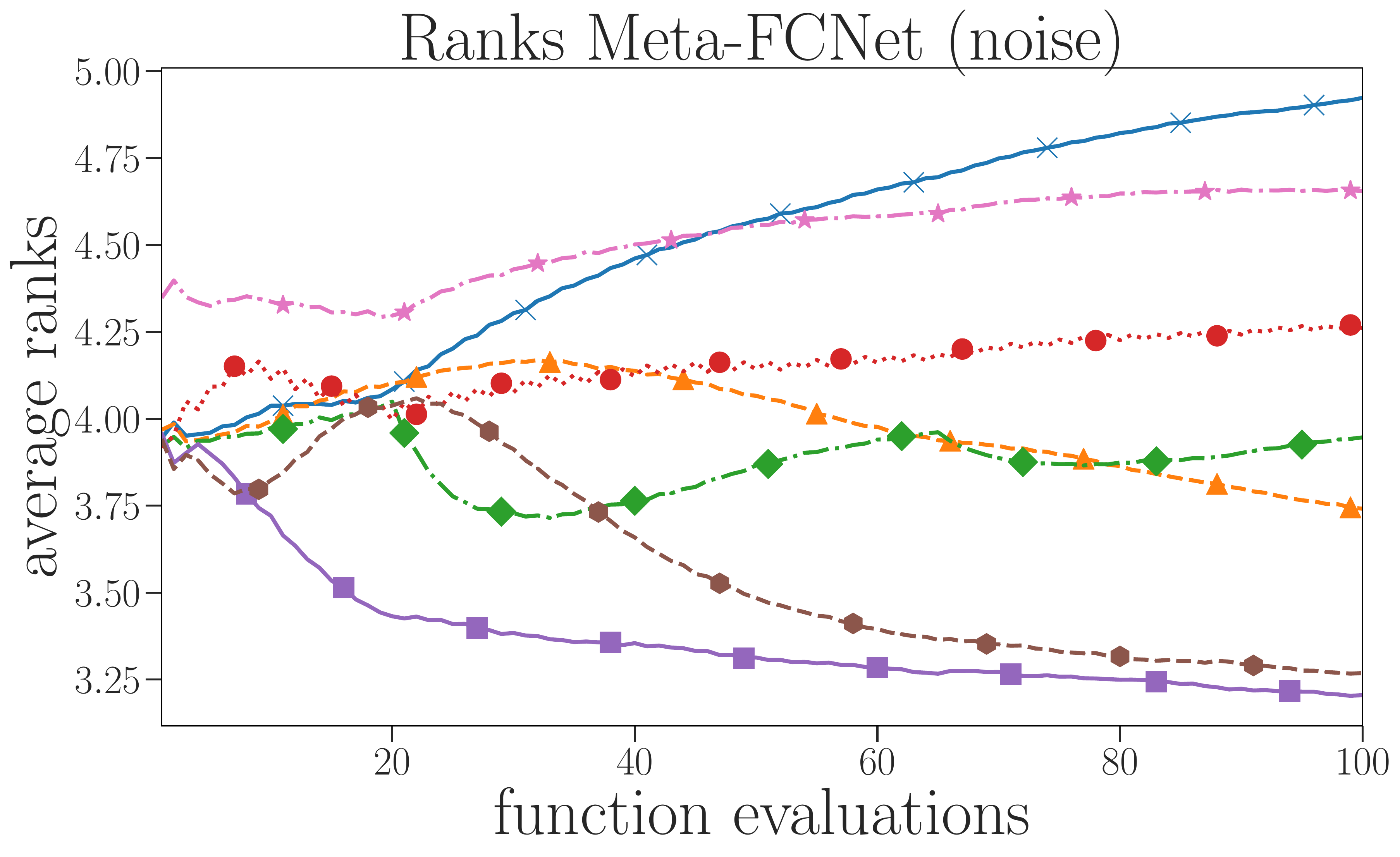}
 \includegraphics[width=0.32\textwidth,valign=t]{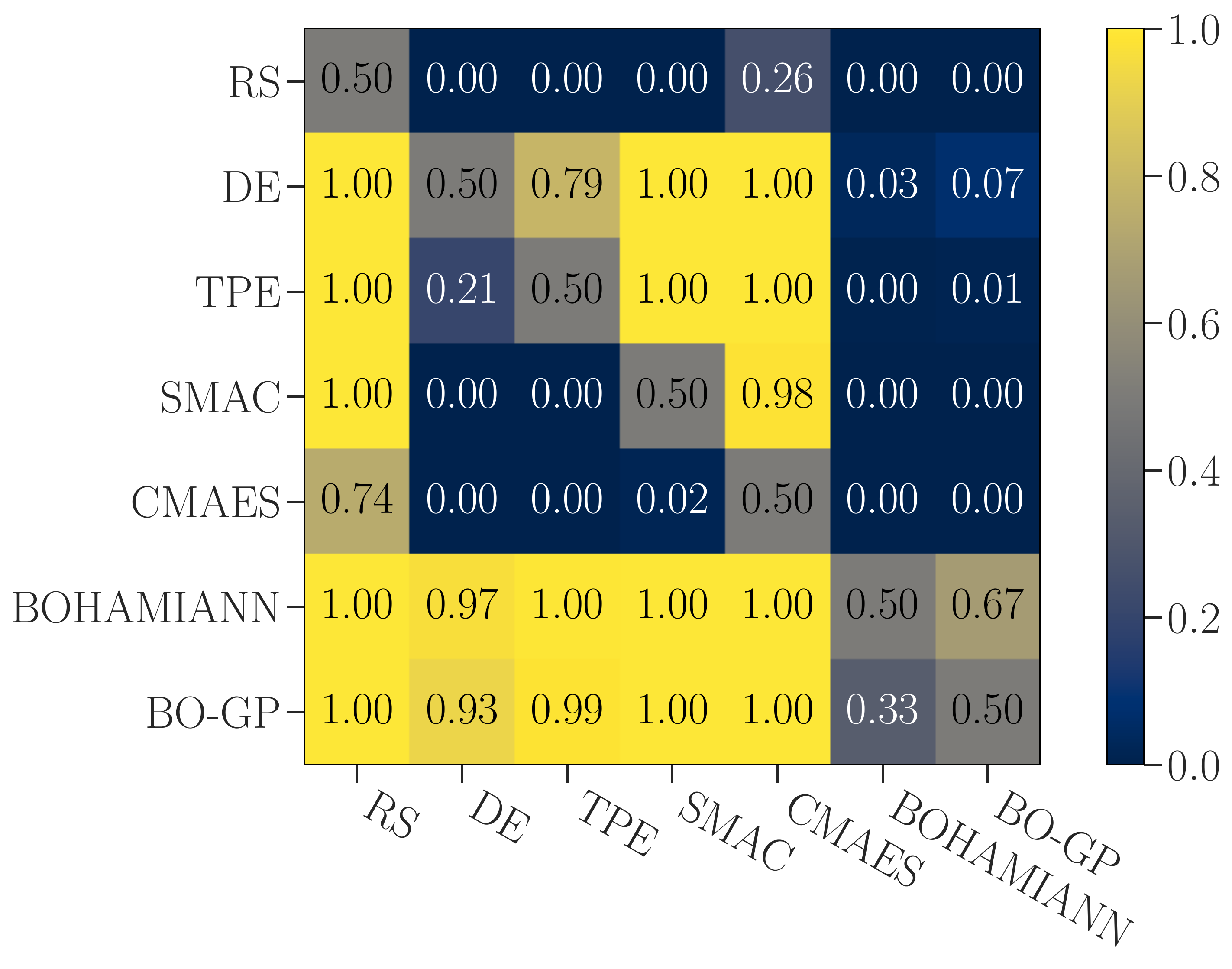}\\
 \includegraphics[width=0.32\textwidth,valign=t]{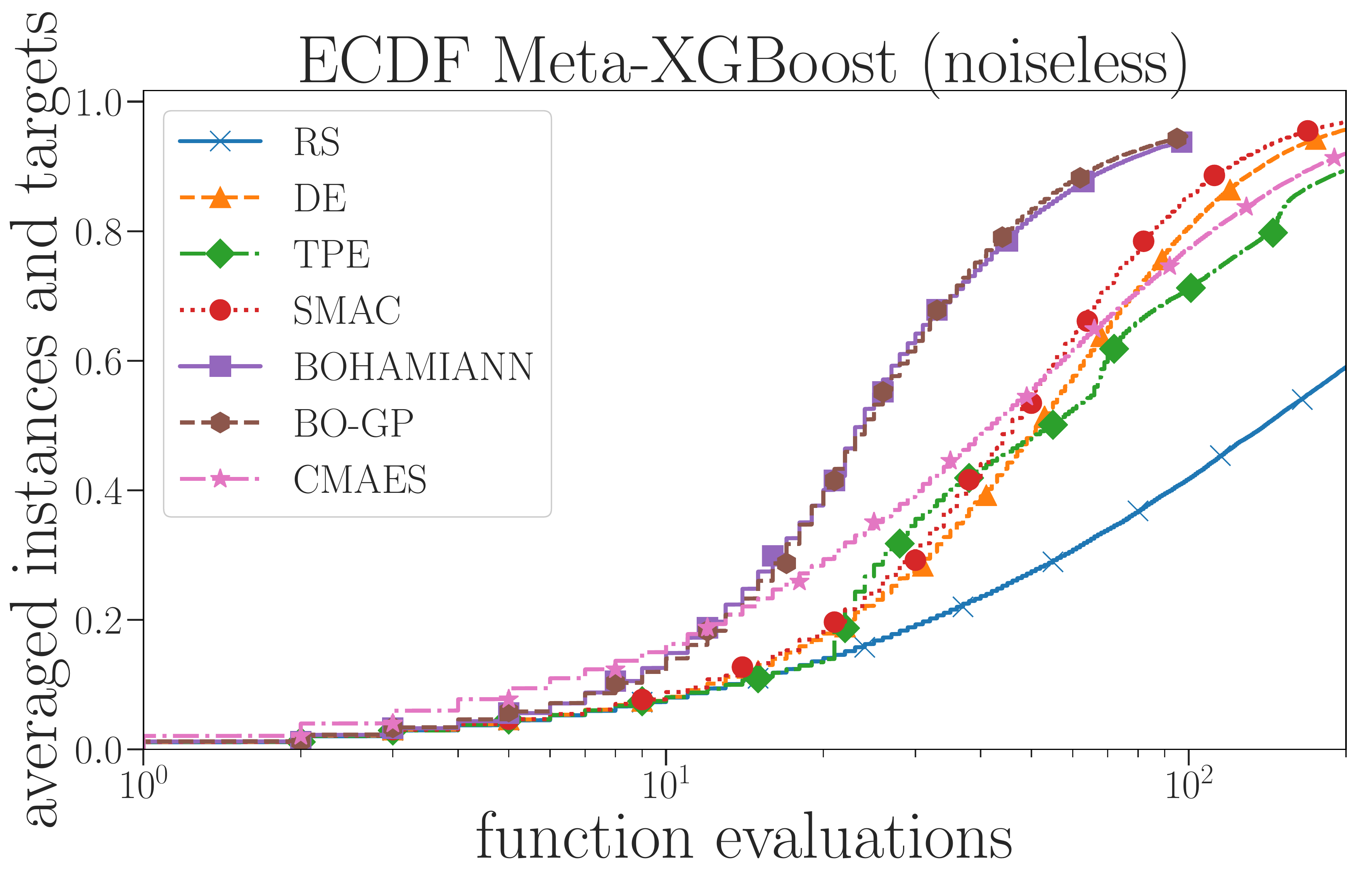}
 \includegraphics[width=0.32\textwidth,valign=t]{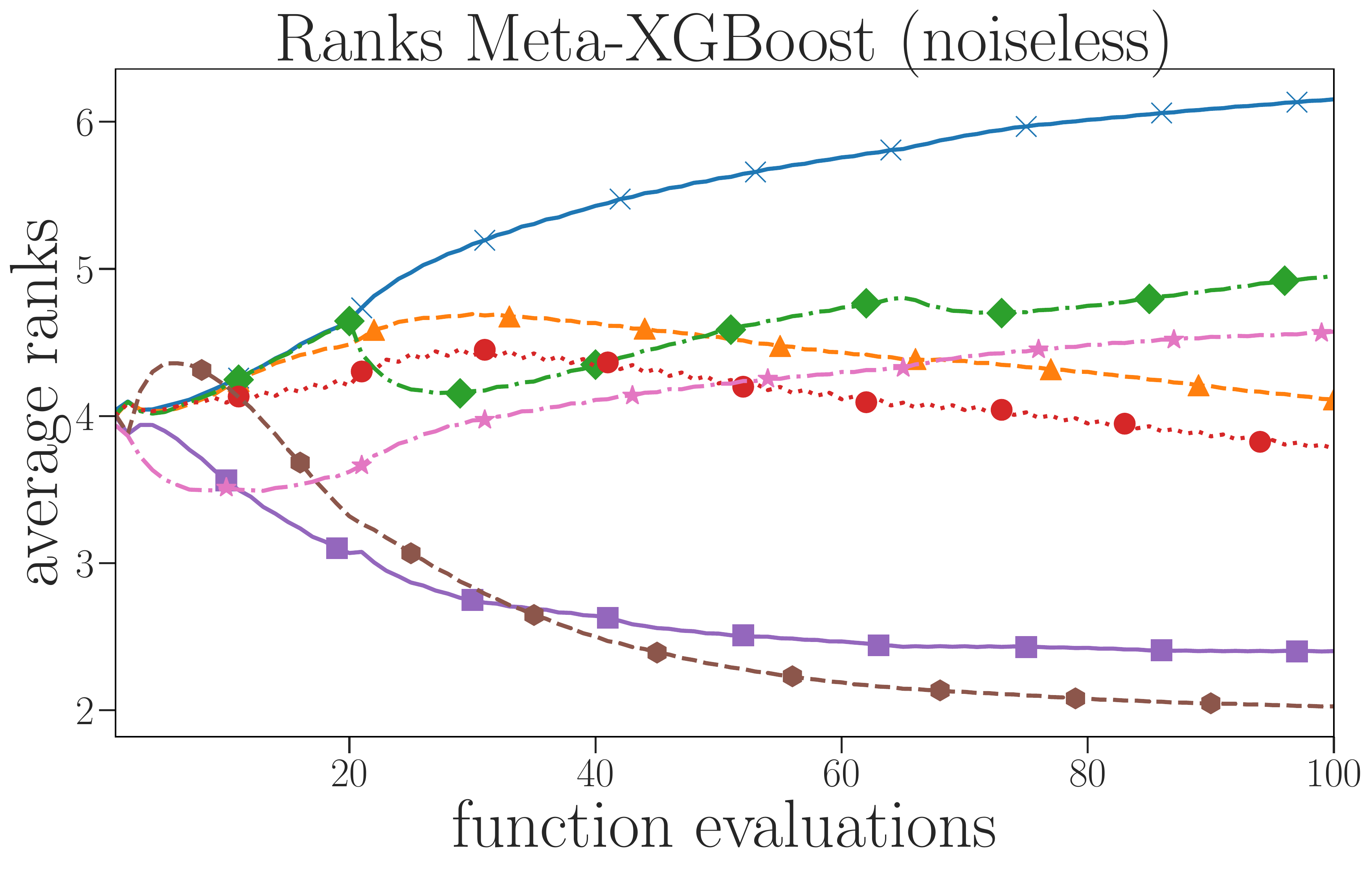}
 \includegraphics[width=0.32\textwidth,valign=t]{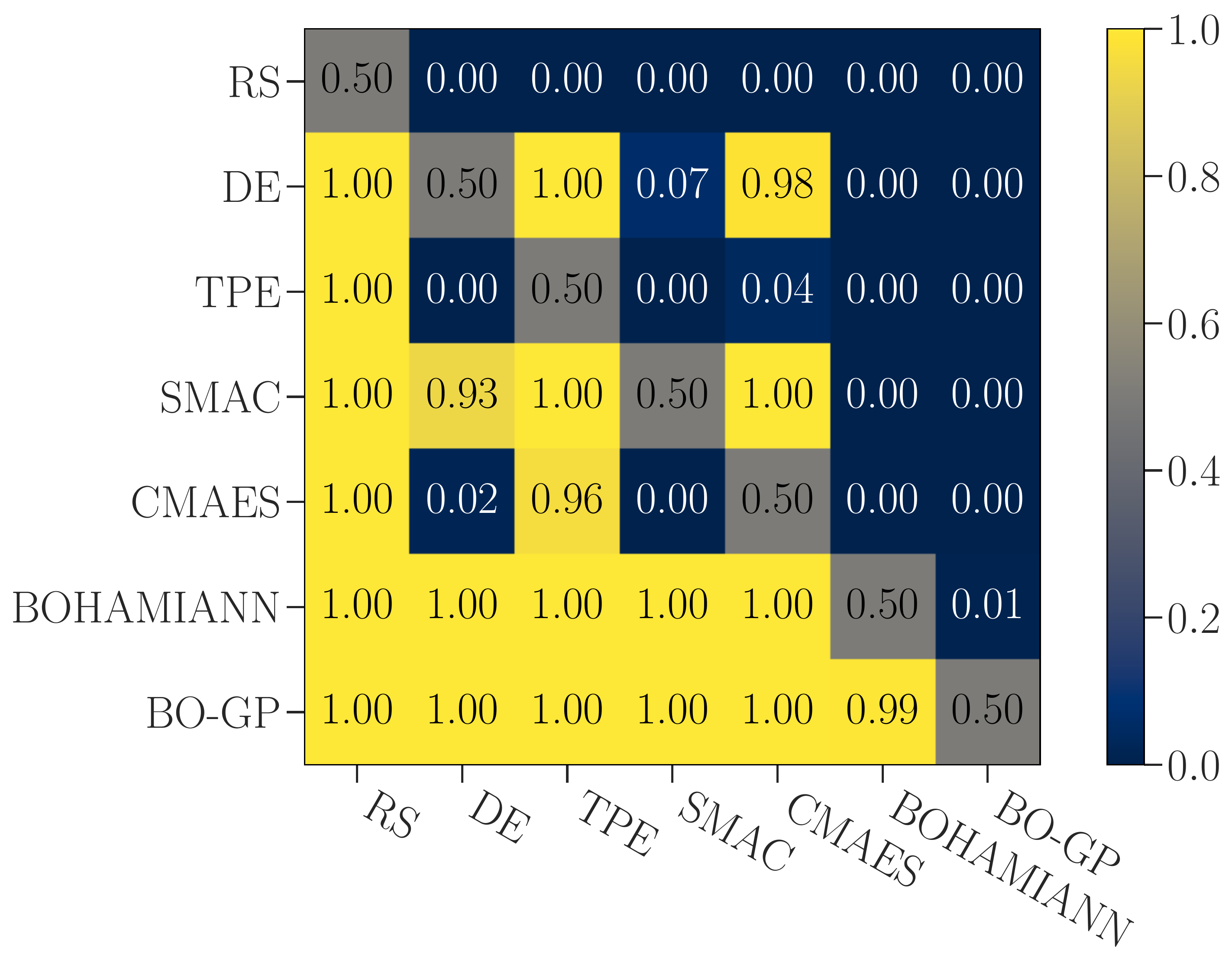}\\
 \includegraphics[width=0.32\textwidth,valign=t]{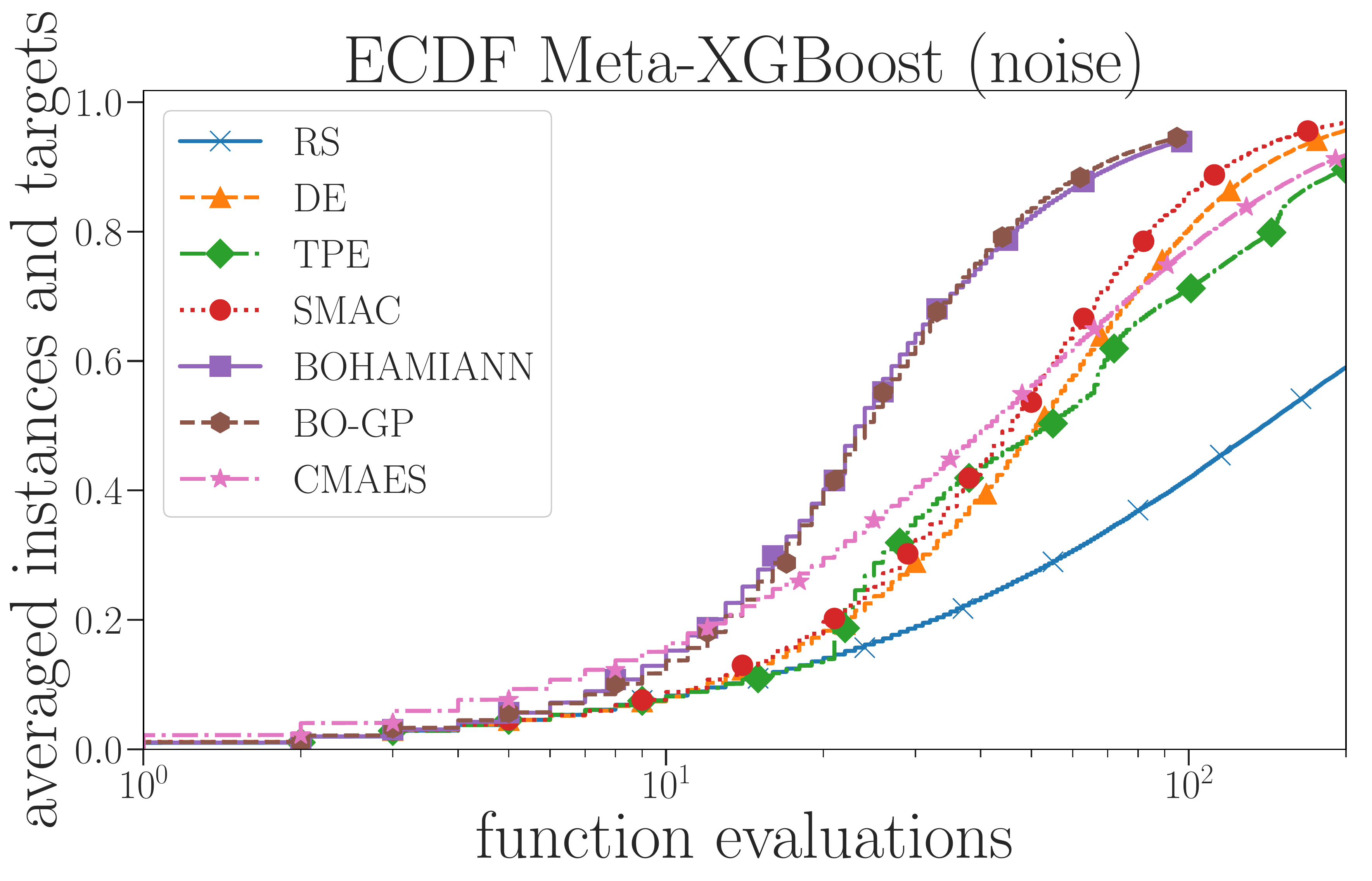}
 \includegraphics[width=0.32\textwidth,valign=t]{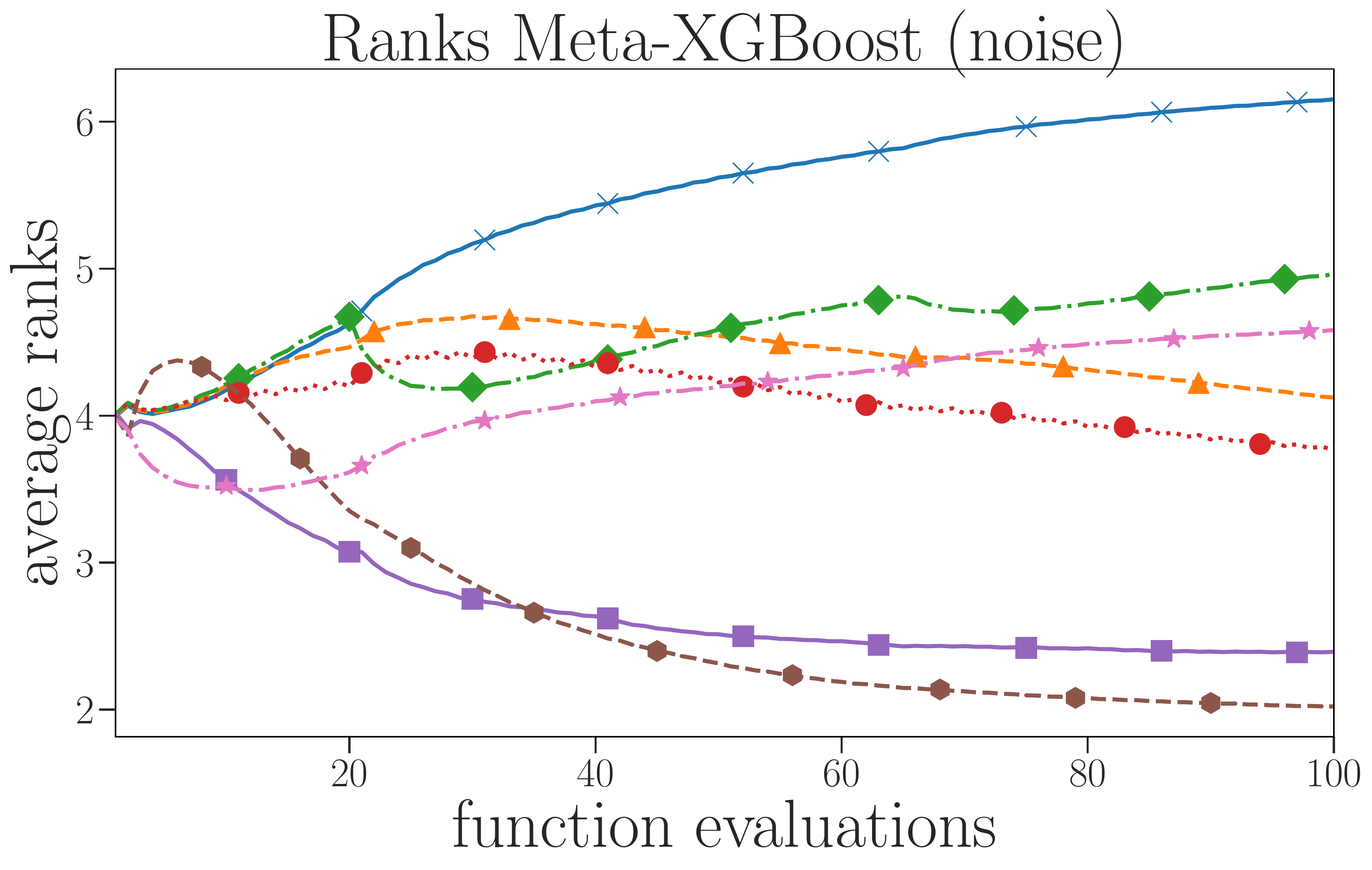}
 \includegraphics[width=0.32\textwidth,valign=t]{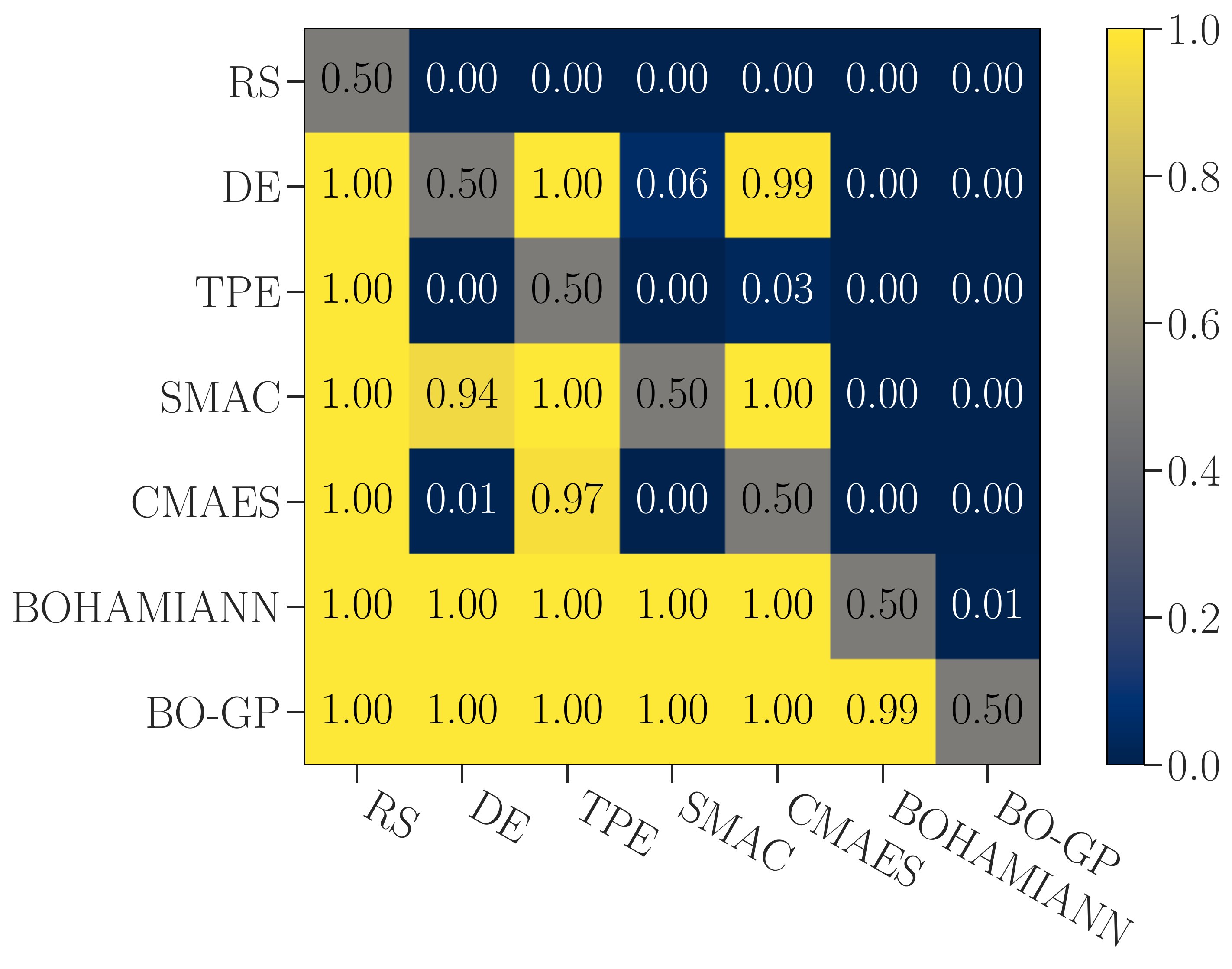}\\
 \caption[Aggregated performance for all benchmarks]{Comparison of various different methods on all three \hpo problems. From above to below 2-dimensional support vector machine, 6-dimensional feed-forward neural network and 8-dimensional XGBoost. The first column shows the ECDF, the second column the ranking and last column the p-values of the Mann-Whitney U test for the noisy and noiseless version of each \hpo problem.}
 \label{fig:results_all_profet}
\end{figure*}

In Figure~\ref{fig:results_all_profet} and Table~\ref{tab:results_all_profet} we show the aggregated results based on the runtime and the ranking for all methods on all three benchmarks.
We also show in Figure~\ref{fig:results_all_profet} the p-values of the Mann-Whitney U test between all methods.
For a detailed analysis of the results see Section 5.3 in the main paper.

\begin{table*}[h!]
\centering
\scriptsize
\begin{tabular}{cccccccc}\hline
Benchmark & RS & DE & TPE & SMAC & BOHAMIANN & CMAES & BO-GP \\\hline
Meta-SVM (noiseless) & $52.19 $ & $74.37$ & $79.64$ & $73.77$ & $90.33$ & $73.69$ & $\mathbf{98.88}$ \\ 
Meta-SVM (noise) & $56.64$ & $77.29$ & $76.44$ & $78.56$ & $\mathbf{89.80}$ & $76.27$ & $88.70$ \\ 
  Meta-FCNet (noiseless) & $45.71$ & $77.99$ & $78.73$ & $72.71$ & $82.50$ & $56.31$ & $\mathbf{84.71}$ \\ 
  Meta-FCNet (noise) & $33.66$ & $49.88$ & $46.84$ & $43.09$ & $\mathbf{57.28}$ & $37.41$ & $56.04$ \\ 
  Meta-XGBoost (noiseless) & $41.59$ & $80.35$ & $71.02$ & $84.95$ & $94.01$ & $77.17$ & $\mathbf{94.69}$ \\ 
  Meta-XGBoost (noise) & $41.71$ & $80.05$ & $71.05$ & $85.34$ & $94.23$ & $77.15$ & $\mathbf{94.87}$ \\ 

\hline
Meta-SVM (noiseless) & $5.89$ & $4.47$ & $4.50$ & $4.64$ & $2.75$ & $4.52$ & $\mathbf{1.22}$ \\ 
  Meta-SVM (noise) & $5.72$ & $4.13$ & $4.42$ & $4.11$ & $\mathbf{2.62}$ & $4.17$ & $2.84$ \\ 
  Meta-FCNet (noiseless) & $5.67$ & $3.70$ & $3.72$ & $4.09$ & $2.90$ & $5.14$ & $\mathbf{2.79}$ \\ 
  Meta-FCNet (noise) & $4.92$ & $3.74$ & $3.95$ & $4.26$ & $\mathbf{3.21}$ & $4.66$ & $3.27$ \\ 
  Meta-XGBoost (noiseless) & $6.15$ & $4.11$ & $4.95$ & $3.78$ & $2.40$ & $4.57$ & $\mathbf{2.03}$ \\ 
  Meta-XGBoost (noise) & $6.15$ & $4.12$ & $4.96$ & $3.76$ & $2.39$ & $4.58$ & $\mathbf{2.02}$ \\ 

\hline
\end{tabular}\caption[Averaged runtime and ranks after 100 function evaluations]{\emph{Top}: Each element of the table shows the averaged runtime after 100 function evaluations for each method-benchmark pair. \emph{Bottom}: Same but for the ranking of the methods.}
 \label{tab:results_all_profet}
\end{table*}

\section{Details of the Meta-Model}\label{sec:meta_model_details_supp}

The neural network architecture for our meta-model consisted of $3$ fully connected layers with $500$ units each and tanh activation functions.
The step length for the MCMC sampler was set to $10^{-2}$ and we used the first 50000 steps as burn-in.
For the probabilistic encoder, we used Bayesian \gplvm\footnote{We used the implementation from~\citet{gpy2014}}\citep{titsias-aistats10} with a Matern52 kernel to learn a $Q=5$ dimensional latent space for the task description.

\section{Pseudo Code for Profet}

Algorithm~\ref{alg:hposcore} shows pseudo code to evaluate an algorithm $\alpha$ with Profet by sampling new surrogate tasks (see Algorithm~\ref{alg:sampling}) sampled from our meta-model.

\begin{algorithm}[h!]
   \caption{Evaluating the performance of \hpo methods. \\
\emph{Inputs}: Datasets $\mathcal{D}_t = \{(\vx_{ti}, y_{ti})\}_{i=1}^{N_t}$ for $t=1,\dots,T$ tasks. \hpo method $\alpha$. Number of tasks $M$. }
   \label{alg:hposcore}
\begin{algorithmic}
	\State Train the probabilistic encoder $p(\vh_t \mid \vy)$ and multi-task model $p(\theta | \mathcal{D})$ as described in Section 3.2 on dataset $\mathcal{D}$
	\State Sample $M$ tasks using Algorithm \ref{alg:sampling}
	\State Solve $t_1 , \dots, t_M$ using $\alpha$ and compute $r(\alpha,t_m)$
	\State Approximate $S_p(\alpha)$ in Equation 1 using $\frac{1}{M}\sum_{m=1}^Mr(\alpha,t_m)$.
\end{algorithmic}
\end{algorithm}

\begin{algorithm}[h!]
  \label{alg:sampling}
   \caption{Sampling new tasks. \\
     \emph{Inputs}: $noiseless \in \{true, false \}$, encoder $p(\vh_t \mid \vy)$ and multi-task model $p(\theta | \mathcal{D}$  as described in Section 3.}
    \label{alg:sampling}

     \begin{algorithmic}
	\State Sample latent task vector $\vh_{t_{\star}} \sim p(\vh_t \mid \vy)$.
	\State Sample a set of weights $\theta \sim p(\theta | \mathcal{D})$ from the posterior of the \bnn.
	\If{$noiseless == true$}\\ 
	$f_{t_{\star}}(\vx) = \hat{\mu}(\vx, \vh_{t_{\star}} | \theta)$
	\Else\\
	$f_{t_{\star}}(\vx) = \hat{\mu}(\vx, \vh_{t_{\star}}  | \theta) + \epsilon \cdot \sigma(\vx, \vh_{t_{\star}}  | \theta) $ where $\epsilon \sim \gauss(0, 1)$
	\EndIf\\
      	\Return $f_{t_{\star}}(\vx)$ 
\end{algorithmic}
\end{algorithm}

\end{document}